\documentclass{article}

\usepackage[main, final]{neurips_2026}
\usepackage{amsmath, amssymb, amsthm, mathtools}
\usepackage{bm}
\usepackage{bbm}
\usepackage{enumitem}
\usepackage{hyperref}
\usepackage[nameinlink,noabbrev]{cleveref}
\usepackage[]{natbib} 
\usepackage{algorithm}
\usepackage{algpseudocode}
\usepackage{graphicx}
\usepackage{subcaption}
\usepackage{tabularx}
\usepackage{booktabs} 
\usepackage{bm}       
\usepackage{arydshln}

\usepackage[utf8]{inputenc} 
\usepackage[T1]{fontenc}    
\usepackage{hyperref}       
\usepackage{url}            
\usepackage{booktabs}       
\usepackage{amsfonts}       
\usepackage{nicefrac}       
\usepackage{microtype}      
\usepackage{xcolor}         
\usepackage{caption}
\usepackage{wrapfig}
\newcommand{\vv}[2]{$#1${\tiny$\,\pm#2$}}
\newcommand{\vb}[2]{$\bm{#1}${\tiny$\,\pm#2$}}
\newcommand{\oo}[4]{\vv{#1}{#2}\,$\rightarrow$\,\vv{#3}{#4}}
\newcommand{\ob}[4]{\vv{#1}{#2}\,$\rightarrow$\,\vb{#3}{#4}}

\newtheorem{assumption}{Assumption}
\newtheorem{definition}{Definition}
\newtheorem{proposition}{Proposition}
\newtheorem{theorem}{Theorem}

\newtheorem{corollary}{Corollary}
\newtheorem{remark}{Remark}

\title{Efficient Diffusion Policy Fine Tuning with Latent Noise Representation Bridging}

\author{
  Yuanchu Liang\\\
    \texttt{yuanchu.liang@anu.edu.au} \\
 The Australian National University.\\ Maincode.
  \And
  Yue Yang\footnotemark[1] \\
    \texttt{yue@maincode.com} \\
  Maincode.
  \And
   An-Chi He \\
 \texttt{anchi@maincode.com} \\
  Maincode.
  \And
  Hanna Kurniawati\footnotemark[1] \\
  \texttt{hanna.kurniawati@anu.edu.au}\\
 The Australian National University.
}

\begin{document}

\maketitle

\begin{abstract}
  Pre-trained diffusion and flow policies have emerged as powerful backbones for robotics, yet adapting them to novel data via reinforcement learning (RL) remains computationally expensive, sample inefficient and lacks theoretical guarantees. Prior approaches that fine-tune the initial noise bypass costly backpropagation through time but struggle with the "noise aliasing" problem: there does not exist a natural link between the noise space value and the action space value. This paper resolves these challenges by introducing the Latent Noise Linear Markov Decision Process (LNL-MDP). We derive the Bridge Equation for LNL-MDPs which eliminates the noise aliasing problem by aligning values with a shared representation between the noise and the action space. Furthermore, our LNL-MDP fine tuning framework falls under the FineTuneRL theory with near-optimal guarantee. Leveraging these insights, we propose Latent Representation Bridging for Diffusion (LARBRID), an algorithm that efficiently learns the representation to optimize the initial distribution that steers the pre-trained model to act optimally in the environment. It yields up to $2\text{-}3\times$ performance increases on challenging control problems across a diverse suite of robotic locomotion and manipulation tasks. Code can be found at: \url{github.com/Maincode-Research/LARBRID.git}.
\end{abstract}

\section{Introduction}
Pre-trained generative models with diffusion \citep{ho2020denoising, song2020denoising} and flow-based methods \citep{lipman2022flow} via behavior cloning on large scale datasets have become the central backbone of robotic foundation models \citep{black2024pi_0, liu2024rdt, team2024octo, chi2025diffusion, kawaharazuka2025vision}. However, these models often fail when dealing with tasks that are not covered in the training dataset. Recently, fine tuning pre-trained models with reinforcement learning (RL) has drawn much attention in the RL and robotics community. It enables learning from online experiences to adapt to the given tasks. Meanwhile, pre-trained models serve as a good prior for RL algorithms, as training them from scratch on complex tasks is notoriously difficult. Although many existing works focus on the integration of diffusion or flow frameworks with traditional RL pipelines \citep{hansenestruch2023idql, dppo2024, fql_park2025, wagenmaker2025steering}, they often encounter heavy computations and lack theoretical justifications on the number of extra fine tuning samples needed to achieve near optimality. 

While some early RL theory provides guarantees in tabular cases \citep{auer2008near, azar2017minimax, jin2018q}, there is a recent surge of interest in finding provably efficient RL structures for commonly encountered large or possibly continuous state and action spaces \citep{jin2020provably, agarwal2020flambe, uehara2021representation, jin2021bellman, chowdhury2021reinforcement, du2021bilinear, ouhamma2023bilinear}. From these theoretical advances, linear Markov Decision Processes (MDPs) \citep{jin2020provably, agarwal2020flambe, uehara2021representation} have been successfully applied by subsequent works to boost sample efficiencies in standard RL problems  \citep{ren2022free, zhang2022making, ren2022spectral, shribak2024diffusion}. Yet, little is known about whether the sample efficiency still holds analytically and practically for fine tuning pre-trained models on complex long horizon robotics problems.  

This work provides affirmative answers to both the theoretical and empirical aspects of utilizing linear MDPs for fine tuning pre-trained robotics generative models. We mainly consider fine tuning only the initial distribution of pre-trained diffusion/flow models \citep{wagenmaker2025steering, eyring2025noise}. Such methods can work for any model whose generative process is deterministic once the initial noise is sampled, e.g. DDIM \citep{song2020denoising} and flow \citep{lipman2022flow}. The benefit with this method is that the weights of the pre-trained model are frozen during fine tuning and the model can be viewed as a black box with only API access. Gradients pass straight to the initial noise network and hence avoid heavy back propagation through time. The key challenge of this setting is the \textbf{noise aliasing problem} \citep{wagenmaker2025steering}, multiple noises can produce the same action, or similar noises can produce different actions, which means that there is no natural association between the Q-values of the actions and the Q-values of the noises. The problem becomes worse in offline RL as out-of-distribution errors often require multiple ($>10$) Q-networks being created independently \citep{an2021uncertainty, wagenmaker2025steering}, amplifying the noise aliasing problem. 

\begin{figure}
    \centering
      \vspace{-2mm}
    \includegraphics[width=0.87\linewidth]{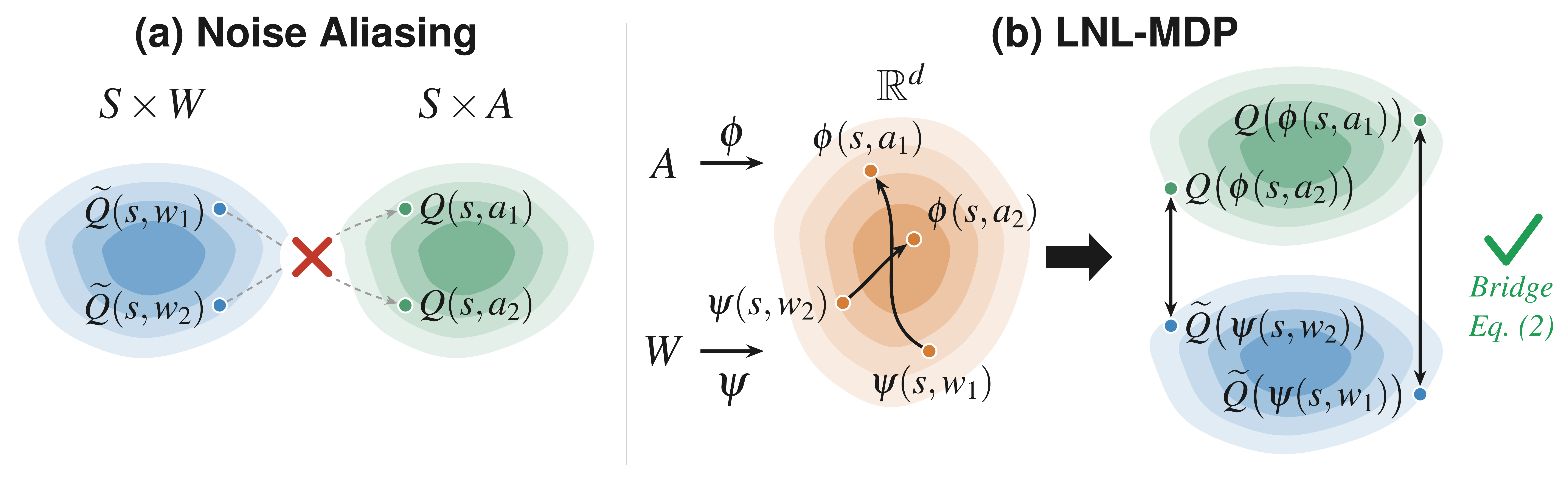}
    \caption{(a) Noise aliasing causes ambiguous associations between the action values $Q$ and noise values $\widetilde{Q}$. (b) LNL-MDP automatically solves the aliasing problem by mapping the state space $S$ and action space $A$ into a shared representation space via $\phi$ and $\psi$, from which the Q values are automatically aligned.}
    \label{fig:bridge-illustration}
        \vspace{-6mm}
\end{figure}

We address these issues analytically by endowing the environment MDP with linear structures such that components of the MDP (e.g. transition, reward, Q-values) can be represented by a single representation. Importantly, a pre-trained diffusion/flow policy operating over this linear MDP induces a Latent Noise Linear MDP (LNL-MDP) over the noise space. The corresponding latent representation is obtained by pulling back the environment representation through the pre-trained generative policy, so state-action and state-noise pairs share a common representation. This shared representation yields the Bridge Equation, which directly aligns action-space and noise-space Q-values without requiring separate critic networks, completely removes the noise-aliasing problem; see Figure~\ref{fig:bridge-illustration}. As a result, learning a single representation is sufficient to evaluate and optimize policies in both the original MDP and its LNL-MDP counterpart. Since much of the information about the environment is encoded into the representation, it is powerful enough to break the sub-optimality of pre-trained models and assist in finding better solutions.


By treating the pre-trained model as a virtual dataset, we can embed the LNL-MDP fine tuning problem under the theoretically well understood FineTuneRL framework \citep{wagenmaker2023leveraging} and inherit its near-optimal sample complexity bound. Our practical algorithm \textit{Latent Representation Bridging for Diffusion} (LARBRID) learns the shared representations from scratch and successfully transfers the theoretical justifications to empirically observed sample efficiencies and demonstrates up to $2\text{-}3\times$ performance increase compared to many state-of-the-art baselines across a wide range of robotic control tasks.

\section{Related Work}
\textbf{Behavior Cloning and Pre-training of Diffusion Policies}.
The success of training generative models on internet scale datasets for languages and visions has inspired the robotics and control community to pursue similar directions known as behavior cloning (BC) \cite{argall2009survey, ross2011reduction, bojarski2016end, zhang2018deep, brohan2022rt, kawaharazuka2025vision}, where robotic agents directly learn from offline demonstration datasets. Breakthroughs in diffusion \cite{ho2020denoising} and flow \cite{lipman2022flow} models have inspired many subsequent BC works to generate robotic controls with them \cite{chi2025diffusion, black2024pi_0, team2024octo}. Although many of these works demonstrate the increasing capabilities of robots handling diverse tasks, they often fail when encountering situations that are not well covered by the training dataset. \textit{Our work aims to fine tune pre-trained models and demonstrate improved empirical results.}

\textbf{Diffusion Fine Tuning with RL}. This is a vast domain with works mainly focusing on directly optimizing the model parameters to act optimally in given environments \citep{chen2022offline, ding2023consistency, hansenestruch2023idql, shao2024deepseekmath, kang2023efficient, psenka2023learning, dppo2024, fql_park2025, wagenmaker2025steering}. For example, \cite{dppo2024} proposes to optimize the first few steps of the diffusion process using PPO \cite{schulman2017proximal} and \cite{psenka2023learning} aims to optimize the diffusion model by matching the score with the Q-function. Unlike these methods, we focus on fine tuning only the initial distribution of the diffusion or flow process, bypassing the need to perform often unstable and computationally expensive backpropagation through time \cite{wagenmaker2025steering}. Unlike \cite{wagenmaker2025steering}, \textit{we rigorously build our algorithm on the linear MDP \cite{jin2020provably} framework and show theoretical and empirical evidence of sample efficiency for fine tuning diffusion policies.}

\textbf{RL under linear MDPs}. Early studies of linear structures in sequential decision processes are carried out in the bandit literature \cite{lattimore2020bandit}. Recently, linear MDPs are proposed \cite{jin2020provably, agarwal2020flambe, uehara2021representation} and shown to achieve sample efficient regret bounds. \cite{wagenmaker2023leveraging} later showed a near-optimal rate in the number of extra online interactions needed to find near-optimal policies while having access to an offline dataset. While the theoretical algorithms are not directly implementable, works such as \cite{ren2022free, zhang2022making, ren2022spectral, shribak2024diffusion} use different approximation schemes to bypass the intractability and demonstrate the practicality of linear MDPs on robotic control tasks. However, \textit{none of these works focus on demonstrating the effectiveness of linear MDPs for fine tuning generative models. It is even unclear whether the linear MDP structure can be properly defined. Our work provides positive answers to these doubts.}
\section{Preliminaries}
\textbf{Markov Decision Processes}. We consider an episodic Markov decision process \cite{puterman2014markov, wagenmaker2023leveraging},
$
M = (\mathcal S,\mathcal A,\{P_h\}_{h=1}^H,\{r_h\}_{h=1}^H,H,s_1),
$
with maximum horizon $H$, a fixed \footnote{the initial state can be probabilistic in our empirical studies} initial state \(s_1\), unknown state transition probability $\{P_h\}_{h=1}^H$ and unknown reward distribution $\{r\}_{h=1}^H$ supported on $[0,1]$. A policy is written as \(\pi=(\pi_h)_{h=1}^H\). For any policy \(\pi: \mathcal S \times [H] \rightarrow \Delta_{\mathcal A}\), we define the Q-value as, 
$Q^\pi_h(s,a) = \mathbb{E}_\pi \left[\sum_{h'=h}^H r_{h'}(s_{h'}, a_{h'})\mid s_h=s, a_h = a\right],$
and the corresponding V-value as  \(V_h^\pi(s) =\mathbb{E}_{a\sim\pi}[Q^\pi_h(s,a)]\). Let $V_0^\pi := V_1^{\pi}(s_1)$ denote the value of the policy $\pi$. Then the optimal policy is the one that maximizes the value $V_0^\star = V_0^{\pi^\star}$. The following definition characterizes the notion of linear MDP,
\begin{definition}[Linear MDP \cite{jin2020provably, wagenmaker2023leveraging}]
\label{def:linear_mdp}
There exists a known feature map $
\phi^\star : \mathcal S\times\mathcal A \to \mathbb R^d
$
such that \(\|\phi_\star(s,a)\|_2\le 1\) for all \((s,a)\), and for each stage \(h\in[H]\) there exist a signed vector-valued measure \(\nu_h: S\rightarrow\mathbb{R}^d\) and an unknown reward vector \(\theta^r_h\in\mathbb R^d\) satisfying
$
\|\nu_h(\mathcal S)\|_2\le \sqrt d,
$ and $
\|\theta_h\|_2\le \sqrt d,
$
such that
\[
P_h(s'\mid s,a)=\langle \phi^\star(s,a),\nu_h(s')\rangle,
\qquad
\mathbb{E}[r_h(s,a)]=\langle \phi^\star(s,a),\theta^r_h\rangle.
\]
\end{definition}
As shown in \cite{jin2020provably}, the advantage of this linear structure is that the Q-value can be represented linearly with the feature map $\phi$ as,
\begin{equation}\label{eq: linear q}
    Q^\pi_h(s,a) = \langle \phi^\star(s,a), \zeta^\pi_h \rangle 
\end{equation}
for some learnable weights $\{\zeta_h^\pi \in \mathbb{R}^d\}_{h=1}^H$. Linear MDP can model a wide range of non-linear dynamics when considering infinite dimensional cases (e.g. kernel embeddings), see Appendix~\ref{appendix: linear mdp example}.

\textbf{Diffusion and Flow Models}. Diffusion and flow models gradually transform an easy to sample initial distribution (e.g. standard Gaussian) to a complex goal distribution $q(x_0)$ \cite{ho2020denoising, lipman2022flow}. Here we brief the popular Denoising Diffusion Probabilistic Model (DDPMs). Its central transform process is governed by $p_\theta(x_0) = \int p_\theta(x_{0: T})dx_{1: T}$, where the joint distribution can be decomposed to $p_\theta(x_{0: T}) = p_\theta(x_T) \prod ^T_{t=1} p_\theta^{(t)}(x_{t-1} \mid x_t)$. The forward process samples $x \sim q(x_0)$ and gradually adds noise $\epsilon_t \sim \mathcal N (0, I)$ to the sample for $T$ steps. The reverse denoising process trains a network $\epsilon_\theta^{(t)}$ to predict the noise added at time $t$ and obtain a denoised sample via $x_{t-1} = \alpha_t(x_t - \beta_t \epsilon_\theta^{(t)}(x_t)) + \sigma_t\epsilon_t$ for coefficients $\alpha_t, \beta_t, \sigma_t$. DDPM sampling process can be formulated as stochastic differential equations \cite{anderson1982reverse}, however, when $\sigma_t = 0$, the process becomes deterministic and boils down to the Denoising Diffusion Implicit Model (DDIM) \cite{song2020denoising}. Flow models share tight connections with diffusion, but are rooted in ordinary differential equations \cite{lipman2022flow, song2020score}. Importantly, DDIM and flow generate deterministically once the initial noise is sampled.

\textbf{Problem Setup}. BC often augments diffusion or flow models with an extra state input $s$ and trains the diffusion or flow model $\Phi(s, w)$ with $w \sim \mathcal{N}(0, I)$ to predict actions $a \in A$ in the dataset $\mathcal D = \{(s_i, a_i)\}_{i=1}^N$. Our problem setup requires the sampling process of $\Phi$ to be deterministic once the input state and noise are given (e.g. DDIM or flow) and keep \(\Phi\) frozen throughout the fine-tuning process. Let $M$ denote the environment MDP $\Phi$ operates in, because of its deterministic generative process after initial noise, sampling noises $w$ from the noise space $W \subseteq \mathbb{R}^{\mathrm{dim}(\mathcal{A})}$ is enough to steer the reverse denoising process with $\Phi(s,w) = a$. This induces the latent noise MDP,
\begin{definition}[Latent Noise MDP \cite{wagenmaker2025steering}]
\label{def:latent_mdp}
Given the environment MDP \(M\) and a pre-trained model \(\Phi\) with weights frozen, define the latent MDP
$
\widetilde M
=
(\mathcal S, W,\{\widetilde P_h\}_{h=1}^H,\{\widetilde r_h\}_{h=1}^H,H,s_1)
$
by
\[
\widetilde P_h(s'\mid s,w):=P_h(s'\mid s,\Phi(s,w)),
\qquad
\widetilde r_h(s,w):=r_h(s,\Phi(s,w)).
\]
\end{definition}
Our goal is to find an optimal noise policy $\pi_{\mathrm{dp}}^\star: S \rightarrow \Delta_{W}$ in this induced latent MDP. Such a general problem structure encounters the noise aliasing problem stated earlier and does not leverage sample efficient structures that bring extra theoretical and empirical benefits. These issues motivate the LNL-MDP formulation and the LARBRID algorithms we will next introduce.

\section{Latent Representation Bridging for Diffusion Fine Tuning}
This section introduces the novel LNL-MDP framework by endowing a linear structure on the environment MDP $M$. The Bridge Equation solves the noise aliasing problem by aligning Q-values between the action and noise spaces together via a shared representation. Our practical algorithm use an existing representational learning algorithm to learn the shared representation for which the rest of the RL fine tune pipeline is conditioned on.

\subsection{Latent Noise Linear MDP}

When the environment MDP $M$ is linear (Definition.\ref{def:linear_mdp}), then the latent noise MDP $\Tilde{M}$ is also linear. The following proposition establishes the Latent Noise Linear MDP (LNL-MDP).
\begin{proposition}[LNL-MDP]\label{prop:latent_linearity}
    When the original MDP $M$ that $\Phi$ acts in satisfies the linear MDP definition given in \ref{def:linear_mdp}. For any $w\in W$, define the pulled back latent feature
    $\psi^\star(s, w) := \phi^\star(s, \Phi(s, w)),$
    then $\widetilde{M}$ is also a linear MDP, called the Latent Noise Linear MDP (LNL-MDP), specifically, its transition and reward function can be written as,
    $$
    \widetilde{P}(s'\mid s, w) = \langle \psi^\star(s, w), \nu(s')\rangle, \; \mathbb{E}[\widetilde{r}(s, w)] = \langle \psi^\star(s, w), \theta^r_h\rangle 
    $$
\end{proposition}
\begin{proof}
See Appendix~\ref{proof:prop_latent_linearity} for detailed proof.
\end{proof}

\textbf{Noise Aliasing Removal} By mapping state-action and state-noise pairs to the same representation space and conditioning the Q-value on this space, we can analytically solve the noise aliasing problem. Denote the Q-value in $\widetilde{M}$ as $\widetilde{Q}^{\pi_{\mathrm{dp}}}$. For any $w \sim \pi_{\mathrm{dp}}(\cdot\mid s)$, 
\begin{equation}\label{eq: shared q}
Q^\pi(s, \Phi(s, w)) = \langle \phi^\star(s, \Phi(s, w)), \zeta^\pi_h \rangle =   \langle \psi^\star(s, w), \zeta^\pi_h \rangle = \widetilde{Q}^{\pi_{\mathrm{dp}}}(s, w)
\end{equation}
Since the Q values in both $M$ and $\widetilde{M}$ are represented by their corresponding feature $\phi^\star$ and $\psi^\star$ due to \ref{eq: linear q}, but $\psi^\star$ is just a pulled back feature from $\phi^\star$, knowing $\phi^\star$ is enough to associate the noise values with the action values. Furthermore, the same Q function works for both $M$ and $\widetilde{M}$, we no longer need to refit a separate noise Q-network against the actor Q-network as done in \cite{wagenmaker2025steering}. We hence call \eqref{eq: shared q} the "Bridge" Equation and drop the superscript $\pi$ in $Q^\pi$. We emphasize that LNL-MDPs and the Bridge Equation are independent of the choice of RL fine tuning pipelines, making it adaptable to different algorithms and settings, as we show soon in Section~\ref{sec: experiments}.

\begin{algorithm}
    \caption{Latent Representation Bridging for Diffusion (LARBRID)}
    \label{alg: LARBRID}
    \begin{algorithmic}[1]
        \Require Pre-trained $\Phi$, offline data $D_{\mathrm{off}}$, total steps $T$, parameters $\theta, \zeta, \rho$ and target entropy $\Bar{H}$. 
        \State Initialize $D = \{\}$ or $D = D_{\mathrm{off}}$ and collect warm-up rollouts.
        \For {$t = 1, \dots, T$}
        \State Sample $(s, a, s', r)$ from $D$.
        \State Update $\phi_{\theta_1}$ via the score matching objective \eqref{eq: feature score match}.
        \State Update $\psi_{\theta_2}$: $\min_{\theta_2} \mathbb{E}_D\left[||\psi_{\theta_2}(s, w) - \phi_{\theta_1}(s, \Phi(s, w))||\right].$
        \State Update $Q_\zeta$ via the TD error defined in \eqref{eq: td error}.
        \State Update $\pi_{\mathrm{dp}, \rho}$: $\min_\rho \mathbb{E}_{D, w \sim \pi_{\mathrm{dp}, \rho}}[\alpha \log \pi_{\mathrm{dp}, \rho}(w\mid s)  - Q(\psi_{\theta_2}(s, w))].$
        \State Update $\alpha$: $\min \alpha \mathbb{E}_{D, w \sim \pi_{\mathrm{dp}, \rho}} [-\alpha\log\pi_{\mathrm{dp}, \rho}(w\mid s) - \alpha \Bar{H}]$.
        \If {online} 
            \State Sample noise $w \sim \pi_{\mathrm{dp}, \rho}(\cdot \mid s)$ and obtain $a = \Phi(s, w)$.
            \State Execute $a$ and obtain $s'$ and $r$.
            \State Add $D = D \cup (s, a, w, s', r)$.
        \EndIf
        \EndFor 
    \end{algorithmic}   
\end{algorithm}

\textbf{Representation Learning} LNL-MDP shifts the noise aliasing problem to a representation learning problem as in practice one needs to learn $\phi^\star$ from scratch. We choose diffusion spectral representation \cite{shribak2024diffusion} to learn $\phi^\star$, see Appendix.~\ref{appendix: diffsr}. When the transition model of the LNL-MDP is parameterized as an energy-based model (EBM), the feature $\phi^\star$ can be approximated by a fully connected neural network with 1 hidden layer, $\phi_{\theta_1}(s,a) = \sigma(W_2^\top \sigma(W_1^T\xi(s, a)))$, where $\sigma$ denotes any activation function and the Q-value can be represented as $Q_\zeta(s,a) = \langle \phi_{\theta_1}(s, a), \zeta \rangle$ because of \eqref{eq: linear q}. Here the score components $\xi$ and $\nu$ of the EBM can be learned via diffusion score matching with noise schedule $\beta$,
\begin{equation}\label{eq: feature score match}
l_{\mathrm{diff}}(\xi, \nu) =  \min_{\xi, \nu} \mathbb{E}_\beta \mathbb{E} (s, a, \Tilde{s}', s') \left [|| \Tilde{s}' + \beta\xi(s,a)^\top\nabla\nu(s') -\sqrt{1-\beta}s'|| \right]
\end{equation}

\begin{remark} There are multiple empirical works \cite{ren2022free, zhang2022making, ren2022spectral} that admit the linear MDP structure. We choose the diffusion spectral representation since it admits universal approximation property with efficient learning.
\end{remark}

\subsection{Latent Representation Bridging for Diffusion (LARBRID)}
We are now ready to introduce our practical algorithm, Latent Representation Bridging for Diffusion (LARBRID) (Algorithm \ref{alg: LARBRID}). We present LARBRID with the full generality in mind, but the algorithm can be simplified in different situations. At the beginning, we learn the feature $\phi_{\theta_1}$ from the diffusion score matching objective given in \eqref{eq: feature score match}. In the full general case when we do not have noise action recorded in the offline dataset and $\Phi$ can be a black box generator (e.g. API access) with no gradient passing through, we may need to match the noise feature $\psi_{\theta_2}(s, w)$ against the original feature $\phi_{\theta_1}(s, \Phi(s, w))$ with a mean squared error (MSE) objective.

Since we aim to find an optimal noise distribution $\pi_{\mathrm{dp}}$, we naturally parameterize $\pi_{\mathrm{dp}, \rho}$ as Gaussian policies with parameter $\rho$ and employ Soft Actor Critic (SAC) \cite{haarnoja2018soft} to solve the exploration and exploitation problem. A caveat is that due to the non-linearity created by the entropy term in SAC, the complex optimization pipelines, we parameterize $Q$ to have extra nonlinearities to saturate these factors,
$Q_\zeta(s,a) = \zeta_2^\top \sigma (\zeta_1^\top \phi_{\theta_1}(s,a)])$ where now $\zeta$ denotes the combined weights $\zeta = \{\zeta_1, \zeta_2\}$. We then use the following TD error to fit $Q_\zeta$ against the target network $\Bar{Q}_{\zeta'}$
\begin{equation}\label{eq: td error}
    \min_\zeta \mathbb{E}_{D, w'\sim\pi_{\mathrm{dp},\rho}(\cdot\mid s')}\left[(Q_\zeta(\phi_{\theta_1}(s, a)) - r - \gamma( \Bar{Q}_{\zeta'}(\phi_{\theta_1}(s', a')) - \alpha \log(\pi_{\mathrm{dp}, \rho}(w'\mid s'))))^2\right],
\end{equation}
where $\zeta'$ is the parameter updated from $\zeta$ using Polyak averaging. To compute this loss (line 6 of LARBRID), we need to sample the next action $a'$ by firstly sampling $w'\sim \pi_{\mathrm{dp},\rho}(\cdot\mid s')$ and diffuse the noise sample via $\Phi$ to obtain $a'=\Phi(s', w')$. The expectation with respect to $D$ means the data $(s,a,s',r)$ is taken from the replay buffer. The actor is then updated by maximizing $Q_\zeta$ in line 7 and we also update $\alpha$ the temperature term according to SAC. Details of these two objectives are presented in Algorithm \ref{alg: LARBRID}. The Bridge Equation is very flexible and allows us to work in any level of MDP that is the most convenient, meaning LARBRID can be simplified in different situations. 

\textbf{$\psi$-only Simplifications}. By operating only in $\widetilde{M}$, one can replace $\phi_{\theta_1}$ with $\psi_{\theta_2}$ in LARBRID. This has the benefit of skipping the feature matching MSE step in line 5 of LARBRID, which saves massive computation when sampling from the pre-trained model is expensive. However, such form can only handle online datasets in which each transition contains the augmented noise actions $w$. 

\textbf{$\phi$-only Simplifications}. Implicitly, LARBRID can work directly in the environment MDP $M$ by only learning $\phi_{\theta_1}$. This has the benefit of bypassing offline feature out-of-distribution errors. The downside is that one may need to perform backpropagation through time to optimize the pre-trained model parameters. This however can be mitigated by the modification below.



\subsection{Offline RL Fine Tuning with LARBRID-I}\label{sec: offline rl}
Offline RL has a notorious data out-of-distribution error where the next action predicted by the current policy may fall outside of the available dataset. \cite{an2021uncertainty} proposes to use multiple critic networks (>10) for learning the Q function, creating unnecessary computation burdens \cite{wagenmaker2025steering} for RL fine tuning. Instead we exploit the Bridge Equation and the $\phi$-only simplification of LARBRID. Since the Bridge Equation says the Q-values between the noise and action space are aligned, we propose to directly optimize the policy $\pi$ in the action space $A$. Indeed, our theory says fine tuning policies in the noise or the action space are simply different sides of the same coin. Specifically, given a noise policy $\pi_W \in \Delta_w$, its pushforward via the pre-trained diffusion policy is the action level policy $\pi_A = (\Phi)_\#\pi_W$. Conversely, if LARBRID-I fine tuned action policy is supported on $\text{Im}(\Phi)$ and $\Phi$ admits a right inverse $G$ such that $\Phi \circ G = 1$, then $\pi_W = (G)_\#\pi_A$, that is fine tuning at action level also induces fine tuning at the noise level. In practice, it is difficult to enforce $G$ to exist, however, this insight motivates us to modify LARBRID based on FQL \citep{fql_park2025}, an efficient offline RL algorithm that eliminates OOD error by distilling from the BC trained flow model. We call this modification LARBRID-I (`I' for implicit) and present it in Algorithm~\ref{alg:larbrid_i}. It fine tunes the BC distilled action policy from FQL using learned representation $\phi$ using Eq.~\ref{eq: feature score match}, avoids back propagation through time and brings massive empirical gains.

\section{Near-Optimal Fine Tuning Guarantees for LNL-MDPs}
We now theoretically show that if the representation $\psi^\star$ is obtained (e.g. via Eq.~\ref{eq: feature score match}), the rest of the fine-tuning regret is near-optimal. For a fixed state \(s\), write \(\Phi_s: \mathcal W \to \mathcal A\) for the map
\(
\Phi_s(w) := \Phi(s, w).
\)
For a distribution \(\mu \in \Delta \mathcal W\), let \((\Phi_s)_\#\mu \in \Delta \mathcal A\) denote its pushforward through \(\Phi_s\), i.e.,
\(
((\Phi_s)_\#\mu)(B)
:=
\mu(\{w\in\mathcal W:\Phi(s,w)\in B\})
\)
for every measurable set \(B\subseteq\mathcal A\). Equivalently, for every bounded measurable function \(f:\mathcal A\to\mathbb R\),
\(
\mathbb E_{a\sim(\Phi_s)_\#\mu}[f(a)]
=
\mathbb E_{w\sim\mu}[f(\Phi(s,w))].
\)
The theory relies on three mild assumptions listed.

\begin{assumption}\label{ass:linear_mdp}
    The MDP that $\Phi$ is trained on satisfies the linear MDP structure defined in \ref{def:linear_mdp}.
\end{assumption}

\begin{assumption}
\label{ass:deterministic_decoder}
For every state \(s\in\mathcal S\), the map \(w\mapsto \Phi(s,w)\) is deterministic.
\end{assumption}

\begin{assumption}[Optimal-policy realizability through the decoder].
\label{ass:surj_optimal}
There exists an optimal Markov policy \(\pi^\star=(\pi_h^\star)_{h=1}^H\) for the original MDP \(M\) and a Markov latent policy \(\mu^\star=(\mu_h^\star)_{h=1}^H\) for the latent MDP \(\widetilde M\) such that, for every stage \(h\in[H]\) and every state \(s\) reachable at stage \(h\) under \(\pi^\star\),
\(
\pi_h^\star(\cdot\mid s)
=
(\Phi_s)_\#\mu_h^\star(\cdot\mid s).
\)
Equivalently, for every bounded measurable \(f:\mathcal A\to\mathbb R\),
\(
\mathbb E_{a\sim\pi_h^\star(\cdot\mid s)}[f(a)]
=
\mathbb E_{w\sim\mu_h^\star(\cdot\mid s)}
[f(\Phi(s,w))].
\)
\end{assumption}
Assumption~\ref{ass:surj_optimal} only requires the reachability of optimal trajectories to ensure that the optimal policy is preserved when bridging $M$ to $\widetilde{M}$. It does not require full surjectivity onto all possible trajectories.

These assumptions and the Bridge Equation work together to ensure that the optimal value and the optimal policy are preserved as shown in the proposition below,
\begin{proposition}[Value preservation]\label{prop:value-preservation}
Suppose Assumption~\ref{ass:deterministic_decoder} and \ref{ass:surj_optimal} hold, then the optimal values of the original MDP and the latent MDP are equivalent:
\(
V_0^\star=
\widetilde V_0^\star
.
\)
\end{proposition}
\begin{proof}
See Appendix~\ref{proof:prop_value_preservation} for detailed proof
\end{proof}

The central insight to our theory is to view the pre-trained frozen $\Phi$ as a virtual offline dataset. 
\begin{definition}[Warm-up rollout dataset]
\label{def:warmup_dataset_copy}
Execute $\Phi$ for \(N_{\mathrm{off}}\) episodes in the real environment. In episode \(i\) and step \(h\), sample $w_h^i\sim \mathcal N(0,I)$ and $a_h^i=\Phi(s_h^i, w_h^i),$, then observe reward \(r_h^i\) and next state \(s_{h+1}^i\) to collect,
\(
D_{\mathrm{off}}
=
\{(s_h^i,w_h^i,r_h^i,s_{h+1}^i)\}_{h\in[H],\,i\in[N_{\mathrm{off}}]}.
\)
The empirical covariate generated by $D_{\mathrm{off}}$ is defined as,
\(
\Lambda_{\mathrm{off}}^h
:=
\sum_{i=1}^{N_{\mathrm{off}}}
\psi^\star(s_h^i,w_h^i)\psi^\star(s_h^i,w_h^i)^\top.
\)
\end{definition}
We show in Proposition~\ref{prop:virtual_offline} that above definition indeed provides the correct offline dataset according to the FineTuneRL framework \cite{wagenmaker2023leveraging}.
\begin{proposition}[Validity of $D_{\mathrm{off}}$]
\label{prop:virtual_offline}
Under Assumptions~\ref{ass:linear_mdp} and~\ref{ass:deterministic_decoder}, the warm-up dataset \(D_{\mathrm{off}}\) is a valid offline dataset on the latent MDP \(\widetilde M\). More precisely: \(D_{\mathrm{off}}\) consists of \(N_{\mathrm{off}}\) i.i.d.\ trajectories of \(\widetilde M\) generated by the policy \(\pi_{\mathrm{dp}}\);
\end{proposition}

\begin{proof}
See Appendix~\ref{proof:prop_virtual_offline} for detailed proof
\end{proof}

Denote the expected covariate generated by playing policy $\pi_{\mathrm{dp}}$ as 
$\Lambda^h := \mathbb E_{\pi_{\mathrm{dp}}}\big[\psi^\star(s_h,w)\psi^\star(s_h,w)^\top\big]$. The pre-trained model when rolled out approximate this covariate and serves as a valid prior as described in the following proposition.
\begin{definition}[Offline-to-online Concentrability Coefficient \cite{wagenmaker2023leveraging}]
\label{def:concentrability-copy}
For the latent MDP \(\widetilde M\) with offline covariance \(\Lambda_{\mathrm{off}}^h\), define the offline-to-online concentrability coefficient, with some $\epsilon > 0$,
\[
C_{o2o}^h(\Lambda_{\mathrm{off}}^h,\varepsilon,T)
:=
\inf_{\pi_{\exp}}
\max_{\pi\in\Pi}
\frac{
\|\psi_{\pi,h}^\star\|^2_{(T\Lambda_{\pi_{\exp}}^h+\Lambda_{\mathrm{off}}^h)^{-1}}
}{
(\widetilde V_0^\star-\widetilde V_0^\pi)^2 \vee \varepsilon^2
},
\]
where
\(
\psi_{\pi,h}^\star := \mathbb E_\pi[\psi^\star(s_h,w_h)],
\)
\(\Lambda_{\pi_{\exp}}^h\) is the stage-\(h\) exploration covariance.
\end{definition}
In the LNL-MDP context, this coefficient additionally measures the quality of $\Phi$. Intuitively, a good $\Phi$ should contain sufficient amount of information such that minimum amount of online interactions is needed to find an optimal policy. The previous steps complete the reduction: the frozen diffusion decoder induces the LNL-MDP, and warm-up rollouts on the diffusion decoder produce a valid offline dataset on the LNL-MDP. We can therefore apply FineTuneRL \cite{wagenmaker2023leveraging} directly to obtain the following bound that depends on the warm up rollout dataset.
\begin{theorem}[Non-asymptotic Guarantee]
\label{thm:main_theory}
Under Assumptions~\ref{ass:linear_mdp}, \ref{ass:deterministic_decoder}, and \ref{ass:surj_optimal}, suppose that we first collect \(D_{\mathrm{off}}\) via Definition~\ref{def:warmup_dataset_copy}, there is an algorithm (FTPedel \cite{wagenmaker2023leveraging}) such that the number of additional online episodes needed to obtain an \(\varepsilon\)-optimal policy satisfies
\[
T_{\mathrm{on}}(\varepsilon)
=
\widetilde O\!\left(
\max_{h\in[H]}
\min\left\{
T:\,
C_{o2o}^h(\Lambda_{\mathrm{off}}^h,\varepsilon,T)\le \frac{1}{\bar{\beta}}
\right\}
\cdot H
\right),
\]
where \(\bar{\beta}=\widetilde\Theta(d^2H^2)\). Moreover, any algorithm in this setting requires at least
\[
T_{\mathrm{on}}(\varepsilon)
=
\widetilde \Omega\!\left(
\max_{h\in[H]}
\min\left\{
T:\,
C_{o2o}^h(\Lambda_{\mathrm{off}}^h,\varepsilon,T)\le \frac{1}{\bar{\beta}}
\right\}
\cdot \frac{1}{H}
\right)
\]
online episodes to return an \(\varepsilon\)-optimal policy with constant probability.
\end{theorem}
\begin{proof}
    See Appendix~\ref{proof:main_theory} for detailed proof
\end{proof}
\vspace{-3mm}
Theorem \ref{thm:main_theory} proves that latent diffusion-policy fine-tuning is minimax-optimal in the LNL-MDP setting up to $H$ factors. The quality of $\Phi$ is translated into the quality of the offline data it can collect. Extending the result to incorporate feature learning and SAC style optimization is beyond the scope of this work, we expanded on this in Appendix.~\ref{appendix:theory-to-real-gap}. The main point here is that once $\psi$ is known, the fine tuning problem is theoretically easy to solve. Indeed, we next show that empirically this is also true.


\section{Experiments}\label{sec: experiments}
We conduct thorough experiments to understand the capability of LARBRID through online (Section \ref{sec: online fine tune}), offline (Section \ref{sec: offline fine tune}) and offline-to-online fine tuning settings (Section \ref{sec: o2o fine tune}) across diverse benchmarks. We compare LARBRID against multiple state-of-the-art baselines. Further experimental details can be found in Appendix~\ref{appendix: exp-details}.

\subsection{Online Fine Tuning}\label{sec: online fine tune}
\begin{figure}[htbp]
     \centering
     \begin{subfigure}[b]{0.35\textwidth}
         \centering
         \includegraphics[width=\textwidth, keepaspectratio]{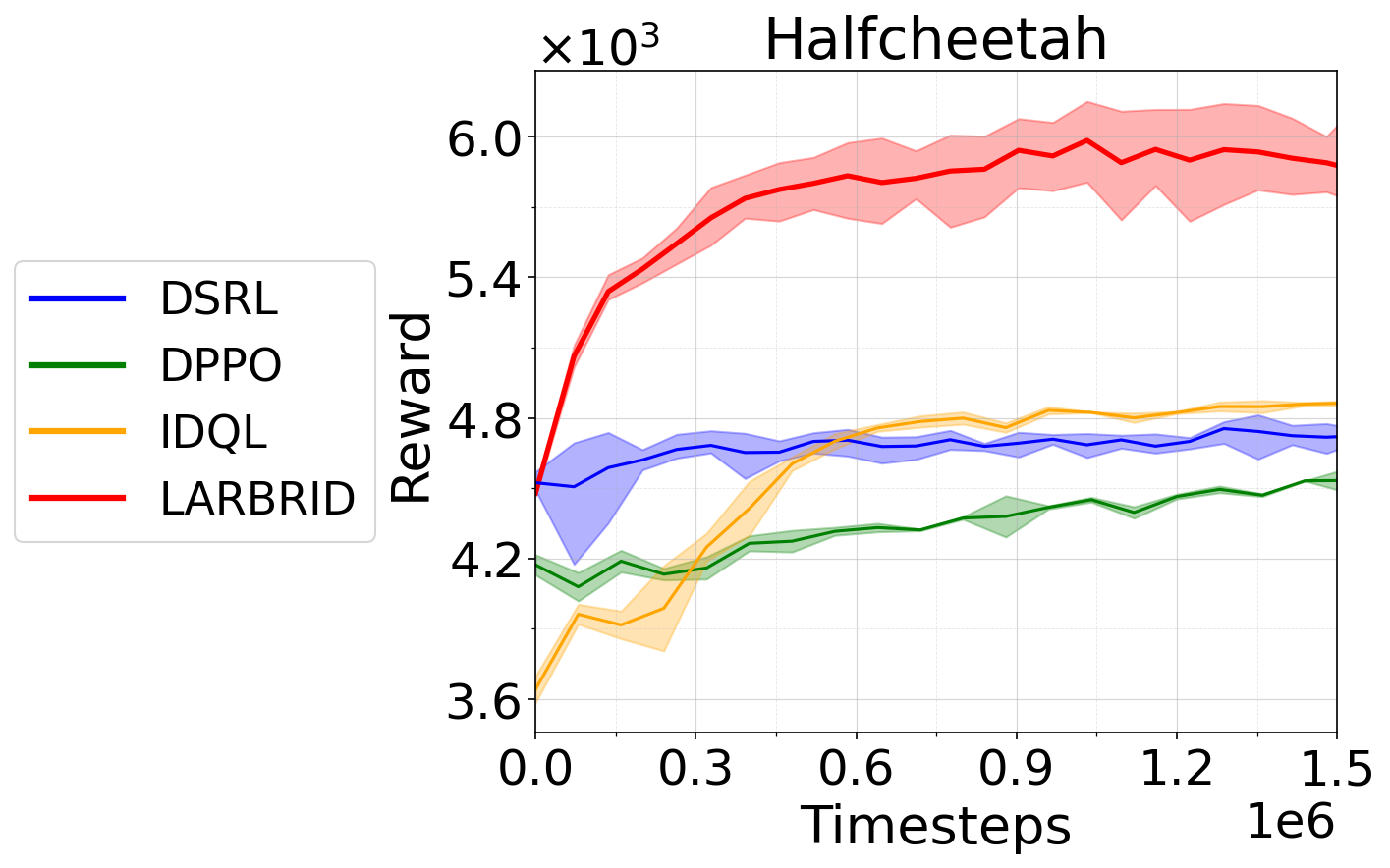}
     \end{subfigure}
     \hfill
     \begin{subfigure}[b]{0.28\textwidth}
         \centering
         \includegraphics[height=3cm, width=\textwidth]{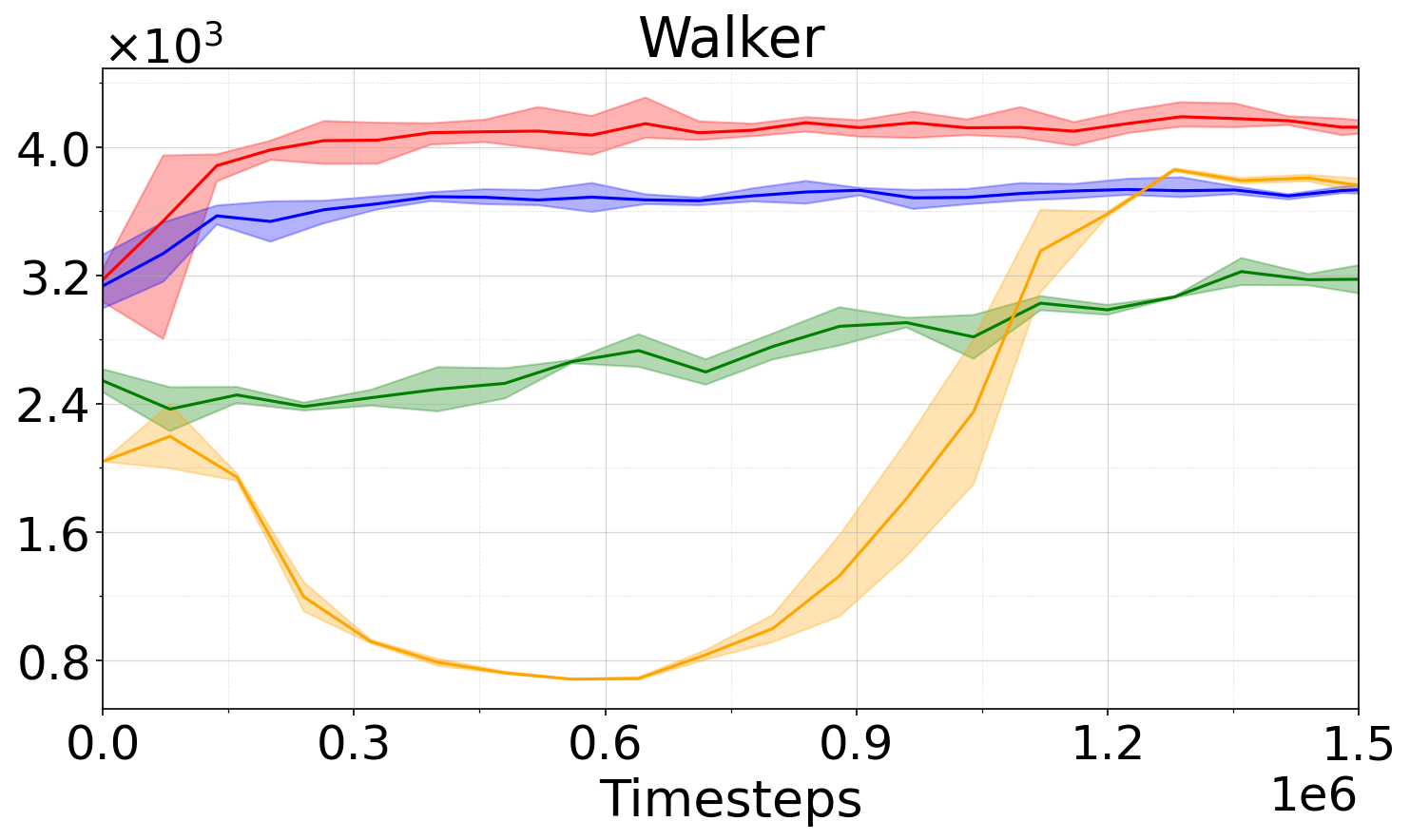}
     \end{subfigure}
     \hfill
     \begin{subfigure}[b]{0.28\textwidth}
         \centering
         \includegraphics[height=3cm, width=\textwidth]{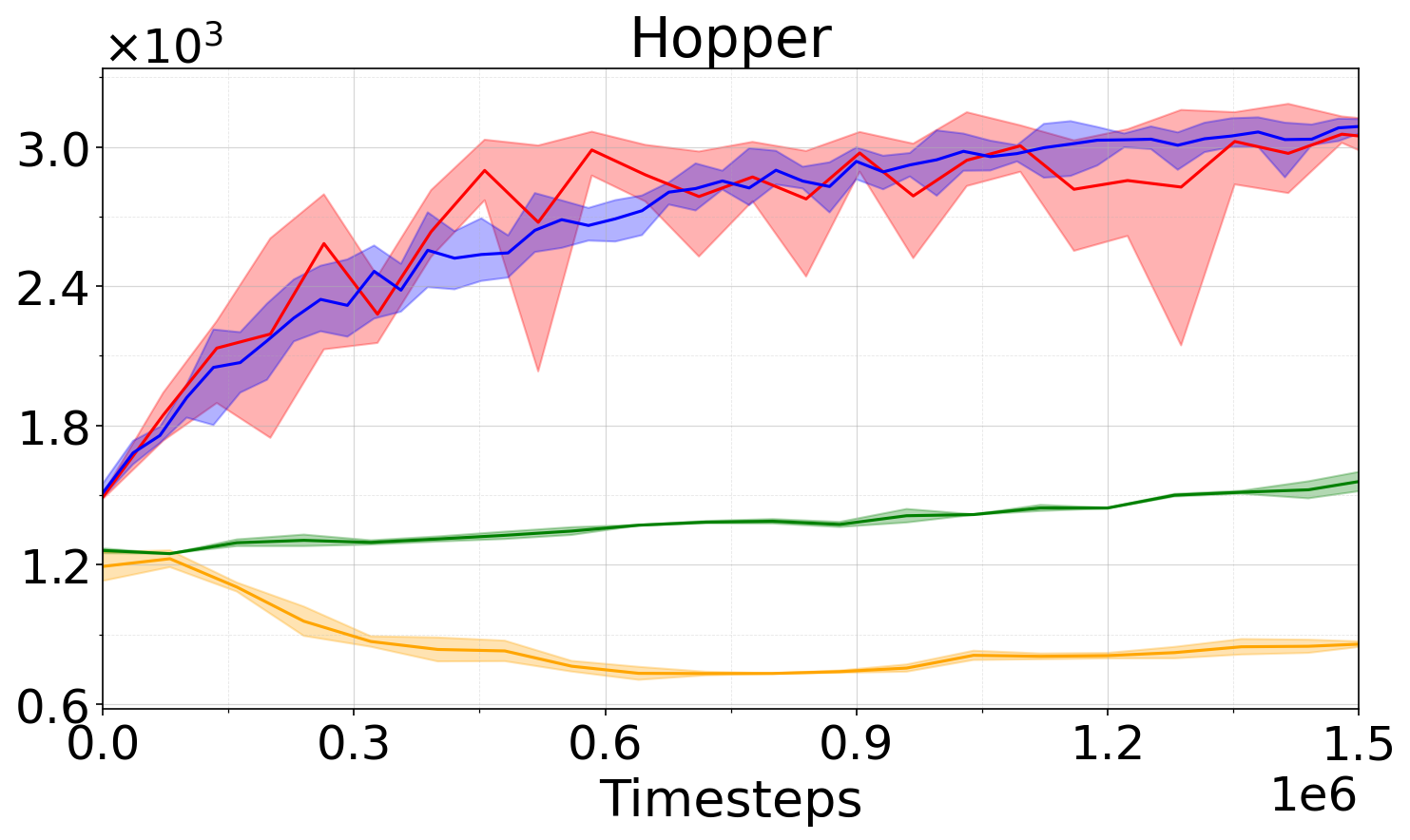}
     \end{subfigure}
        \caption{Online fine tuning results of LARBRID on Gymnasium \cite{towers2024gymnasium} control tasks.}
        \label{fig:gym-results}
\end{figure}

\begin{figure}[htbp]
     \centering
     \begin{subfigure}[b]{0.24\textwidth}
         \centering
         \includegraphics[width=\textwidth]{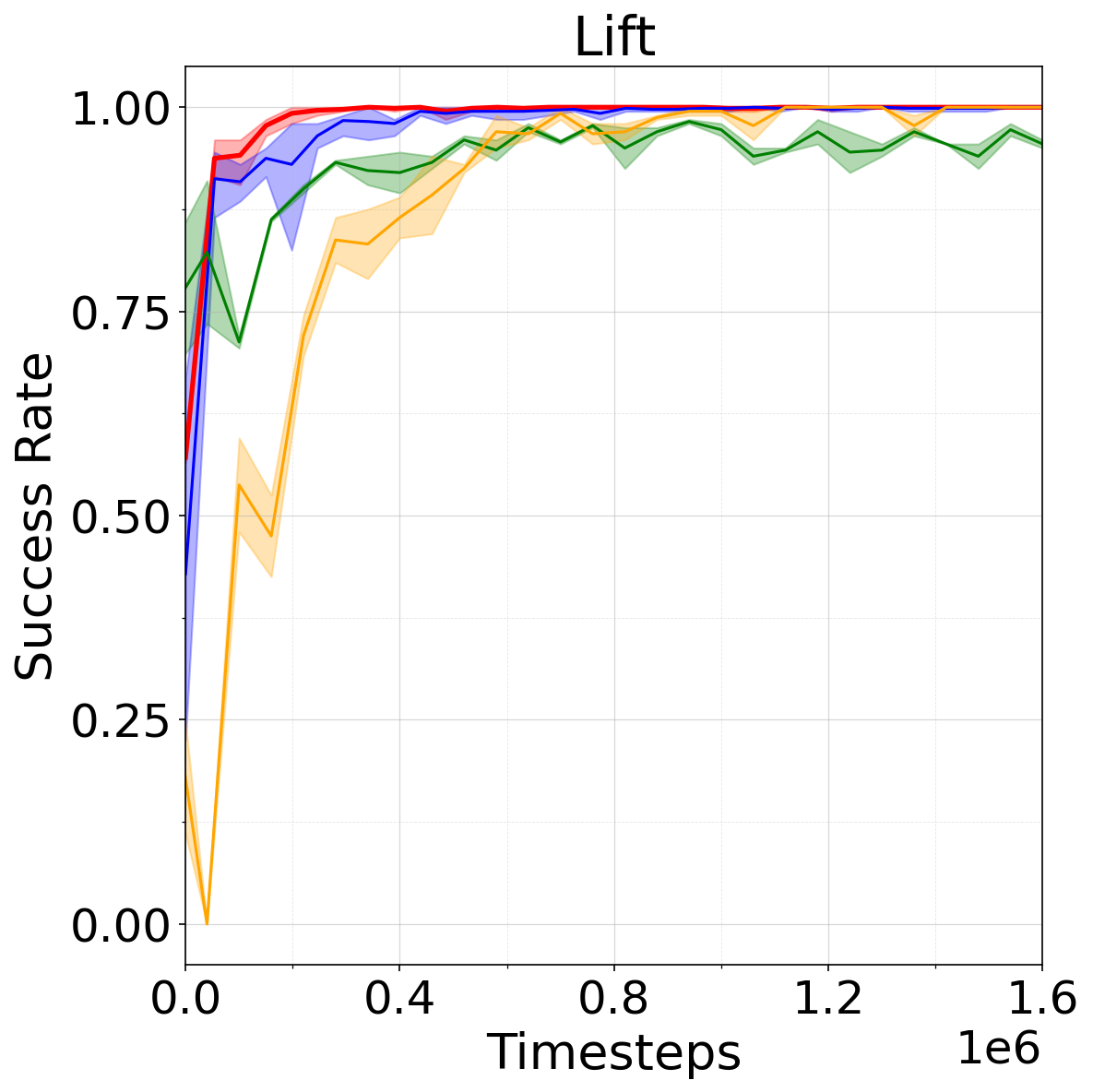}
     \end{subfigure}
     \hfill
     \begin{subfigure}[b]{0.24\textwidth}
         \centering
         \includegraphics[width=\textwidth]{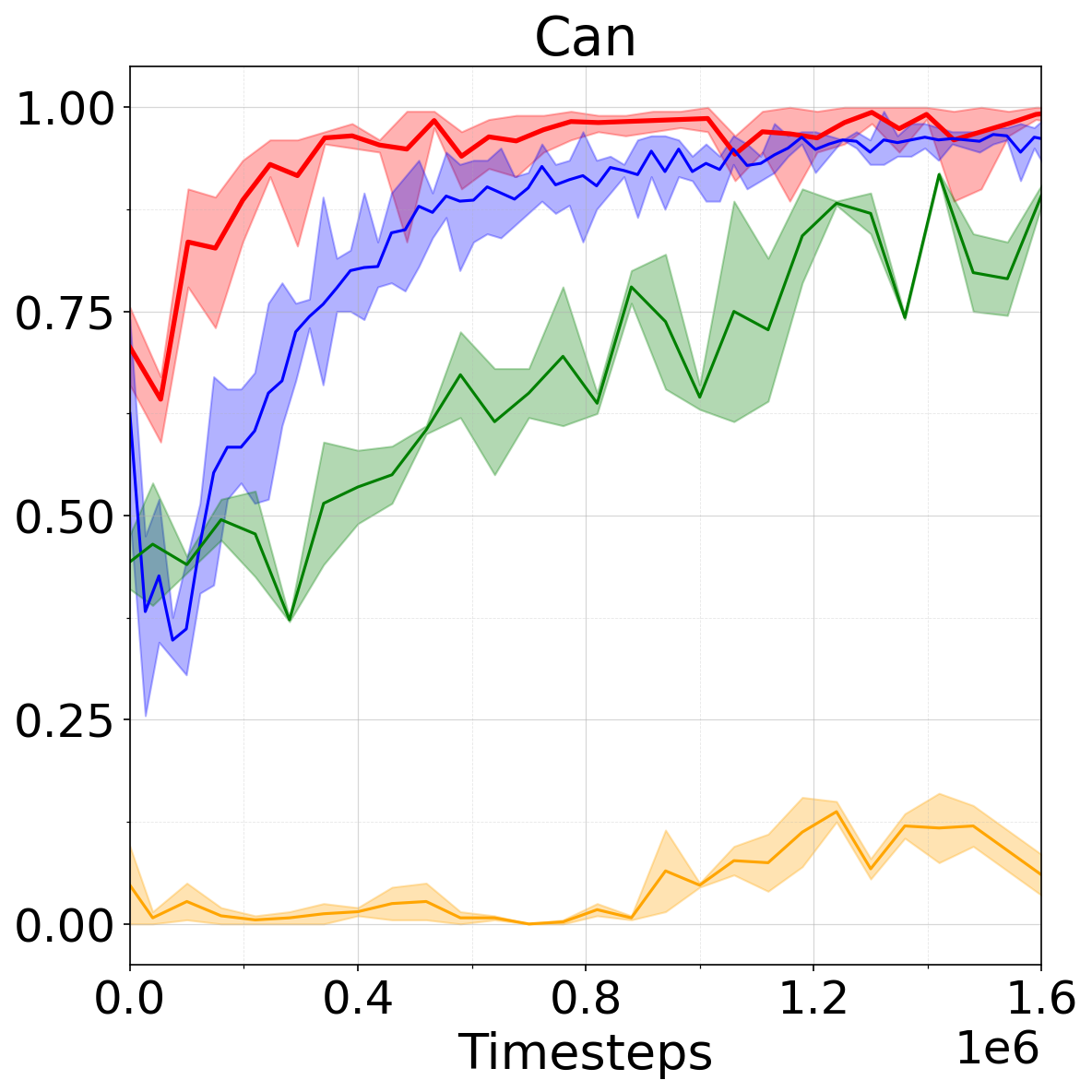}
     \end{subfigure}
     \hfill
     \begin{subfigure}[b]{0.24\textwidth}
         \centering
         \includegraphics[width=\textwidth]{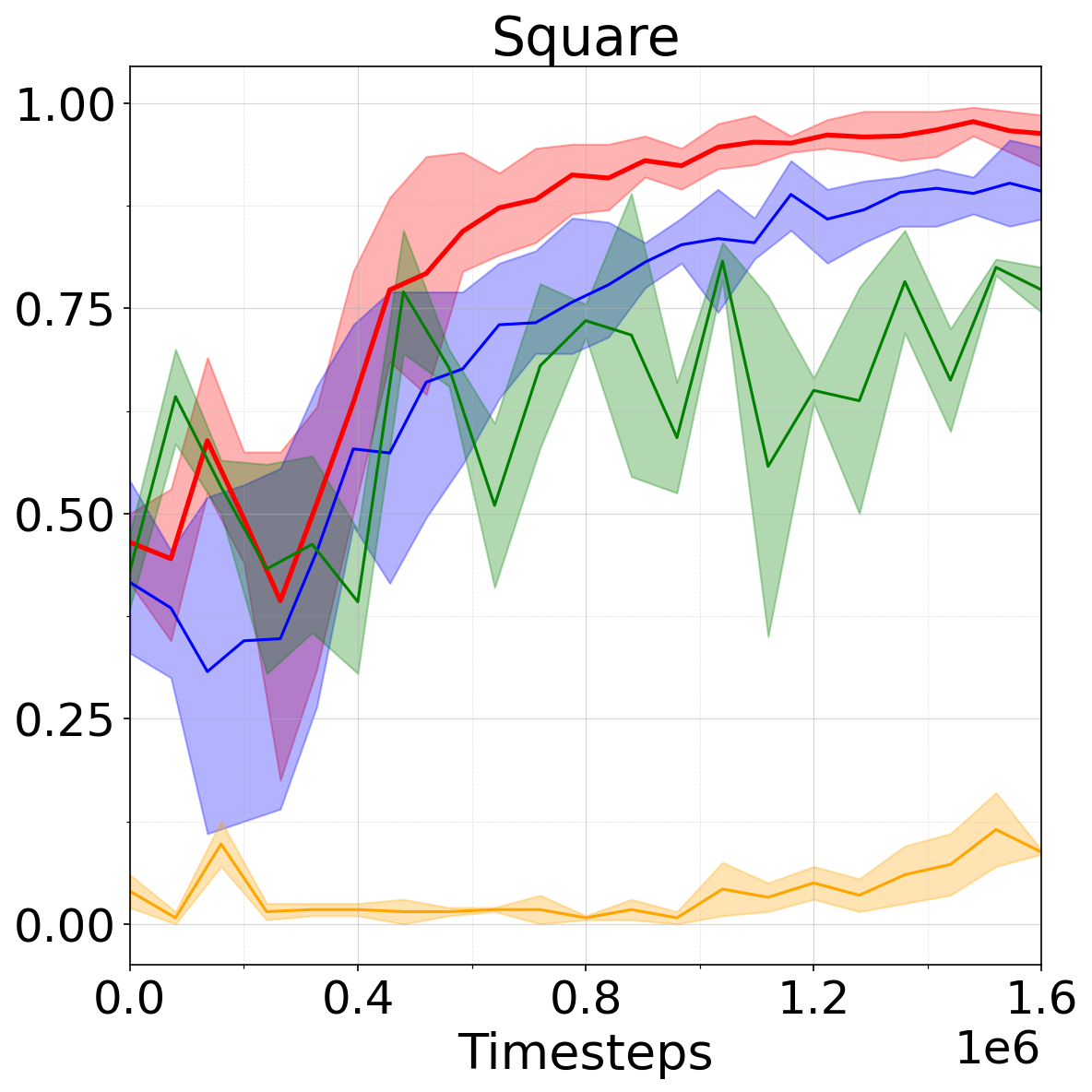}
     \end{subfigure}
     \hfill
     \begin{subfigure}[b]{0.24\textwidth}
         \centering
         \includegraphics[width=\textwidth]{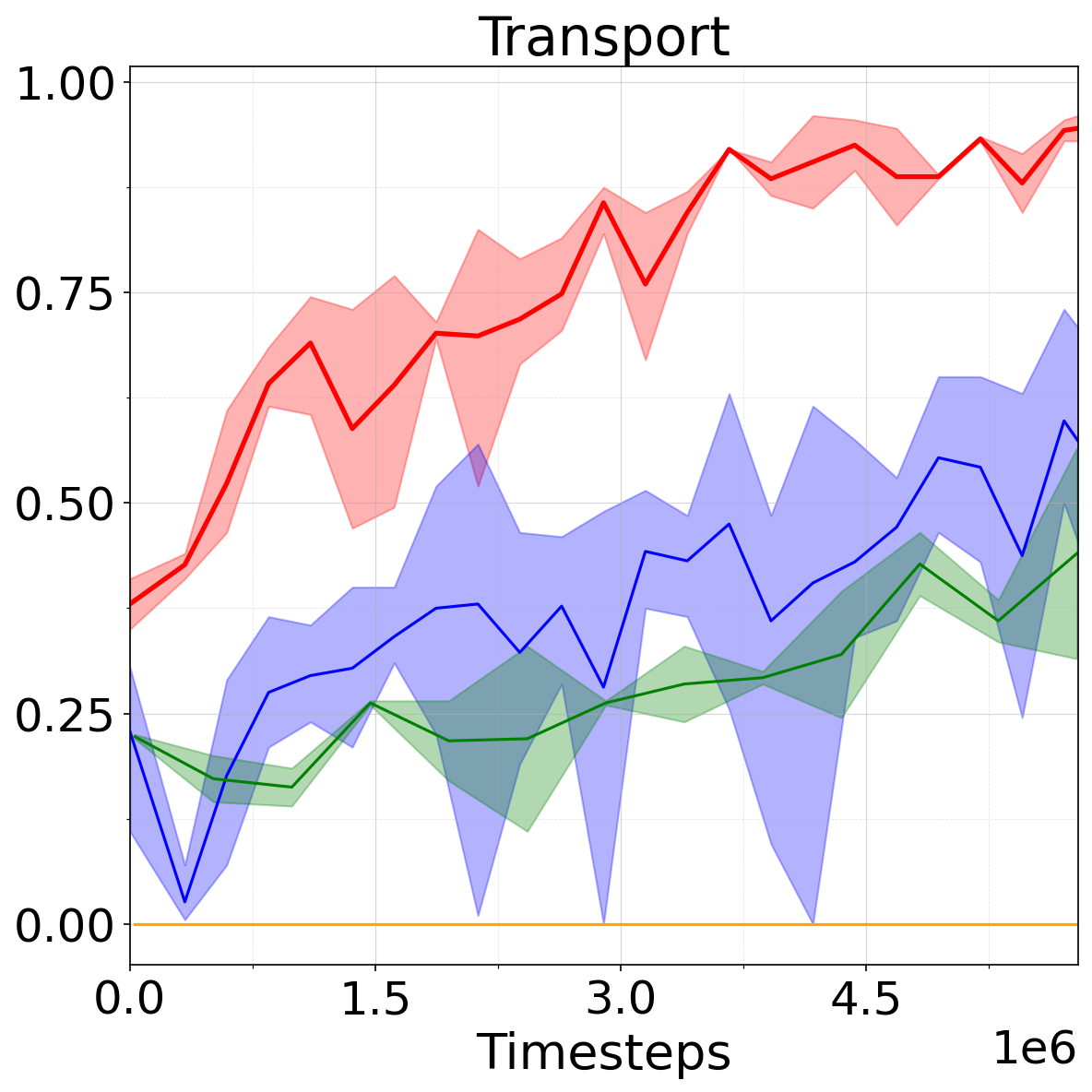}
     \end{subfigure}
    \caption{Online fine tuning results of LARBRID on Robomimic \cite{robomimic2021} control tasks.}
    \label{fig:robomimic-results}
\end{figure}

We run online fine tuning of pre-trained DDIM experiments on two standard benchmarks, Gymnasium \cite{towers2024gymnasium} and Robomimic \cite{robomimic2021}. We compare our method against three other recent competitive baselines, DSRL \cite{wagenmaker2025steering}, DPPO \cite{dppo2024} and IDQL \cite{hansenestruch2023idql}. Some earlier baselines such as DIPO \cite{yang2023policy} and QSM \cite{psenka2023learning} are dropped, because they aim to adapt from scratch and have been shown to be less competitive than the latest baselines. We set all methods to use the same DDIM checkpoints provided by \cite{dppo2024} and use the same hyperparameters and public code released by these methods. LARBRID can be easily implemented on top of DSRL by conditioning DSRL on the representation space $\mathbb{R}^d$ of the LNL-MDP. DSRL runs their full algorithm for all experiments except for Transport task due to heavy DDIM sampling. For fair comparisons, we also run LARBRID in full and only apply $\psi$-only simplification to the Transport task. The results are averaged over 4 random seeds, with their mean and standard errors displayed in Figure~\ref{fig:gym-results} and Figure~\ref{fig:robomimic-results}.

Since LARBRID learns good representations fast (see Figure~\ref{fig:robomimic-diffusion-losses}), which resolve the noise aliasing problem for free, it is able to break the sub-optimality of the pre-trained model and discover better noises to sample conditioned on the learned representation. This is shown across both the Gymnasium and Robomimic tasks. Notably for all tasks, LARBRID converges very fast with training variance at least on par with the baselines, demonstrating its sample efficiency and stability. In particular, LARBRID is able to achieve at least 21\% more reward in Halfcheetah, and $\sim$70\% better performance than DSRL in the Transport task, the hardest problem considered in this setting.  

\subsection{Offline Fine Tuning}\label{sec: offline fine tune}

OOD errors in offline setting is broadly discussed in Section~\ref{sec: offline rl}. DSRL handles OOD issues by deploying ensembles of 10 separate Q-networks, causing heavy computational cost. Instead of resolving to offline RL tricks, we deploy LARBRID-I by directly working with the environment MDP $M$, bypassing the OOD errors and computational burdens of large critic ensembles. We run LARBRID-I on OGBench \cite{ogbench_park2025} across 10 locomotion and manipulation tasks with 4 random seeds and compare against recent offline RL algorithms, including IQL \cite{kostrikov2021offline}, REBRAC \cite{tarasov2023revisiting}, IDQL \cite{hansenestruch2023idql}, SRPO \cite{chen2023score}, CAC \cite{ding2023consistency}, FAWAC \cite{nair2020awac}, FBRAC \cite{wang2022diffusion}, IFQL \cite{hansenestruch2023idql}, FQL \cite{fql_park2025} and DSRL \cite{wagenmaker2025steering}. We also include 5 manipulation tasks with visual observations and report the average performance (see Appendix~\ref{appendix: offline} for individual results). Due to high computational costs, we select 6 strong performing baselines to compare against ours for the visual input tasks. All tasks are set to their default single task variant in OGBench and each method is trained with 1M steps (500K for visual tasks).  The results are displayed in Table~\ref{tab:offline-ogbench} and all baseline results are copied from \cite{wagenmaker2025steering}. As also observed in the online fine tuning setting, the representation power provided by LNL-MDP allows LARBRID-I to achieve high success rates across 7 out of 10 tasks. It runs particularly well on the difficult humanoid maze problem and brings at least $1.5\times$ performance increase for the medium maze and $3\times$ performance increase in the large maze. LARBRID-I also handles visual inputs well and brings an extra 7\% improvement on top of the second strongest baseline FQL.

\begin{table*}[htbp]
\centering
\caption[Offline LARBRID-I task success results on OGBench]{Offline LARBRID-I task
success results on OGBench~\cite{ogbench_park2025}.\\
\footnotesize{hum: humanoidmaze, med: medium, s: single, d: double, pz: puzzle.\\
visual: visual manipulation tasks, averaged over 5 visual tasks.}}
\label{tab:offline-ogbench}
\small
\setlength{\tabcolsep}{2.pt}
\renewcommand{\arraystretch}{1.1}
\begin{tabular}{@{}l*{11}{c}@{}}
\hline\hline
Method & ant-large & ant-giant & hum-med & hum-large & soccer
& cube-s & cube-d & scene & pz-3x3 & pz-4x4 & visual \\
\hline\hline
IQL    & \vv{48}{9}  & \vv{0}{0}  & \vv{32}{7}  & \vv{3}{1}   & \vv{3}{2}  & \vv{85}{8}  & \vv{1}{1}  & \vv{12}{3}  & \vv{2}{1} & \vv{5}{2}  & \vv{42}{4} \\
REBRAC & \vv{91}{10} & \vb{27}{22}& \vv{16}{9}  & \vv{2}{1}   & \vv{0}{0}  & \vv{92}{4}  & \vv{7}{3}  & \vv{50}{13} & \vv{2}{1} & \vv{10}{3} & \vv{60}{2} \\
IDQL   & \vv{0}{0}   & \vv{0}{0}  & \vv{1}{1}   & \vv{0}{0}   & \vv{0}{1}  & \vb{96}{2}  & \vv{16}{10}& \vv{33}{14} & \vv{0}{0} & \vv{26}{6} & $-$ \\
SRPO   & \vv{0}{0}   & \vv{0}{0}  & \vv{0}{0}   & \vv{0}{0}   & \vv{0}{0}  & \vv{82}{16} & \vv{0}{0}  & \vv{2}{2}   & \vv{0}{0} & \vv{7}{4}  & $-$ \\
CAC    & \vv{42}{7}  & \vv{0}{0}  & \vv{38}{19} & \vv{1}{1}   & \vv{0}{0}  & \vv{80}{30} & \vv{2}{2}  & \vv{50}{40} & \vv{0}{0} & \vv{1}{1}  & $-$ \\
FAWAC  & \vv{1}{1}   & \vv{0}{0}  & \vv{6}{2}   & \vv{0}{0}   & \vv{12}{13}& \vv{81}{9}  & \vv{2}{1}  & \vv{18}{8}  & \vv{1}{1} & \vv{0}{0}  & $-$ \\
FBRAC  & \vv{70}{20} & \vv{0}{1}  & \vv{25}{8}  & \vv{0}{1}   & \vv{24}{4} & \vv{83}{13} & \vv{22}{12}& \vv{46}{10} & \vv{2}{2} & \vv{5}{1}  & \vv{22}{2} \\
IFQL   & \vv{24}{17} & \vv{0}{0}  & \vv{69}{19} & \vv{6}{2}   & \vv{16}{9} & \vv{73}{3}  & \vv{9}{5}  & \vv{0}{0}   & \vv{0}{0} & \vv{21}{11}& \vv{50}{5} \\
FQL    & \vv{80}{8}  & \vv{4}{5}  & \vv{19}{12} & \vv{7}{6}   & \vb{39}{6} & \vb{97}{2}  & \vv{36}{6} & \vv{76}{6}  & \vb{16}{5}& \vv{11}{3} & \vv{65}{2} \\
DSRL   & \vv{40}{29} & \vv{0}{0}  & \vv{34}{20} & \vv{10}{12} & \vv{28}{9} & \vb{93}{14} & \vb{53}{14}& \vb{88}{9}  & \vv{0}{0} & \vb{37}{13}& $-$ \\
\hdashline
\textbf{LARBRID-I}
       & \vb{95}{0}  & \vv{0}{0}  & \vb{96}{1}  & \vb{35}{15} & \vb{36}{9} & \vb{93}{5}  & \vb{51}{2} & \vb{89}{5}  & \vv{9}{2} & \vv{16}{1} & \vb{70}{3} \\
\hline
\end{tabular}
\vspace{-3mm}
\end{table*}

\subsection{Offline to Online Fine Tuning}\label{sec: o2o fine tune}
We also test the capabilities of both LARBRID and LARBRID-I to handle offline-to-online fine tuning tasks from OGBench\cite{ogbench_park2025}. To obtain the offline pre-trained policy for LARBRID, we pre-train the default BC model in FQL \cite{fql_park2025} for 1M offline steps and fine tune it with LARBRID online for another 1M steps. For LARBRID-I we continue its training after 1M offline steps with another 1M online steps. We add two specific offline to online baselines, RLPD \cite{ball2023efficient} and Cal-QL \cite{nakamoto2023cal}. The results are displayed in Table\ref{tab:o2o_results} with other baseline results copied from \cite{fql_park2025}. See Figure~\ref{fig:ogbench-larbrid-i-o2o} and \ref{fig:ogbench-larbrid-o2o} for additional training plots. Overall, both models combined together are able to solve all benchmark problems displayed here. LARBRID is able to learn the representative features from scratch and steer the poor performing pre-trained BC model to achieve remarkably high success across multiple tasks. It failed to solve humanoid maze, likely due to the combined factors of the BC model not covering the near-optimal trajectories well and the learned feature being not representative enough. LARBRID-I on the other hand is able to find policies to solve all tasks except Puzzle-4x4 which LARBRID performs better in comparison. Empirically, this gap between LARBRID and LARBRID-I is expected. Since $\psi^*$ and $\pi$ are learned in practice, training them in the noise space or the action space may incur different optimization landscapes, causing the results to differ especially in hard tasks like humanoid and ant soccer. The different results show the flexibility of the LNL-MDP framework, as if one space is too hard to solve in practice, one can work with the other space.

\begin{table*}[htbp]
\centering
\caption[Offline-to-online fine-tuning results of LARBRID on OGBench]{Offline-to-online
fine-tuning results of LARBRID on OGBench~\cite{ogbench_park2025}.\\
\footnotesize{Left number denotes offline fine tuned success rate, right number denotes online fine tuned success rate}}
\label{tab:o2o_results}
\small
\setlength{\tabcolsep}{3.5pt}
\renewcommand{\arraystretch}{1.3}
\begin{tabular}{@{}l*{5}{c}@{}}
\hline\hline
Method & cube-double & scene & puzzle-4x4 & hum-med & ant soccer \\
\hline\hline
IQL    & \oo{0}{1}{0}{0}    & \oo{14}{11}{10}{9}  & \oo{5}{2}{1}{1}    & \oo{21}{13}{16}{8}  & \oo{2}{1}{0}{0} \\
REBRAC & \oo{6}{5}{28}{28}  & \oo{55}{10}{100}{0} & \oo{8}{4}{14}{35}  & \oo{16}{20}{1}{1}   & \oo{0}{0}{0}{0} \\
Cal-QL & \oo{0}{0}{0}{0}    & \oo{1}{2}{50}{53}   & \oo{0}{0}{0}{0}    & \oo{0}{0}{0}{0}     & \oo{0}{0}{0}{0} \\
RLPD   & \oo{0}{0}{0}{0}    & \ob{0}{0}{100}{0}   & \ob{0}{0}{100}{1}  & \oo{0}{0}{0}{8}     & \oo{0}{0}{0}{0} \\
IFQL   & \oo{12}{9}{40}{5}  & \oo{0}{1}{60}{39}   & \oo{23}{6}{19}{33} & \oo{56}{35}{82}{20} & \oo{26}{15}{39}{10} \\
FQL    & \oo{40}{11}{92}{3} & \ob{82}{11}{100}{1} & \oo{8}{3}{38}{52}  & \oo{12}{7}{22}{12}  & \oo{28}{8}{86}{5} \\
\hdashline
\textbf{LARBRID}
       & \ob{0}{0}{99}{1}   & \ob{0}{0}{100}{0}   & \ob{0}{0}{100}{0}  & \oo{0}{0}{0}{0}     & \oo{0}{0}{0}{0} \\
\textbf{LARBRID-I}
       & \ob{51}{2}{98}{2}  & \ob{88}{5}{100}{0}  & \oo{16}{1}{68}{47} & \ob{96}{1}{100}{0}  & \ob{36}{9}{99}{1} \\
\hline
\end{tabular}
\vspace{-3mm}
\end{table*}

\subsection{Fine Tuning Pre-trained Robotic Foundation Models}
\begin{figure}[htbp]
 \vspace{-2mm}
     \centering
     \begin{subfigure}[b]{0.23\textwidth}
         \centering
         \subcaption{\scriptsize Aloha Cube Transfer}
         \includegraphics[width=\textwidth]{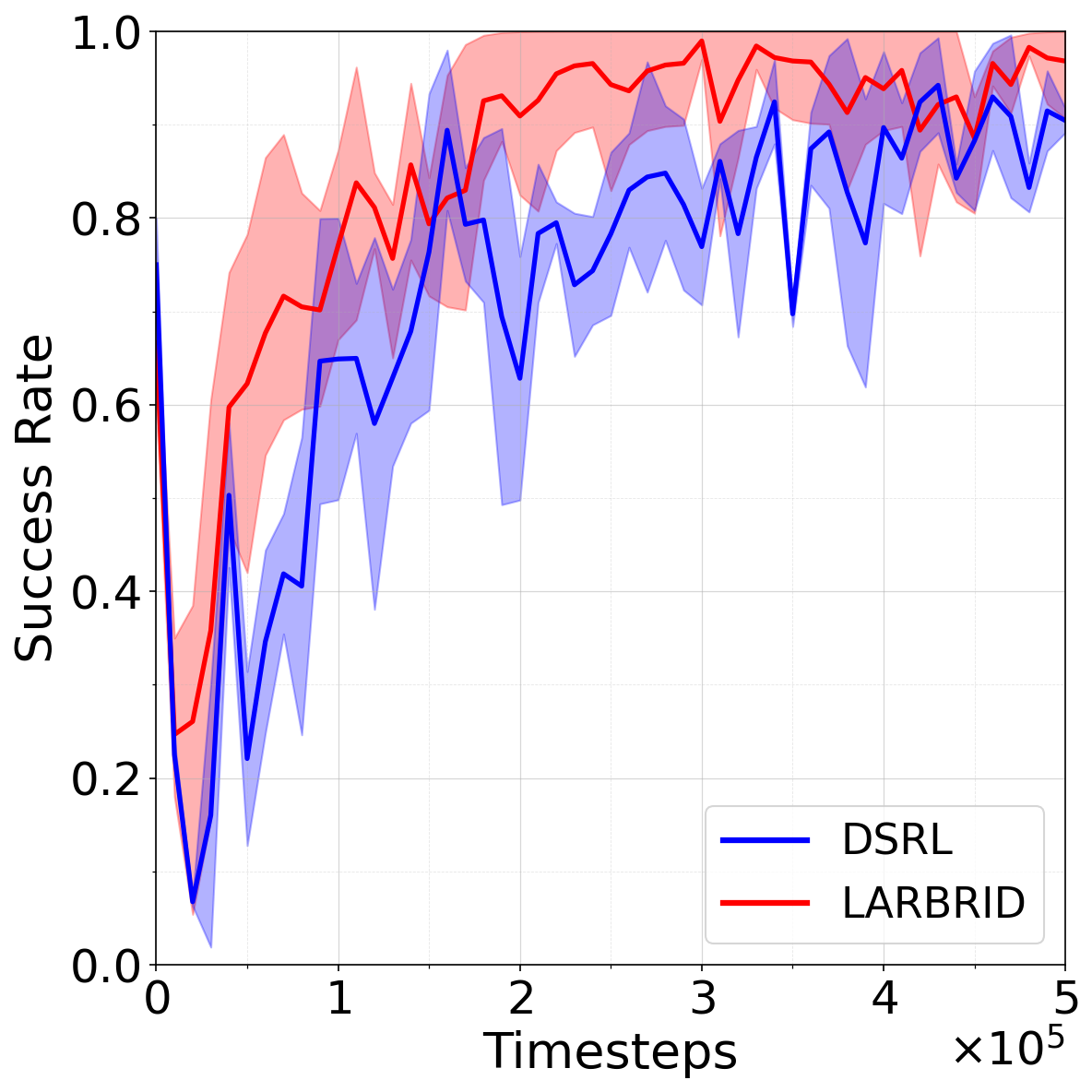}
     \end{subfigure}
     \hfill
     \begin{subfigure}[b]{0.23\textwidth}
         \centering
         \subcaption{\scriptsize Libero Open Drawer}
         \includegraphics[width=\textwidth]{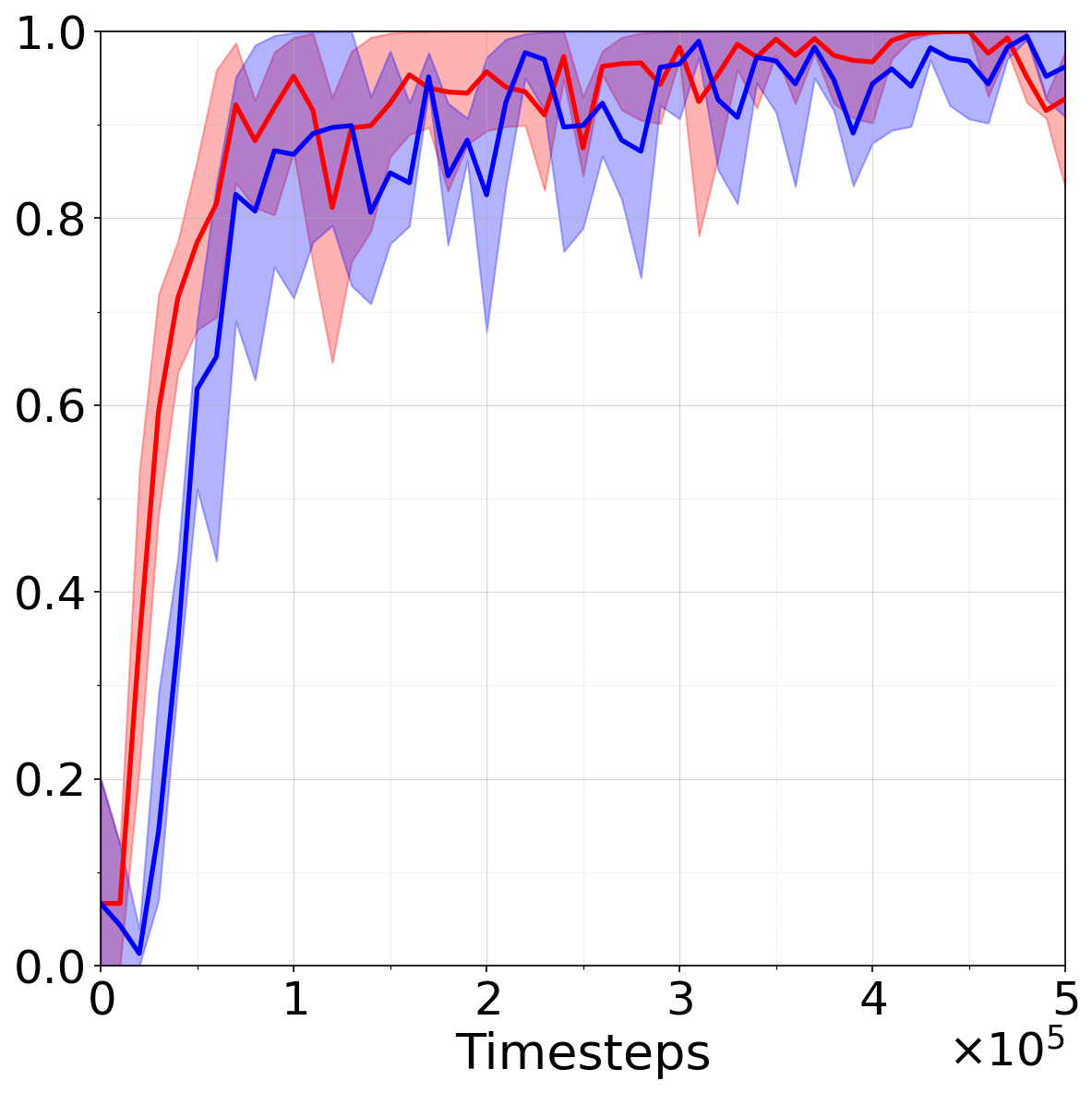}
     \end{subfigure}
     \hfill
     \begin{subfigure}[b]{0.23\textwidth}
         \centering
         \subcaption{\scriptsize Libero Turn On Stove}
         \includegraphics[width=\textwidth]{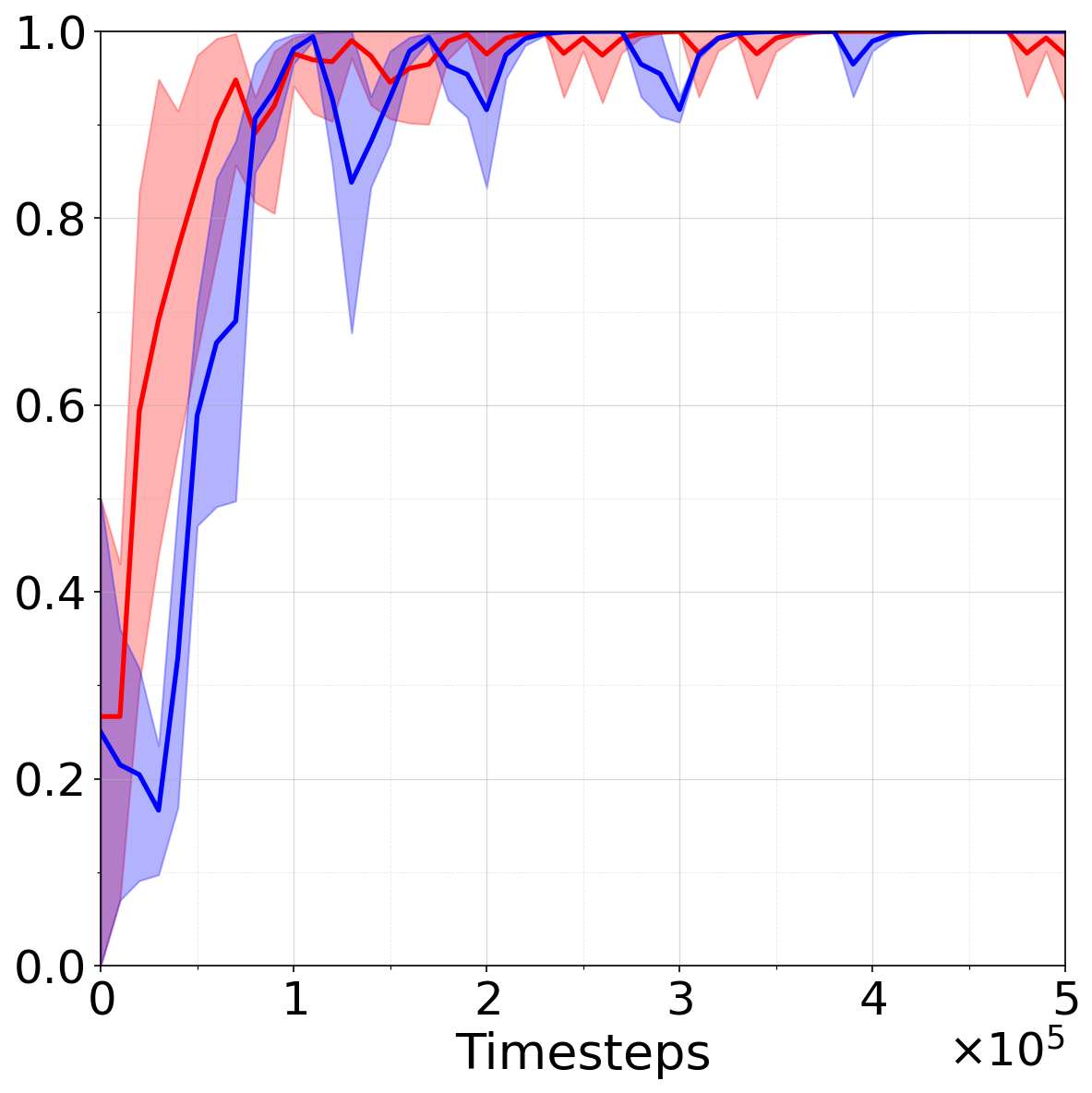}
     \end{subfigure}
     \hfill
     \begin{subfigure}[b]{0.24\textwidth}
         \centering
         \subcaption{\scriptsize SC Ablation -- Halfcheetah}
         \includegraphics[width=\textwidth]{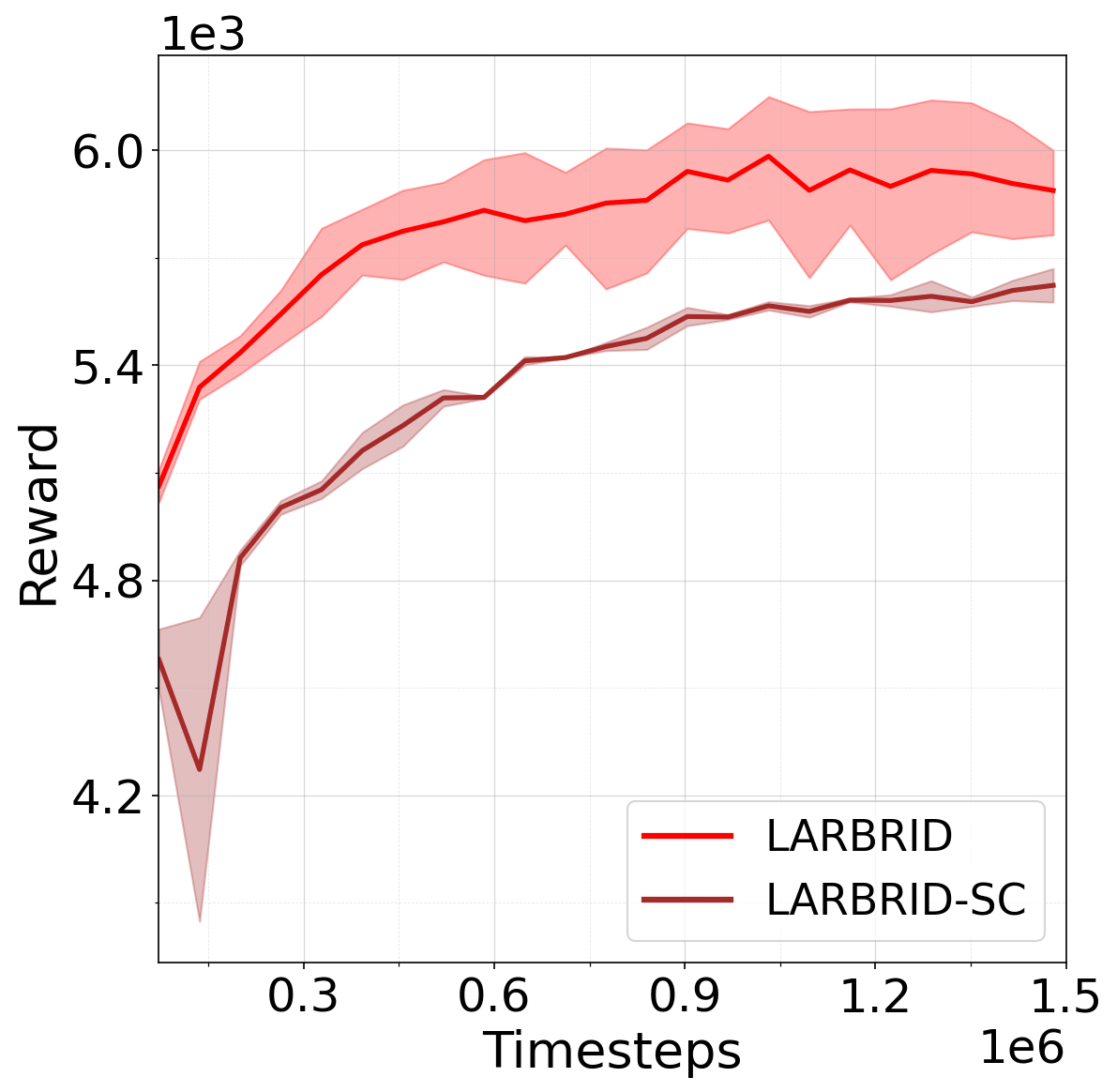}
     \end{subfigure}
\caption{When fine tuning $\pi_0$ using LARBRID, its evaluated success rate converges faster than DSRL across all plots shown in (a), (b) and (c). Separate critic (SC) ablation studies carried in Halfcheetah shown in (d) confirms the benefits brought by the Bridge Equation.}\label{fig: pi0 and ablation}
\vspace{-5mm}
\end{figure}
This section further stretches the capability of LARBRID by testing whether it can be used to fine tune robotic foundation models online. We choose to fine tune $\pi_0$ \cite{black2024pi_0}, a 3.3B parameter VLA model that uses a VLM-based backbone to process visual and language inputs and adopt a flow policy to output control actions. We test LARBRID on three simulated benchmarks with different dynamics. The first one is \texttt{AlohaTransferCube} \cite{zhao2023learning} that requires two aloha robots to cooperate, one picks up a cube and transfers it to another with high precision. The other two tasks are from the Libero \cite{liu2023libero} benchmark that requires the robot to open a drawer and turn on a stove. We compare against DSRL, the strongest online fine-tuning baseline and dropped RLPD \cite{ball2023efficient} which has been shown to be dominated by DSRL. The plots are shown in Figure~\ref{fig: pi0 and ablation}. Both LARBRID and DSRL are able to fine tune $\pi_0$ to achieve high success rate, however, LARBRID converges faster than DSRL across all tasks. In the \texttt{AlohaTransferCube} task, LARBRID is able to achieve a success rate $~2\times$ higher than DSRL at early training steps (50k-80k) and improves steadily afterwards whereas DSRL's performance fluctuates more and converge to a suboptimal point.
\vspace{-3mm}

\subsection{Ablations}
To isolate the effect of the Bridge Equation and demonstrate that the noise aliasing problem is indeed solved by it, we carefully design an ablation of LARBRID by creating a separate noise critic network that takes in the noise and state inputs instead of the learned representation $\psi$ and matches this new critic against the action critic trained from the learned action-state representation $\phi$. In other words, we keep the state-action representation but do not bridge it to the noise-action space such that the variant needs to re-solve the noise-aliasing problem online. We call this variant LARBRID-SC (`SC' for separate critics). We compare LARBRID-SC against LARBRID on the Halfcheetah task and the training plot is given in (d) of Figure~\ref{fig: pi0 and ablation}. We can see that the representation from LNL-MDP still brings massive performance gains, but because LARBRID-SC now needs to handle the noise-aliasing problem, it converges much slower and to a point inferior to that of LARBRID. Additional separate critic ablation results is provided in Figure~\ref{fig: ablation plots}. Furthermore, in Appendix~\ref{appendix: ablations}, we also test the sensitivity of LARBRID to representation learning and matching errors by introducing weights to the representation learning loss (line 4 of Algorithm~\ref{alg: LARBRID}) and to the representation matching loss (line 5 of Algorithm~\ref{alg: LARBRID}). Overall, we found that LARBRID is quite robust to the underlying representation shifts.

\vspace{-3mm}
\section{Discussions}
\textbf{Limitations} We suspect that when LARBRID gets stuck in local optimal features due to complex optimization landscape of hard problems like humanoid maze, the representation is not good enough for downstream fine-tuning, causing poor performances. Understanding such failure modes better can further boost the practical performances of LARBRID. It is also important to understand when a pre-trained generative model can be efficiently fine tuned by LARBRID, as this helps with providing a good prior for fine tuning. Due to the extra DDPM sampling induced by \eqref{eq: feature score match}, our method trains slower than the method LARBRID is based on but share the same inference speed, see Figure.\ref{fig:train-time}. This is somewhat expected as sample efficient algorithms typically require extra computation to process more information out of each datum.


\textbf{Conclusion} The representation power brought by LNL-MDP allows LARBRID to handle diverse RL fine tuning settings across different tasks. By embedding the fine tuning process into the learned latent representation space, we theoretically remove the noise-aliasing problem and show a near-optimal fine tuning rate when the representation is known. Empirically, our algorithm LARBRID and LARBRID-I work well in a diverse set of tasks and achieves a performance increase of up to 2-3 $\times$ compared to other baselines. In particular, the flexibility of LNL-MDP allows one to pick an easier space to fine-tune a model when the other space is too hard. The generality of LNL-MDP means it may even benefit future RL fine tuning pipelines or problems from different domains and motivate further representation learning for RL fine tunings.
\newpage
\section{Acknowledgments}
Yuanchu Liang is partially supported by Australia RTP scholarship and conducted this research under the AI Residency program in Maincode, Australia. Yuanchu Liang and Hanna Kurniawati are partially supported by ARC
Research Hub in Intelligent Robotic Systems for Real-Time Asset Management (IH210100030). Hanna Kurniawati is partially supported by the Australasian Space Innovation Institute (fka SmartSat CRC). Yue Yang and An-Chi He are supported by Maincode.
\bibliographystyle{abbrv}
\bibliography{ref}

\newpage
\appendix
\section{Linear MDPs for Non-Linear Dynamics}\label{appendix: linear mdp example}
Linear MDPs impose a linear decomposition on the transition and reward functions of a MDP. This may seem restrictive in the beginning, however, representations can be made arbitrarily non-linear and indeed many non-linear dynamics can be captured by this model when considering infinite dimensional spaces. We here provide a simple example of how this can be done. Consider the stochastic system $f: S\times A \rightarrow \mathbb{R}$,
$$
s' = f(s,a) + \epsilon, \text{ for } \epsilon \sim \mathcal{N}(0, \sigma^2),
$$
which can be equivalently written as,
$$
P(s'\mid s, a) \propto \exp\left(-\frac{||s' - f(s,a)||^2_2}{2\sigma^2}\right).
$$
Note there is no restrictions on the form $f$ might take, a linear system can be written as $f(s,a) = As + Ba$ for some design matrices $A, B$. Using the reproducing property of Gaussian kernels, we can equivalently write the density above as,
$$
P(s'\mid s, a) = \left\langle k(f(s,a), \cdot), (2\pi\sigma^2)^{d/2}k(s', \cdot)\right\rangle_{\mathcal{H}},
$$
Here $k(x, y) = \exp{\frac{-||x-y||^2}{2\sigma^2}}$ is the Gaussian kernel with bandwidth $\sigma$ and $\mathcal{H}$ is the reproducing Hilbert space associated with the kernel \cite{wainwright2019high}. And as one can see the associated density satisfies the linear decomposition. A even more general example that uses EBMs instead of Gaussian to obtain the linear decomposition is given below.

\section{Diffusion Spectral Representation}\label{appendix: diffsr}
Here we give a brief overview of the diffusion spectral representation \cite{shribak2024diffusion}, the core method we use to learn the representation needed for LARBRID. It relies on parameterising the transition function as an energy based model (EBM), which admits linear decomposition with respect to the Gaussian kernel and whose score components can be learned via score matching. Specifically, when the transition function $P(\cdot)$ is parameterized as an EBM,
\begin{equation}
    P(s'\mid s, a) = \exp(\xi(s,a)^\top \nu(s') - \log Z(s,a)), Z(s,a) = \int\exp(\xi(s,a)^\top\nu(s'))\mathrm{d}s'
\end{equation}
where $\xi(\cdot)$ and $\nu(\cdot)$ together compute the score function of the EBM and $\log Z(\cdot)$ is the log partition function. Applying random Fourier feature \citep{rahimi2007random, dai2014scalable}, the spectral decomposition of the transition function admits,
\begin{equation}
    P(s'\mid s, a) = \left\langle \phi_w(s,a), \mu_w(s')\right\rangle_{\mathcal{N}(w)}
\end{equation}
such form can model arbitrary transition densities \citep{sasaki2018neural}. With a slight abuse of notation we also use $w$ to denote the noises used in this decomposition, specifically for $w\sim \mathcal N (0,I)$, we have
\begin{align}
    \phi_w(s, a) &= \exp(-i w^\top \xi(s,a))\exp(||\xi(s,a)||^2 / 2 - \log Z(s,a)) \\
    \mu_w(s) &= \exp(-i w^\top \nu(s'))\exp(||\nu(s)||^2 / 2)
\end{align}
and the log partition function can be linearly represented as $Z(s,a) = \langle \exp(-i w^\top \xi(s,a)+ ||\xi(s,a)||^2 / 2), u \rangle_{\mathcal{N}(w)}$. Hence knowing $\nu(\cdot)$ and $\xi(\cdot)$ are enough to obtain $\phi_w(\cdot)$ and $\mu_w(\cdot)$. But these features can be learned via diffusion score matching using the following objective,
\begin{equation}\label{eq: feature score match copy}
l_{\mathrm{diff}}(\xi, \nu) =  \min_{\xi, \nu} \mathbb{E}_\beta \mathbb{E} (s, a, \Tilde{s}', s') \left [|| \Tilde{s}' + \beta\xi(s,a)^\top\nabla\nu(s') -\sqrt{1-\beta}s'|| \right]
\end{equation}
where $\Tilde{s}'$ is the noise corrupted next state and $\beta$ is the noise schedule in standard DDPM style diffusion. Substituting $\phi_w(s, a)$ into \eqref{eq: linear q}, we obtain,
\begin{equation}    
    Q(s,a) = \langle \exp(-i w^T\xi(s,a) - \log\langle\exp(-iw^\top \xi(s,a)), u\rangle_{\mathcal{N}(w)}, \zeta^\pi_h\rangle) \rangle_{\mathcal{N}(w)}.
\end{equation}
That is $\phi^\star(s,a)$ can be represented by a fully connected neural network with 1 hidden layer, $\phi_{\theta_1}(s,a) = \sigma(W_2^\top \sigma(W_1^T\xi(s, a)))$ and the Q-value can be represented as $Q_\zeta(s,a) = \langle \phi_{\theta_1}(s, a), \zeta \rangle$. We refer readers to \cite{shribak2024diffusion} for more details.

\newpage
\section{Proof for Proposition~\ref{prop:latent_linearity}}
\label{proof:prop_latent_linearity}
\begin{proof}
Fix any stage \(h\in[H]\), state \(s\in\mathcal S\), latent code \(w\in\mathbb R^{d_w}\), and next state \(s'\in\mathcal S\). By Definition~\ref{def:latent_mdp},
\[
\widetilde P_h(s'\mid s,w)
=
P_h(s'\mid s,\Phi(s, w)).
\]
Applying Assumption~\ref{ass:linear_mdp} to the action \(a=\Phi(s,w)\) gives
\[
P_h(s'\mid s,\Phi(s, w))
=
\langle \phi_\star(s,\Phi(s,w)),\nu^\star_h(s')\rangle.
\]
By the definition of \(\psi^\star\),
\[
\langle \phi^\star(s,\Phi(s,w)),\nu^\star_h(s')\rangle
=
\langle \psi^\star(s,w),\nu^\star_h(s')\rangle.
\]
Hence
\[
\widetilde P_h(s'\mid s,w)=\langle \psi^\star(s,w),\nu^*_h(s')\rangle.
\]

Similarly,
\[
\mathbb{E}[\widetilde r_h(s,w)]
=
\langle \psi^\star(s,w),\theta^r_h\rangle.
=
\langle \phi^\star(s,\Phi(s, w)),\theta^r_h\rangle
=
\mathbb{E}[r_h(s,\Phi(s,w))]
\]

Finally, since \(\Phi(w;s)\in\mathcal A\) and \(\|\phi^\star(s,a)\|_2\le 1\) for all \((s,a)\),
\[
\|\psi^\star(s,w)\|_2
=
\|\phi^\star(s,\Phi(s,w))\|_2
\le 1.
\]
Therefore \(\widetilde M\) satisfies the linear-MDP definition with feature map \(\psi^\star, \nu^*_h\) and parameters \(\{\theta^r_h\}_{h=1}^H\).
\end{proof}
\newpage

\section{Proof for Proposition~\ref{prop:value-preservation}}
\label{proof:prop_value_preservation}
\begin{proof}
Let
\[
\bar r_h(s,a):=\mathbb E[r_h(s,a)]
\]
denote the expected reward in the original MDP, and define the latent expected reward by
\[
\widetilde{\bar r}_h(s,w)
:=
\bar r_h(s,\Phi(s, w)).
\]
By the definition of the latent MDP,
\[
\widetilde P_h(\cdot\mid s,w)
=
P_h(\cdot\mid s,\Phi(s, w)).
\]

We first prove that
\[
\widetilde V_0^\star\le V_0^\star.
\]

Let \(\mu=(\mu_h)_{h=1}^H\) be any latent Markov policy. Define its induced action policy \(\pi^\mu\) in the original MDP by
\[
\pi_h^\mu(\cdot\mid s)
:=
(\Phi_s)_\#\mu_h(\cdot\mid s).
\]
Equivalently, for every bounded measurable function \(f:\mathcal A\to\mathbb R\),
\[
\mathbb E_{a\sim\pi_h^\mu(\cdot\mid s)}[f(a)]
=
\mathbb E_{w\sim\mu_h(\cdot\mid s)}
[f(\Phi(s,w))].
\]

We claim that, for all \(h\in[H+1]\) and all \(s\in\mathcal S\),
\[
\widetilde V_h^\mu(s)
=
V_h^{\pi^\mu}(s),
\]
and, for all \(h\in[H]\), \(s\in\mathcal S\), and \(w\in\mathcal W\),
\[
\widetilde Q_h^\mu(s,w)
=
Q_h^{\pi^\mu}(s,\Phi(s, w)).
\]

We prove this by backward induction on \(h\). At the terminal stage \(H+1\), both value functions are zero by convention.

Assume the claim holds at stage \(h+1\). Then, for any \(s,w\),
\[
\begin{aligned}
\widetilde Q_h^\mu(s,w)
&=
\widetilde{\bar r}_h(s,w)
+
\mathbb E_{s'\sim \widetilde P_h(\cdot\mid s,w)}
[\widetilde V_{h+1}^\mu(s')]
\\
&=
\bar r_h(s,\Phi(s,w))
+
\mathbb E_{s'\sim P_h(\cdot\mid s,\Phi(s,w))}
[V_{h+1}^{\pi^\mu}(s')]
\\
&=
Q_h^{\pi^\mu}(s,\Phi(s,w)).
\end{aligned}
\]
Note this is essentially the Bridge Equation \eqref{eq: shared q}. The second equality uses the definition of the latent MDP and the induction hypothesis.

Now integrate both sides over \(w\sim\mu_h(\cdot\mid s)\):
\[
\begin{aligned}
\widetilde V_h^\mu(s)
&=
\mathbb E_{w\sim\mu_h(\cdot\mid s)}
[\widetilde Q_h^\mu(s,w)]
\\
&=
\mathbb E_{w\sim\mu_h(\cdot\mid s)}
[Q_h^{\pi^\mu}(s,\Phi(s, w))]
\\
&=
\mathbb E_{a\sim\pi_h^\mu(\cdot\mid s)}
[Q_h^{\pi^\mu}(s,a)]
\\
&=
V_h^{\pi^\mu}(s).
\end{aligned}
\]
Thus the induction is complete. In particular,
\[
\widetilde V_0^\mu
=
V_0^{\pi^\mu}
\le
V_0^\star.
\]
Since this holds for every latent policy \(\mu\),
\[
\widetilde V_0^\star
=
\sup_\mu \widetilde V_0^\mu
\le
V_0^\star.
\]

We now prove the reverse inequality:
\[
\widetilde V_0^\star\ge V_0^\star.
\]

By Assumption \ref{ass:surj_optimal}, there exists an optimal Markov policy \(\pi^\star\) for the original MDP and a latent Markov policy \(\mu^\star\) such that, for every stage \(h\) and every state \(s\) reachable under \(\pi^\star\),
\[
\pi_h^\star(\cdot\mid s)
=
(\Phi_s)_\#\mu_h^\star(\cdot\mid s).
\]
Therefore, when starting from the initial state \(s_1\), the original rollout under \(\pi^\star\) and the latent rollout under \(\mu^\star\), after decoding \(a_h=\Phi(w_h;s_h)\), induce the same distribution over state-action-reward trajectories. Equivalently, by the same backward-induction argument above applied on the reachable trajectory support,
\[
\widetilde V_0^{\mu^\star}
=
V_0^{\pi^\star}
=
V_0^\star.
\]
Since \(\widetilde V_0^\star\) is the supremum over all latent policies,
\[
\widetilde V_0^\star
\ge
\widetilde V_0^{\mu^\star}
=
V_0^\star.
\]

Combining the two inequalities gives
\[
\widetilde V_0^\star
=
V_0^\star.
\]
\end{proof}
\newpage
\section{Proof for Proposition~~\ref{prop:virtual_offline}}
\label{proof:prop_virtual_offline}


\begin{proof}
We prove the two claims separately.

\textbf{1. validity of the dataset: }Consider one transition tuple \((s_h^i,w_h^i,r_h^i,s_{h+1}^i)\) collected during the warm-up phase. By construction, at state \(s_h^i\) we first sample a noise vector \(w_h^i\) according to the Gaussian prior used by $\Phi$. The environment action executed is
\[
a_h^i=\Phi(s_h^i, w_h^i).
\]
By Definition~\ref{def:latent_mdp}, the latent transition kernel is
\[
\widetilde P_h(\cdot\mid s_h^i,w_h^i)
=
P_h(\cdot\mid s_h^i,\Phi(s_h^i, w_h^i)).
\]
Therefore the observed next state satisfies
\[
s_{h+1}^i \sim \widetilde P_h(\cdot\mid s_h^i,w_h^i),
\]
and the observed reward satisfies
\[
r_h^i = \widetilde r_h(s_h^i,w_h^i).
\]
Hence each collected tuple is a valid latent transition. Since different episodes are generated by independent rollouts, the trajectories are i.i.d.

\paragraph{2. covariance concentration: }
Fix a step \(h\). Define
\[
X_i := \psi^\star(s_h^i,w_h^i)\psi^\star(s_h^i,w_h^i)^\top - \Lambda^h.
\]
Then \(\{X_i\}_{i=1}^{N_{\mathrm{off}}}\) are i.i.d.\ zero-mean random matrices.

Since \(\|\psi^\star(s,w)\|_2\le 1\),
\[
\left\|
\psi^\star(s_h^i,w_h^i)\psi^\star(s_h^i,w_h^i)^\top
\right\|_{\mathrm{op}}
=
\|\psi^\star(s_h^i,w_h^i)\|_2^2
\le 1.
\]
Also, by Jensen's inequality,
\[
\|\Lambda^h\|_{\mathrm{op}}
=
\left\|
\mathbb E[\psi^\star(\psi^\star)^\top]
\right\|_{\mathrm{op}}
\le
\mathbb E\left\|\psi^\star(\psi^\star)^\top\right\|_{\mathrm{op}}
\le 1.
\]
Therefore
\[
\|X_i\|_{\mathrm{op}}\le 2.
\]

Let
\[
S_h := \sum_{i=1}^{N_{\mathrm{off}}} X_i
=
\Lambda_{\mathrm{off}}^h - N_{\mathrm{off}}\Lambda^h.
\]
A standard matrix Bernstein inequality implies that for any \(t>0\),
\[
\mathbb P\!\left(
\|S_h\|_{\mathrm{op}}\ge t
\right)
\le
2d\exp\!\left(
-\frac{t^2/2}{N_{\mathrm{off}} + 2t/3}
\right),
\]
up to absolute constants. Choosing
\[
t = C\sqrt{N_{\mathrm{off}}\, d \log(dH/\delta)}
\]
for a sufficiently large universal constant \(C\), and applying a union bound over \(h\in[H]\), yields that with probability at least \(1-\delta\),
\[
\left\|
\Lambda_{\mathrm{off}}^h - N_{\mathrm{off}}\Lambda^h
\right\|_{\mathrm{op}}
\le
C\sqrt{N_{\mathrm{off}}\, d \log(dH/\delta)}
\qquad
\text{for all } h\in[H].
\]
This proves the claim.
\end{proof}

\section{Proof for Theorem~\ref{thm:main_theory}}
\label{proof:main_theory}


\begin{proof}
We verify the assumptions required by FineTuneRL.
By Proposition~\ref{prop:latent_linearity}, the latent problem \(\widetilde M\) is itself a linear MDP with feature map \(\psi^\star\) and dimension \(d\). By Proposition~\ref{prop:virtual_offline}, the warm-up phase yields a valid offline dataset \(D_{\mathrm{off}}\) on \(\widetilde M\), and its empirical covariance is exactly the quantity required in the FineTuneRL formulation.
By Proposition~\ref{prop:value-preservation}, under Assumption~\ref{ass:surj_optimal}, an \(\varepsilon\)-optimal policy in the latent MDP is also an \(\varepsilon\)-optimal policy in the original MDP.

Therefore the fine-tuning problem is exactly a FineTuneRL problem on \(\widetilde M\). Applying the FineTuneRL upper-bound theorem gives
\[
T_{\mathrm{on}}(\varepsilon)
=
\widetilde O\!\left(
\max_{h\in[H]}
\min\left\{
T:\,
C_{o2o}^h(\Lambda_{\mathrm{off}}^h,\varepsilon,T)\le \frac{1}{\bar{\beta}}
\right\}
\cdot H
\right),
\]
and applying the corresponding lower-bound theorem yields
\[
T_{\mathrm{on}}(\varepsilon)
=
\widetilde \Omega\!\left(
\max_{h\in[H]}
\min\left\{
T:\,
C_{o2o}^h(\Lambda_{\mathrm{off}}^h,\varepsilon,T)\le \frac{1}{\bar{\beta}}
\right\}
\cdot \frac{1}{H}
\right).
\]
\end{proof}
\subsection{FTPedel Algorithm Used in Theorem 1}

For completeness, we restate an informal version of FTPedel from FineTuneRL~\citep{wagenmaker2023leveraging}
adapted to the latent noise MDP $\widetilde{M}_{\theta}$. 

\begin{algorithm}[H]
\caption{FTPedel on the Latent Noise MDP (Informal)}
\label{alg:ftpedel}
\begin{algorithmic}[1]
\Require $D_{\mathrm{off}}$, policy class $\Pi_{\mathrm{lsm}}$, target accuracy $\epsilon$, confidence level $\delta$
\State Construct the empirical offline covariance $\Lambda^{h}_{\mathrm{off}}$ for each stage $h$.
\State Select an exploration policy $\pi_{\mathrm{exp}}$ according to the offline-to-online concentrability criterion.
\State Collect $T$ additional online episodes using $\pi_{\mathrm{exp}}$.
\State Combine the offline dataset and online samples.
\State Estimate the linear value-function parameters over the latent feature space.
\State Return a policy $\widehat{\pi}$ that is $\epsilon$-optimal in $\widetilde{M}$.
\end{algorithmic}
\end{algorithm}

\begin{remark}
    In addition to proving the validity of dataset in Proposition~.\ref{prop:virtual_offline}, it also proves a concentration bound between $\Lambda^h$ and $\Lambda^h_{\mathrm{off}}$. A consequence is that in the asymptotic limit, Theorem.~\ref{thm:main_theory} can be stated in terms of the true covariate generated by $\Phi$, i.e., $\Lambda^h$. This statement reflects the prior information provided by the pre-trained model, however, it is less practical, as in general we do not have infinite interactions with $M$.
\end{remark}

\newpage
\section{Theory to Real Gap}
\label{appendix:theory-to-real-gap}
\subsection{Bridge to the practical learned-feature implementation}
\label{subsec:bridge_practical}

Theorem~\ref{thm:main_theory} proves a minimax guarantee in the idealized setting where the structural feature map \(\phi^\star\) is known. LARBRID relaxes this assumption by also learning $\phi^\star$. Instead of directly using the exact pulled-back feature
\[
\psi^\star(s,w)=\phi^\star(s,\Phi(s,w)),
\]
it learns an action-space feature map and a latent-space feature map and aligns them during training.

Let
\(\phi : \mathcal S\times\mathcal A \to \mathbb R^d
\)
denote the learned action-space feature map used in practice, and let
\(
\tilde\phi : \mathcal S\times W \to \mathbb R^d
\)
denote the learned latent-space feature map. Define again the ideal pulled-back feature
\(
\bar\phi(s,w):=\phi^\star(s,\Phi(s, w)).
\)

Our implementation cannot directly optimize through \(\bar\phi_\theta(s,w)\), because the decoder is frozen and treated as a black box during optimization. Instead, it learns \(\tilde\phi(s,w)\) and aligns it to the decoded action-space feature \(\phi(s,\Phi_\theta(s, w))\). The next proposition shows that the latent representation error is controlled by two quantities: the decoded action-feature approximation error and the latent feature-matching error.

Let \(\rho\) be any distribution over \(\Delta_{S\times W}\). In practice, \(\rho\) may be taken as the actor-induced state-latent distribution, since feature matching is enforced on actor-sampled latent codes.
\begin{definition}[Representation losses]
\label{def:repr_losses}
For any distribution \(\rho\) over \((s,w)\), define
\[
\mathcal L_{\mathrm{act}}(\rho)
:=
\mathbb E_{(s,w)\sim \rho}
\Big[
\|\phi(s,\Phi(s,w))-\phi^\star(s,\Phi(s,w))\|_2^2
\Big],
\]
and
\[
\mathcal L_{\mathrm{match}}(\rho)
:=
\mathbb E_{(s,w)\sim \rho}
\Big[
\|\tilde\phi(s,w)-\phi(s,\Phi(s,w))\|_2^2
\Big].
\]
\end{definition}

\begin{proposition}[In-distribution representation error bound]
\label{prop:repr_error_bound}
For any distribution \(\rho\),
\[
\mathbb E_{(s,w)\sim \rho}
\Big[
\|\tilde\phi(s,w)-\bar\phi(s,w)\|_2^2
\Big]
\le
2\,\mathcal L_{\mathrm{match}}(\rho)
+
2\,\mathcal L_{\mathrm{act}}(\rho).
\]
\end{proposition}

\begin{proof}
Fix any \((s,w)\). By definition,
\[
\bar\phi(s,w)=\phi^\star(s,\Phi(s,w)).
\]
Hence
\[
\tilde\phi(s,w)-\bar\phi(s,w)
=
\Bigl(\tilde\phi(s,w)-\phi(s,\Phi(s,w))\Bigr)
+
\Bigl(\phi(s,\Phi(s,w))-\phi^\star(s,\Phi(s,w))\Bigr).
\]
Let
\[
x:=\tilde\phi(s,w)-\phi(s,\Phi(s,w)),
\qquad
y:=\phi(s,\Phi(s,w))-\phi^\star(s,\Phi(s,w)).
\]
Then
\[
\tilde\phi(s,w)-\bar\phi(s,w)=x+y.
\]
Using
\[
\|x+y\|_2^2
=
\|x\|_2^2+2\langle x,y\rangle+\|y\|_2^2
\le
2\|x\|_2^2+2\|y\|_2^2,
\]
we obtain
\[
\|\tilde\phi(s,w)-\bar\phi(s,w)\|_2^2
\le
2\|\tilde\phi(s,w)-\phi(s,\Phi(s,w))\|_2^2
+
2\|\phi(s,\Phi(s,w))-\phi^\star(s,\Phi(s,w))\|_2^2.
\]
Taking expectation over \((s,w)\sim \rho\) yields
\[
\mathbb E_{(s,w)\sim \rho}
\Big[
\|\tilde\phi(s,w)-\bar\phi(s,w)\|_2^2
\Big]
\le
2\,\mathcal L_{\mathrm{match}}(\rho)
+
2\,\mathcal L_{\mathrm{act}}(\rho).
\]
\end{proof}

\begin{corollary}[Root-mean-square representation error]
\label{cor:repr_rms}
Under the same conditions as Proposition~\ref{prop:repr_error_bound},
\[
\left(
\mathbb E_{(s,w)\sim \rho}
\Big[
\|\tilde\phi(s,w)-\bar\phi(s,w)\|_2^2
\Big]
\right)^{1/2}
\le
\sqrt{2\,\mathcal L_{\mathrm{match}}(\rho)+2\,\mathcal L_{\mathrm{act}}(\rho)},
\]
and
\[
\mathbb E_{(s,w)\sim \rho}
\Big[
\|\tilde\phi(s,w)-\bar\phi(s,w)\|_2
\Big]
\le
\sqrt{2\,\mathcal L_{\mathrm{match}}(\rho)+2\,\mathcal L_{\mathrm{act}}(\rho)}.
\]
\end{corollary}

\begin{proof}
The first inequality follows by taking square roots on both sides of Proposition~\ref{prop:repr_error_bound}. The second follows from Jensen's inequality:
\[
\mathbb E\big[\|\tilde\phi-\bar\phi\|_2\big]
\le
\sqrt{\mathbb E\big[\|\tilde\phi-\bar\phi\|_2^2\big]},
\]
combined with Proposition~\ref{prop:repr_error_bound}.
\end{proof}

\begin{assumption}[Uniform control of component errors on a support set]
\label{ass:uniform_components}
There exists a set \(\Omega\subseteq \mathcal S\times \mathbb W\) such that for all \((s,w)\in\Omega\),
\[
\|\phi(s,\Phi(s,w))-\phi^\star(s,\Phi(s,w))\|_2 \le \varepsilon_{\mathrm{act}},
\]
and
\[
\|\tilde\phi(s,w)-\phi(s,\Phi(s,w))\|_2 \le \varepsilon_{\mathrm{match}}.
\]
\end{assumption}

\begin{corollary}[Uniform representation error bound]
\label{cor:uniform_repr}
Under Assumption~\ref{ass:uniform_components}, for all \((s,w)\in\Omega\),
\[
\|\tilde\phi(s,w)-\bar\phi(s,w)\|_2
\le
\varepsilon_{\mathrm{match}}+\varepsilon_{\mathrm{act}}.
\]
Consequently,
\[
\sup_{(s,w)\in\Omega}
\|\tilde\phi(s,w)-\bar\phi(s,w)\|_2
\le
\varepsilon_{\mathrm{match}}+\varepsilon_{\mathrm{act}}.
\]
\end{corollary}

\begin{proof}
Fix any \((s,w)\in\Omega\). By definition,
\[
\bar\phi(s,w)=\phi^\star(s,\Phi(s,w)).
\]
Therefore
\[
\tilde\phi(s,w)-\bar\phi(s,w)
=
\Bigl(\tilde\phi(s,w)-\phi(s,\Phi(s,w))\Bigr)
+
\Bigl(\phi(s,\Phi(s,w))-\phi^\star(s,\Phi(s,w))\Bigr).
\]
Applying the triangle inequality,
\begin{align*}
\|\tilde\phi(s,w)-\bar\phi(s,w)\|_2
&\le
\|\tilde\phi(s,w)-\phi(s,\Phi(s,w))\|_2
+
\|\phi(s,\Phi(s,w))-\phi^\star(s,\Phi(s,w))\|_2 \\
&\le
\varepsilon_{\mathrm{match}}+\varepsilon_{\mathrm{act}},
\end{align*}
where the last inequality uses Assumption~\ref{ass:uniform_components}. Taking the supremum over \(\Omega\) yields the result.
\end{proof}

\begin{remark}[Interpretation of the bridge result]
Proposition~\ref{prop:repr_error_bound} is the natural statement for the practical algorithm, because feature matching only controls the latent representation on the sampled training distribution. Corollary~\ref{cor:uniform_repr} is stronger, but it requires stronger assumptions: namely, that both the decoded action-feature error and the latent matching error are already uniformly controlled on the support set of interest. In applications, \(\Omega\) may be taken to be the support of the actor-induced state-latent distribution, or any restricted region where uniform approximation quality is plausible.
\end{remark}

\subsection{Scope of the guarantee}
\label{subsec:scope}

At the same time, our practical implementation is more general than this idealized setting. In practice, the exact structural feature map \(\phi_\star\) is not known, and the exact pulled-back feature \(\psi^\star(s,w)=\phi^\star(s,\Phi(s,w))\) cannot be used directly for optimization because the decoder is frozen and treated as a black box. We therefore learn an action-space feature map \(\phi\) and a latent-space feature map \(\tilde\phi\) and align them during training. Proposition~\ref{prop:repr_error_bound} and Corollaries~\ref{cor:repr_rms}--\ref{cor:uniform_repr} make this bridge explicit: if both the decoded action-feature approximation error and the latent feature-matching error are small, then the learned latent representation remains close to the ideal pulled-back feature required by the theory.

Accordingly, the role of Theorem~\ref{thm:main_theory} is to justify the latent diffusion-steering \emph{framework}, while the role of Proposition~\ref{prop:repr_error_bound} and its corollaries is to show why the learned-feature implementation can be viewed as a practical approximation to that idealized framework. We emphasize, however, that the paper does \emph{not} claim a minimax theorem for the entire end-to-end training pipeline, including learned feature matching and SAC-style optimization. Analyzing those components jointly would require additional assumptions and new optimization theory beyond the scope of the present work.
\newpage

\section{Experiment Details}\label{appendix: exp-details}
We provide general implementation details and best practices for LARBRID here. These details and recommendations are applied to all problems. Hyperparameters can be found in Table~\ref{tab: hyperparameters}.

\textbf{Diffusion Spectral Representation}. We found it is sufficient to use the same feature dimension as the hidden dimensions across all tasks. We use 50 DDPM sampling steps when solving the diffusion score matching problem in \eqref{eq: feature score match}. We use variance preserving noise schedule in all trainings. We initialize the noise feature $\psi_{\theta_2}$ (if applicable) with exactly the same network used for $\phi_{\theta_1}$. Importantly we also keep a target copy of the learned feature $\phi_{\theta_1}$, and update the target parameters with Polyak averaging rate of $0.005$. This slow moving target anchors the Q-network target to ensure the underlying representation space is not changing so fast that may cause learning instabilities from Q value bootstrapping. To further stablise the pipeline, we also set a slightly slower learning rate for the DiffSR features than the critic and the actor.

\textbf{Critics and Actors}. Due to the inherent non-linearity of the soft Q-value from SAC, the potential OOD errors from the offline dataset and the complex interactions of different networks within LARBRID, we found generally setting the critic and actor to have 2 or 3 layers of hidden dimensions are important (instead of 1) to saturate these nonlinear factors during learning. We apply the clipped double Q-network trick \cite{fujimoto2018addressing} to stablise TD computations. The learning rate for both the critic and actor is set to $3e\text{-}4$.

\textbf{Visual Encoders}. Instead of directly working in the image space for visual inputs, we encode the input image and embed the inputs in a lower dimensional space and run LARBRID over this lower dimensional encoded space. We use slightly different encoding schemes for OGBench visual manipulation tasks and $\pi_0$ fine tuning tasks. Details of these encoders used in offline OGBench and online $\pi_0$ fine tuning are reported below.

\subsection{Details of Online Fine Tuning Tasks}\label{appendix: online}
\begin{figure}[htbp]
     \centering
     \begin{subfigure}[b]{0.24\textwidth}
         \centering
         \includegraphics[width=\textwidth]{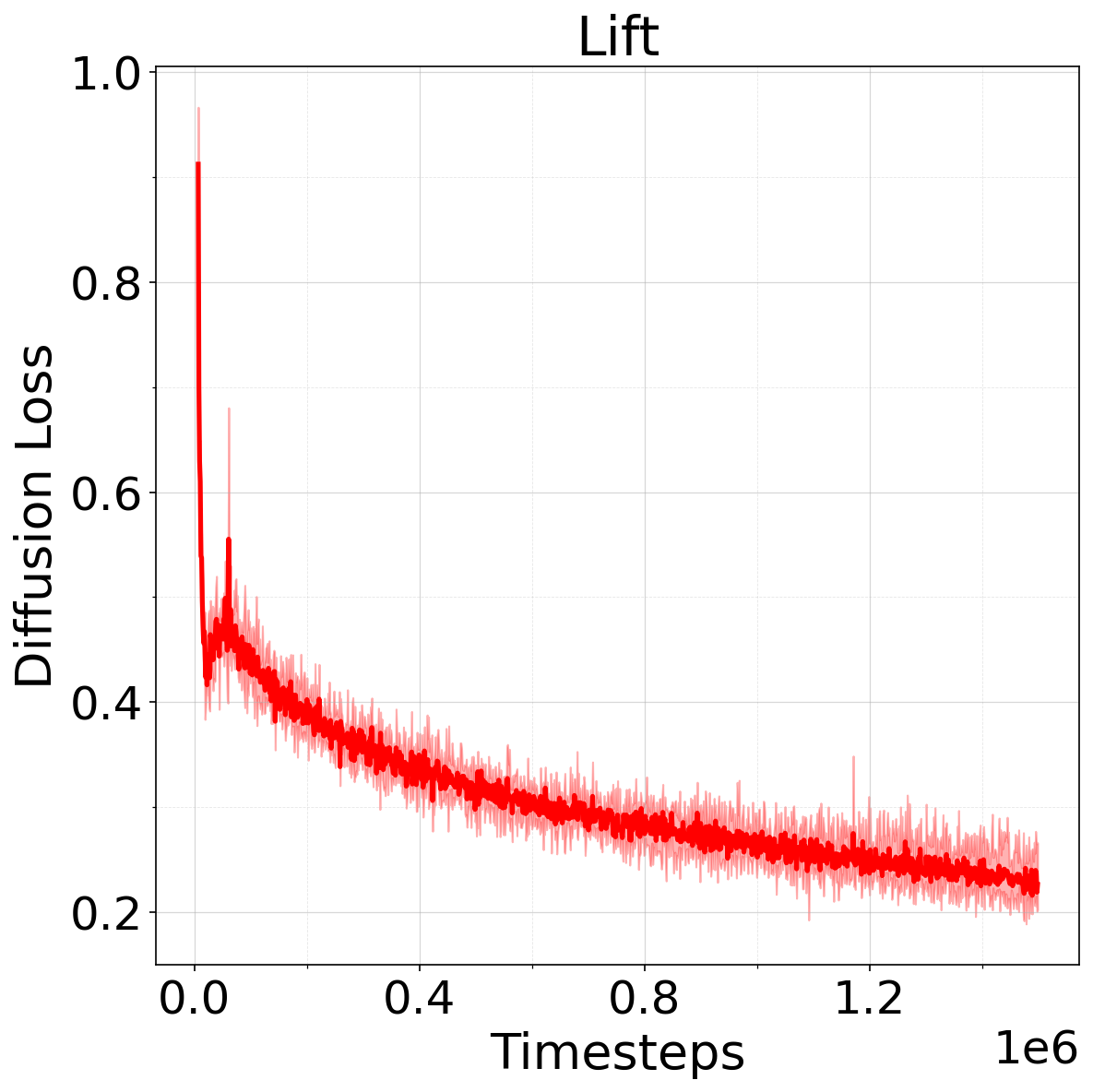}
     \end{subfigure}
     \hfill
     \begin{subfigure}[b]{0.24\textwidth}
         \centering
         \includegraphics[width=\textwidth]{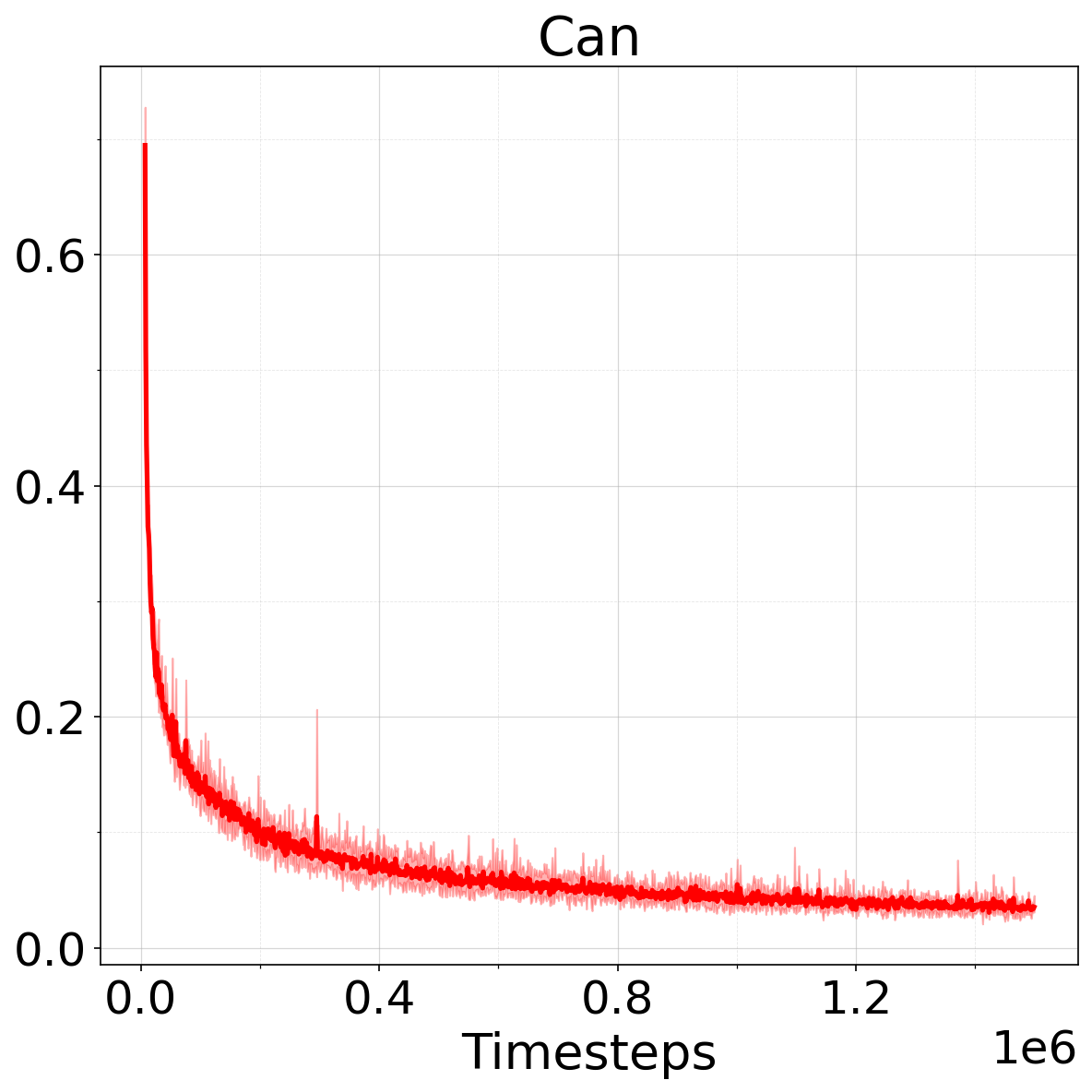}
     \end{subfigure}
     \hfill
     \begin{subfigure}[b]{0.24\textwidth}
         \centering
         \includegraphics[width=\textwidth]{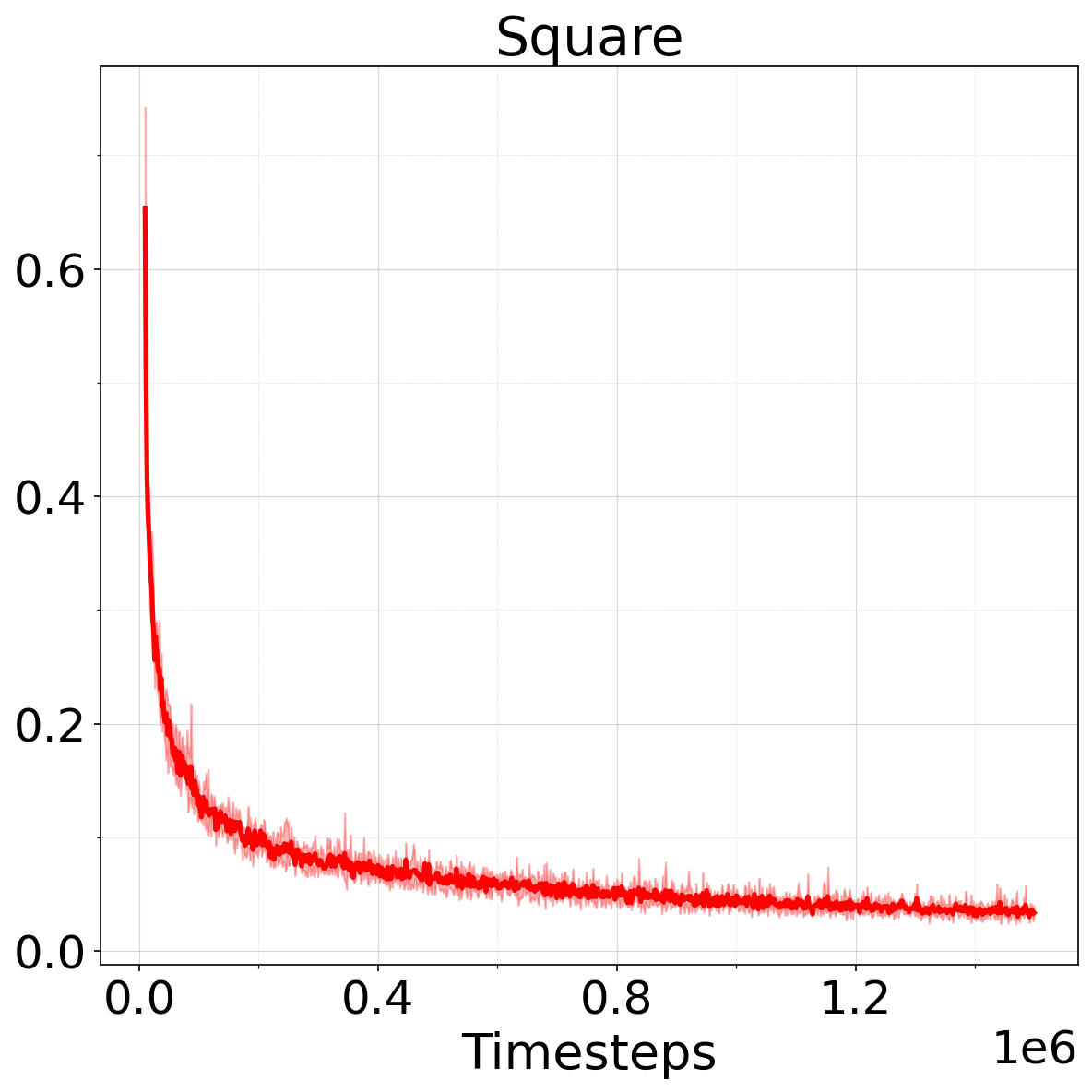}
     \end{subfigure}
     \begin{subfigure}[b]{0.24\textwidth}
         \centering
         \includegraphics[width=\textwidth]{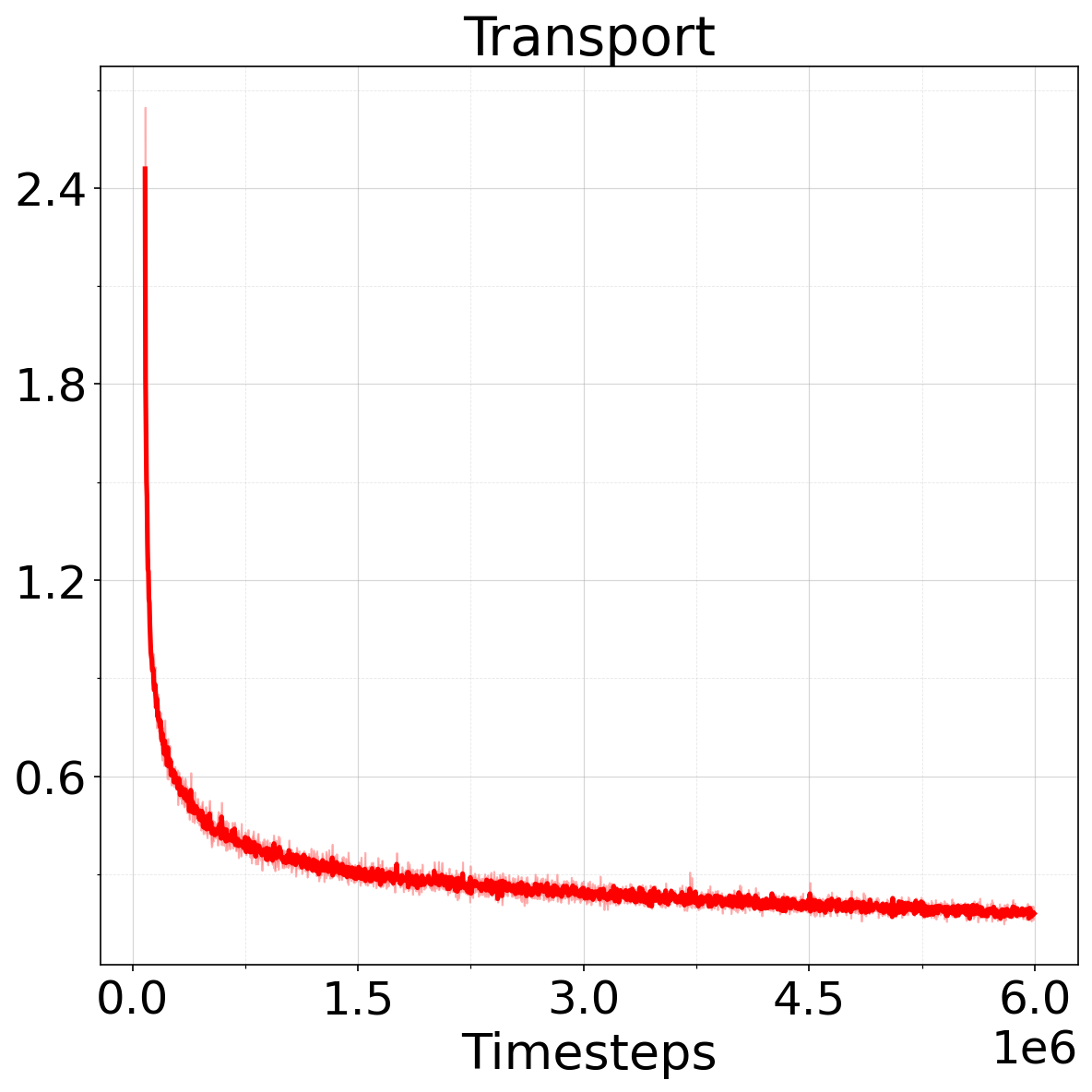}
     \end{subfigure}
        \caption{Spectral representation diffusion losses of LARBRID on Robomimic tasks.}
        \label{fig:robomimic-diffusion-losses}
\end{figure}

Initially, we collect 1500 data samples from the environment by rolling out the pre-trained DDIM. The hidden layer dimension of the critic and actor is set to 1024. We found using two hidden layers is sufficient for all the online experiments. The diffusion spectral representation losses for Robomimic tasks are shown in Figure~\ref{fig:robomimic-diffusion-losses}. Although we do not know the true representation $\phi^\star(\cdot)$ for each experiment, the efficiently minimized losses combined with the fast converging rewards and success rates demonstrate the effectiveness of LNL-MDPs and LARBRID. Figure~\ref{fig:robomimic-diffusion-losses} shows that the representation learning stage converges quickly so that downstream RL training can occur efficiently, as our theory also suggests.


\subsection{Details of Offline Fine Tuning Tasks}\label{appendix: offline}

\textbf{Name Mapping}. The full name of each task run in OGBench is given below.
\begin{itemize}
    \item ant-large: \texttt{antmaze-large-navigate-singletask-v0}
    \item ant-giant: \texttt{antmaze-giant-navigate-singletask-v0}
    \item hum-med: \texttt{humanoidmaze-medium-navigate-singletask-v0}
    \item hum-large: \texttt{humanoidmaze-large-navigate-singletask-v0}
    \item soccer: \texttt{antsoccer-arena-navigate-singletask-v0}
    \item cube-s: \texttt{cube-single-play-singletask-v0}
    \item cube-d: \texttt{cube-double-play-singletask-v0}
    \item scene: \texttt{scene-play-singletask-v0}
    \item pz-3x3:\texttt{puzzle-3x3-play-singletask-v0}
    \item pz-4x4:\texttt{puzzle-4x4-play-singletask-v0}
\end{itemize}

\textbf{Feature OOD Errors}. LARBRID displayed in Algorithm~\ref{alg: LARBRID} can theoretically work with pure offline data, we found doing so causes unstable training and fails to solve hard tasks (e.g. humanoid maze, ant soccer, puzzle 4x4) from OGBench. This is because of the feature representation OOD error. The environment MDP feature $\phi_{\theta_1}(\cdot)$ is learned on the offline dataset, which is collected from the unknown behavior policy. However, if we try to perform feature matching according to line 9 of LARBRID, we will have to generate noises from the current noise policy $\pi_{\mathrm{dp}, \rho}(\cdot)$ and sample a potentially OOD action to match the features, causing wrong $\psi_{\theta_2}(\cdot)$ learned. Even worse, if the offline data does not cover the spaces well, $\phi_{\theta_1}$ will not be representative enough for LARBRID to work. One can resort to offline RL techniques like SAC-N \cite{an2021uncertainty} and BRAC \cite{wu2019behavior}, they often come with heavy hyper-parameter tuning or large computation burden. In fact, this is how DSRL\cite{wagenmaker2025steering} solves their problem with 10 critic ensembles. 

Instead, we leverage the Bridge Equation to work with only the environment MDP $M$, since finding an optimal actor in $M$ is equivalent to finding an optimal noise actor in $\widetilde M$. This avoids the aforementioned feature OOD error, however, we will have to perform backpropagation through time to fine tune the pre-trained model. Luckily, FQL \cite{fql_park2025} has solved this problem by simultaneously learning a behavior flow policy via behavior cloning and a one-step distilled policy whose objective is to stay close to the BC policy while maximizing the critic value. LARBRID-I builds on top of this idea by adding an extra representation learning stage and condition the FQL algorithm on the learned feature space (Algorithm~\ref{alg:larbrid_i}). All FQL parameters are set according to their paper with the only exception that we lower the number of hidden dimensions in the critic and actor from 4 to 3. The spectral representation learning part is set to learned $\phi_{\theta_1}(\cdot)$ with the same hyper-parameters and network architecture as before. There is no extra hyper-parameter tuning needed.

\begin{algorithm}[H]
\caption{Latent Representation Bridging for Diffusion - Implicit (LARBRID-I)}
\label{alg:larbrid_i}
\begin{algorithmic}[1]
\Require Offline dataset $\mathcal{D}$, total steps $T$, representation parameters $\theta_1$, BC flow vector field parameters $\theta$, one-step policy parameters $\omega$, critic parameters $\zeta$, discount factor $\gamma$, distillation weight $\alpha$.
\For{$t=1, \dots, T$}
    \State Sample batch $(s, a, s', r)$ from $\mathcal{D}$.
    
    \Statex \textit{// Update Representation $\phi_{\theta_1}$:}
    \State $\min_{\theta_1} \mathbb{E}_{\beta} \mathbb{E}_{(s,a,\tilde{s}',s')} \left[ ||\tilde{s}' + \beta \phi_{\theta_1}(s,a)^\top \nabla \nu(s') - \sqrt{1-\beta}s'||_2 \right]$
    
    \Statex \textit{// Update BC Flow Policy $\mu_\theta$:}
    \State Sample $x^0 \sim \mathcal{N}(0, I_d)$ and $t \sim \text{Unif}([0,1])$.
    \State Set $x^1 = a$ and interpolate $x^t = (1-t)x^0 + tx^1$.
    \State $\min_\theta \mathbb{E} \left[ ||v_\theta(t, s, x^t) - (x^1 - x^0)||_2^2 \right]$
    
    \Statex \textit{// Update Critic $\zeta$:}
    \State Sample $z \sim \mathcal{N}(0, I_d)$ and obtain next action via one-step policy $a' = \mu_\omega(s', z)$.
    \State $\min_\zeta \mathbb{E}_{\mathcal{D}} \left[ \left( Q_\zeta(\phi_{\theta_1}(s, a)) - r - \gamma \bar{Q}_{\zeta'}(\phi_{\theta_1}(s', a')) \right)^2 \right]$
    
    \Statex \textit{// Update Actor (One-Step Policy) $\omega$:}
    \State Sample $z \sim \mathcal{N}(0, I_d)$ and obtain current action $a^\pi = \mu_\omega(s, z)$.
    \State $\min_\omega \mathbb{E}_{\mathcal{D}} \left[ -Q_\zeta(\phi_{\theta_1}(s, a^\pi)) + \alpha ||a^\pi - \mu_\theta(s, z)||_2^2 \right]$
\EndFor
\end{algorithmic}
\end{algorithm}

\textbf{Visual Manipulation Tasks}
We select 5 classic manipulation tasks from the OGBench environment and test our method's capability of handling visual inputs. 

Similar to \cite{shribak2024diffusion}, we use the CNN autoencoder structure in \cite{zhang2020learning} to learn a latent 1D embedding for each image by training the autoencoder with a standard image reconstruction loss. To handle partial observations, we concatenate 3 image latent embeddings and use such concatenated vector as observation inputs to LARBRID. We run a warm up stage for the autoencoder after the initial rollout so that the DDPM can obtain meaningful inputs at the beginning. The rest of the training pipeline is kept the same.

Due to the high computational costs, we select five strongest performing baselines listed in \cite{fql_park2025} to compare against. 
\begin{table*}[htpb]
\vspace{-2mm}
\centering
\caption{Visual Manipulation Tasks from OGBench}
\label{tab:visual-ogbench}
\footnotesize
\setlength{\tabcolsep}{3pt}
\renewcommand{\arraystretch}{1}
\begin{tabular}{lcccccc}
\hline
Task                                        & IQL       & ReBRAC   & FBRAC    & IFQL      & FQL        & LARBRID-I \\ \hline \hline
visual-cube-single-play-singletask-task1-v0 & $70\pm12$ & $\bm{83\pm6}$ & $55\pm8$ & $49\pm7$  & $\bm{81\pm12}$  & $\bm{83\pm8}$  \\
visual-cube-double-play-singletask-task1-v0 & $\bm{34\pm23}$ & $4\pm4$  & $6\pm2$  & $8\pm6$   & $21\pm11$  & $\bm{32\pm11}$ \\
visual-scene-play-singletask-task1-v0       & $\bm{97\pm2}$  & $\bm{98\pm4}$ & $46\pm4$ & $86\pm10$ & $\bm{98\pm 3}$  & $\bm{97\pm3}$  \\
visual-puzzle-3x3-play-singletask-task1-v0  & $7\pm 15$ & $88\pm4$ & $7\pm2$  & $\bm{100\pm0}$ & $94 \pm 1$ & $87\pm7$  \\
visual-puzzle-4x4-play-singletask-task1-v0  & $0\pm0$   & $26\pm6$ & $0\pm0$  & $8\pm15$  & $33\pm6$   & $\bm{49\pm16}$ \\ \hline
\end{tabular}
\end{table*}

\begin{figure}[htbp]
     \centering
     \begin{subfigure}[b]{0.35\textwidth}
         \centering
         \includegraphics[width=\textwidth]{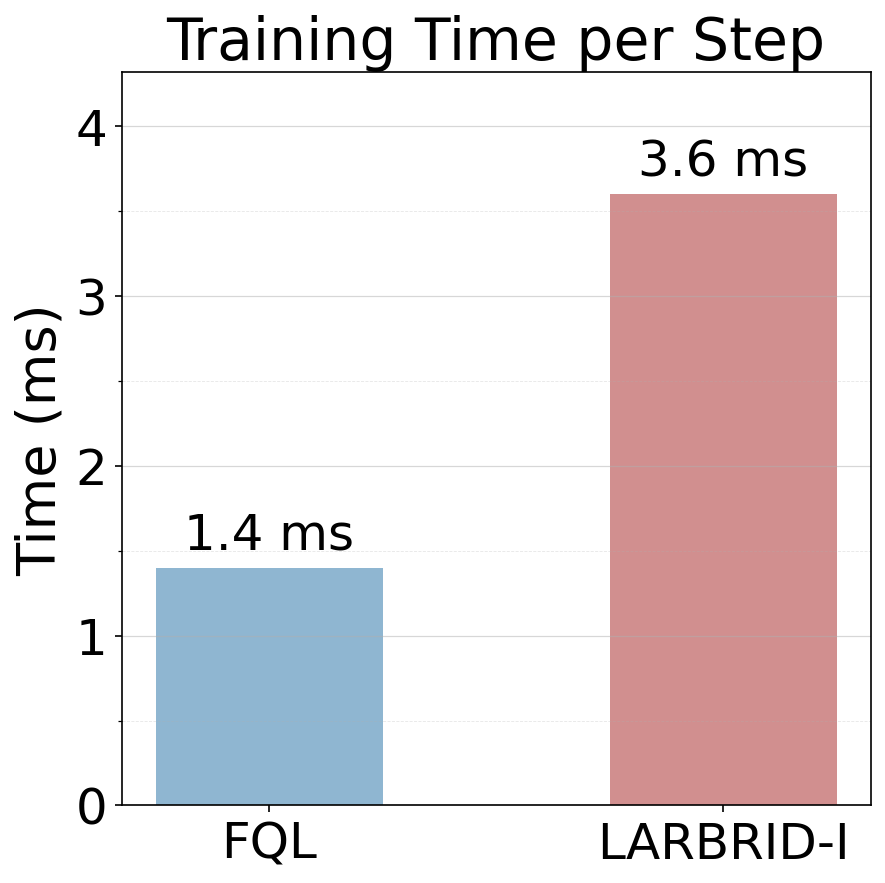}
     \end{subfigure}
     \hfill
     \begin{subfigure}[b]{0.35\textwidth}
         \centering
         \includegraphics[width=\textwidth]{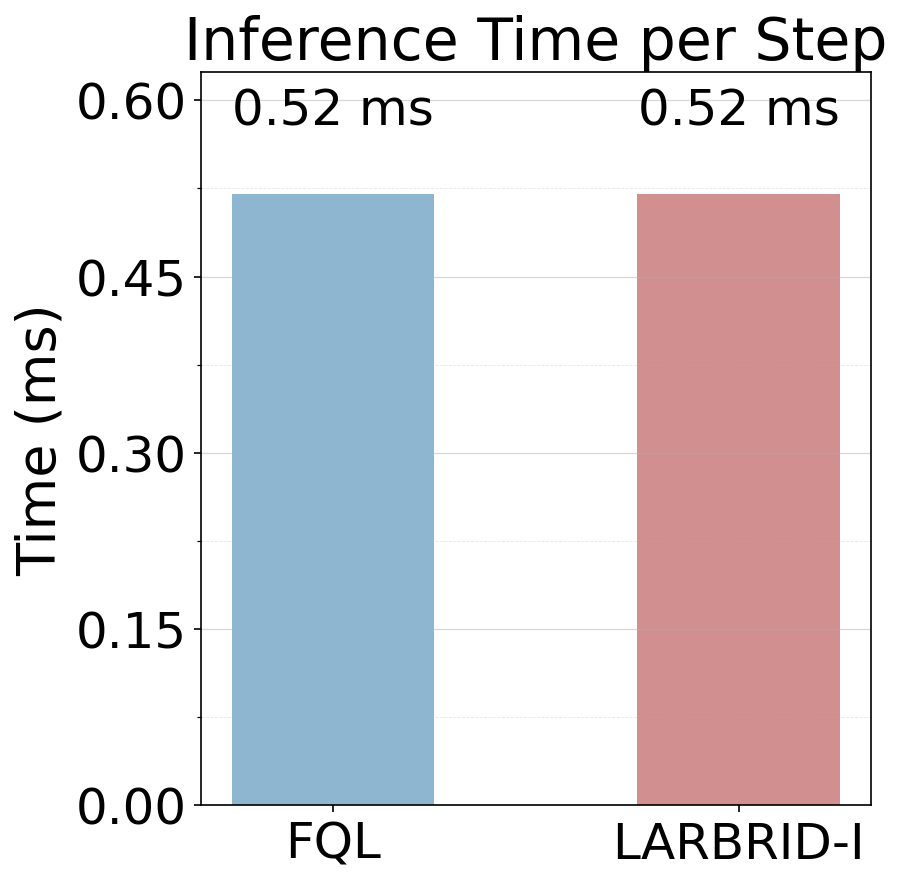}
     \end{subfigure}
        \caption{Training and inference time comparisons between LARBRID-I and FQL.}
        \label{fig:train-time}
\end{figure}

\textbf{Computation Times}. We report the training time of LARBRID-I against FQL on a single NVIDIA 4090 in Figure \ref{fig:train-time}.  LARBRID-I runs $~2.5\times$ slower when compared to FQL due to the extra DDPM sampling (50 sampling steps) induced when performing diffusion score matching during learning. During inference, the same actor is queried for both methods and hence they share the same inference time.

\subsection{Details of Offline-to-online Fine Tuning Tasks}\label{appendix:o2o}
\begin{figure}[htbp]
     \centering
     \begin{subfigure}[b]{0.24\textwidth}
         \centering
         \includegraphics[width=\textwidth]{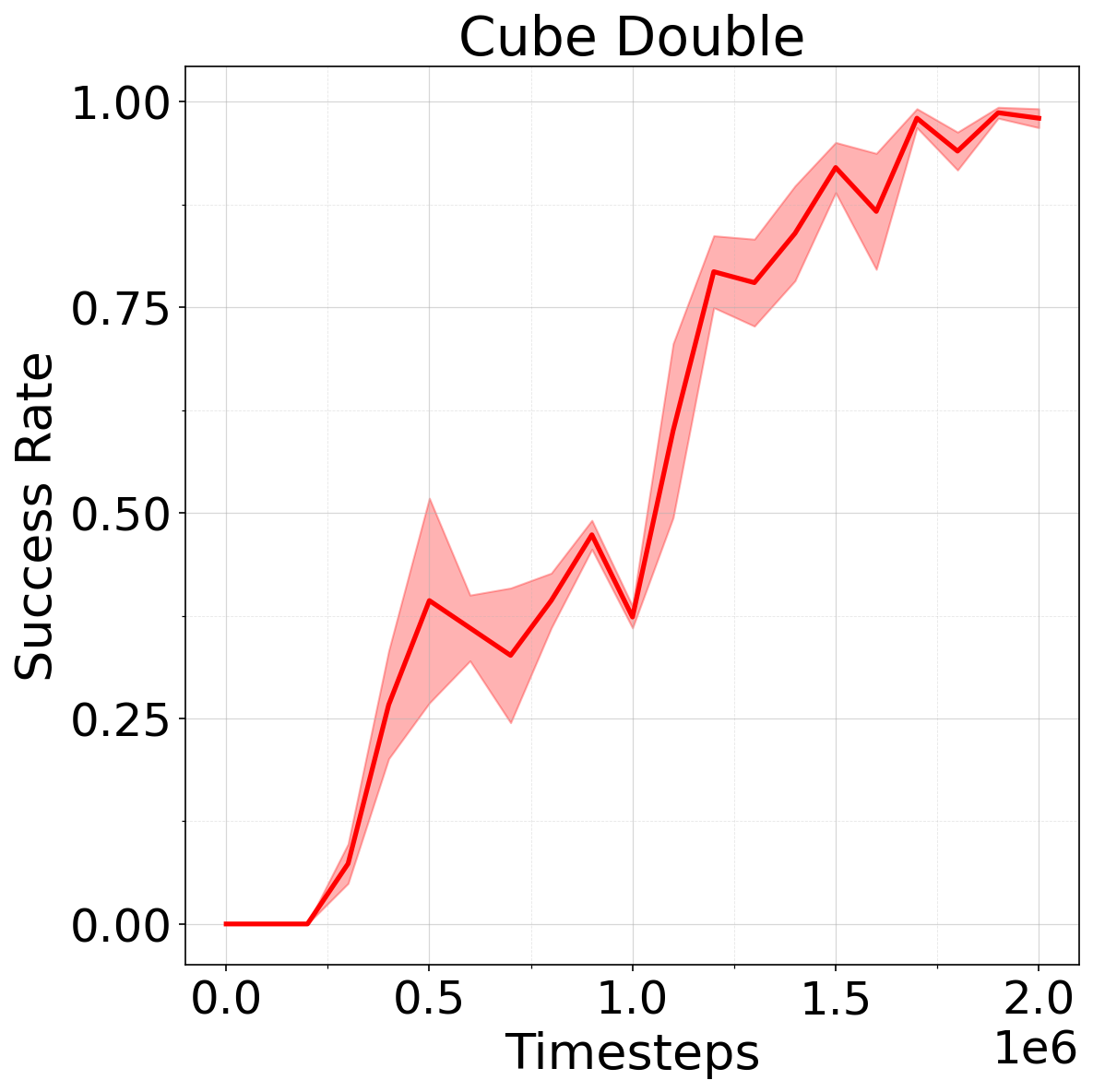}
     \end{subfigure}
     \hfill
     \begin{subfigure}[b]{0.24\textwidth}
         \centering
         \includegraphics[width=\textwidth]{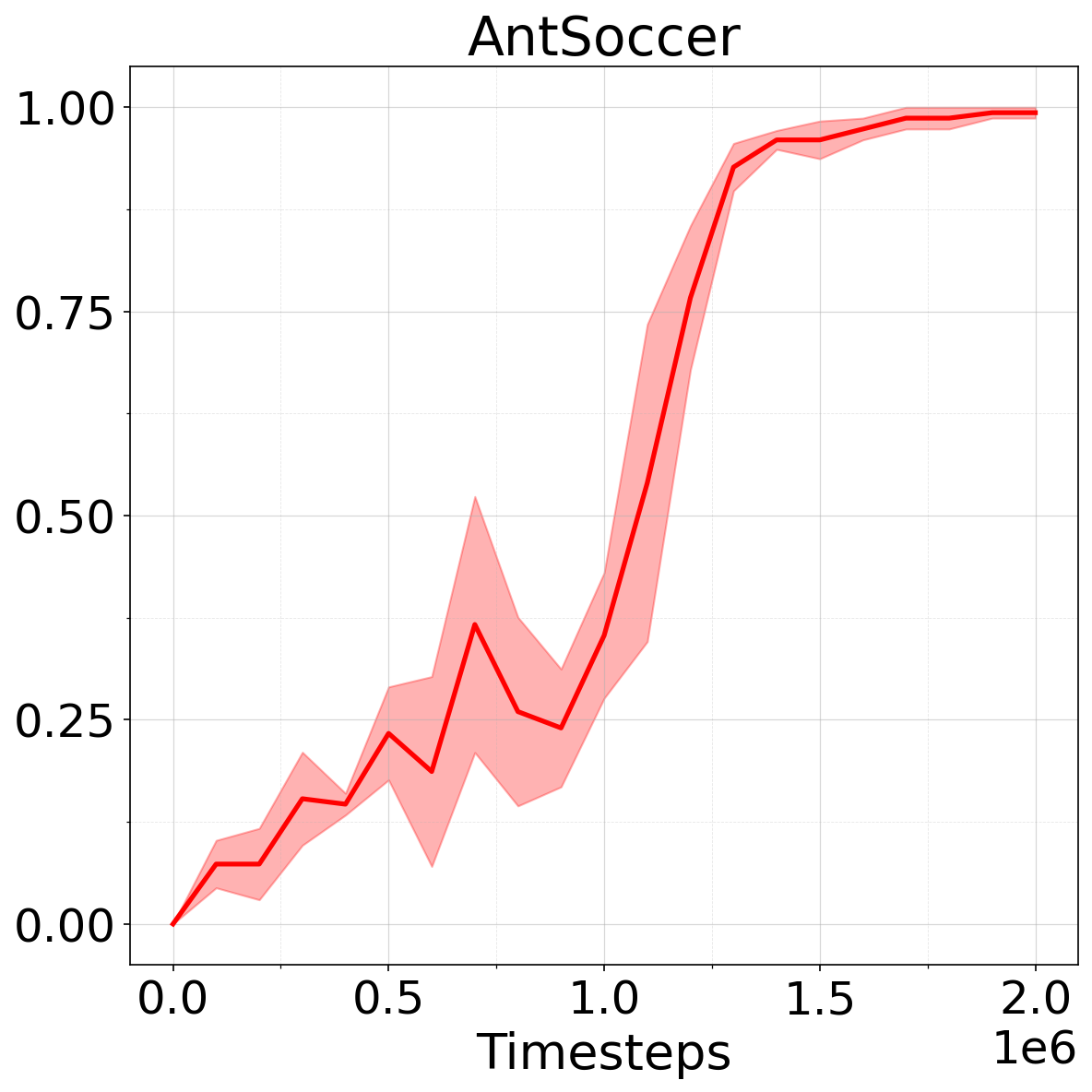}
     \end{subfigure}
     \hfill
     \begin{subfigure}[b]{0.24\textwidth}
         \centering
         \includegraphics[width=\textwidth]{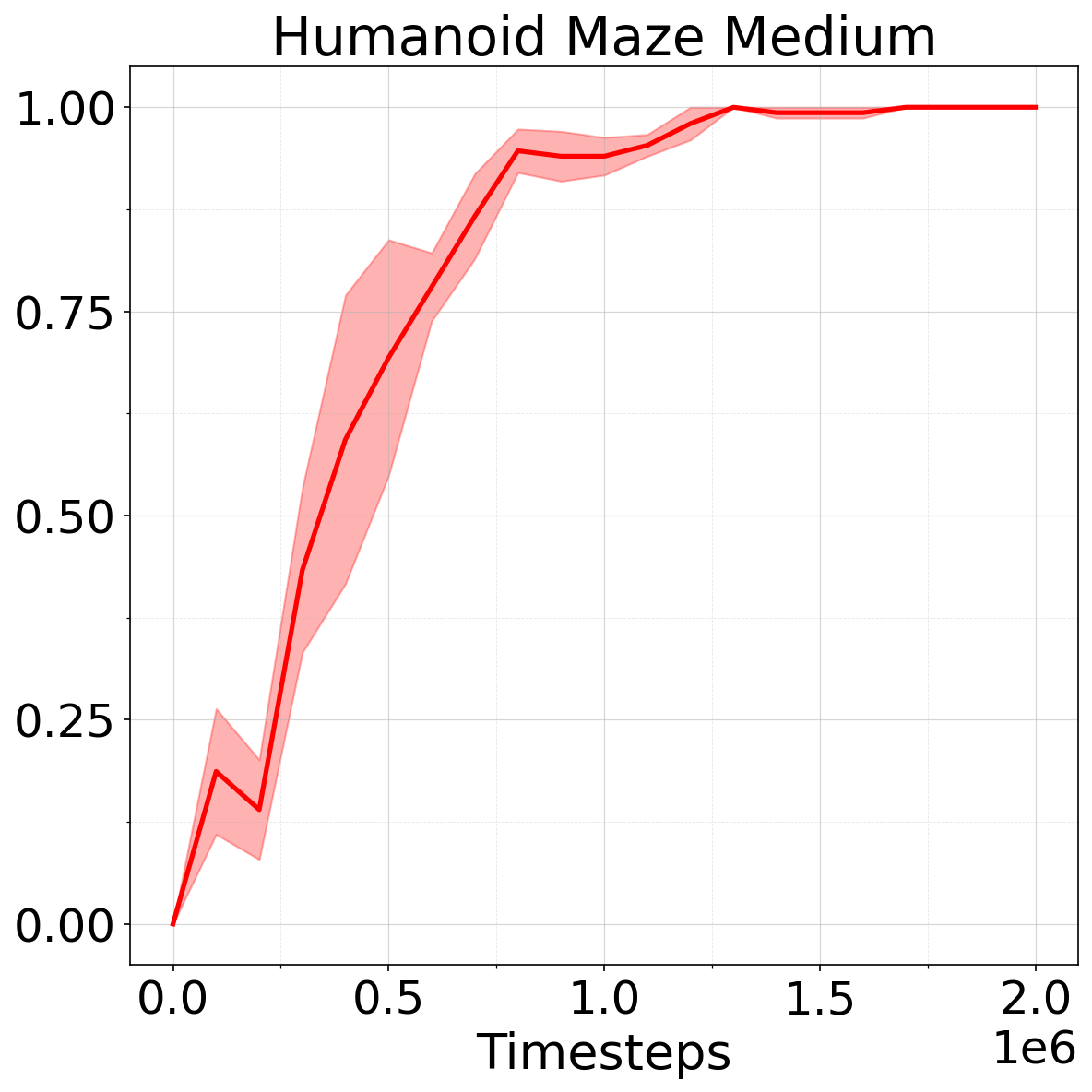}
     \end{subfigure}
     \hfill
     \begin{subfigure}[b]{0.24\textwidth}
         \centering
         \includegraphics[width=\textwidth]{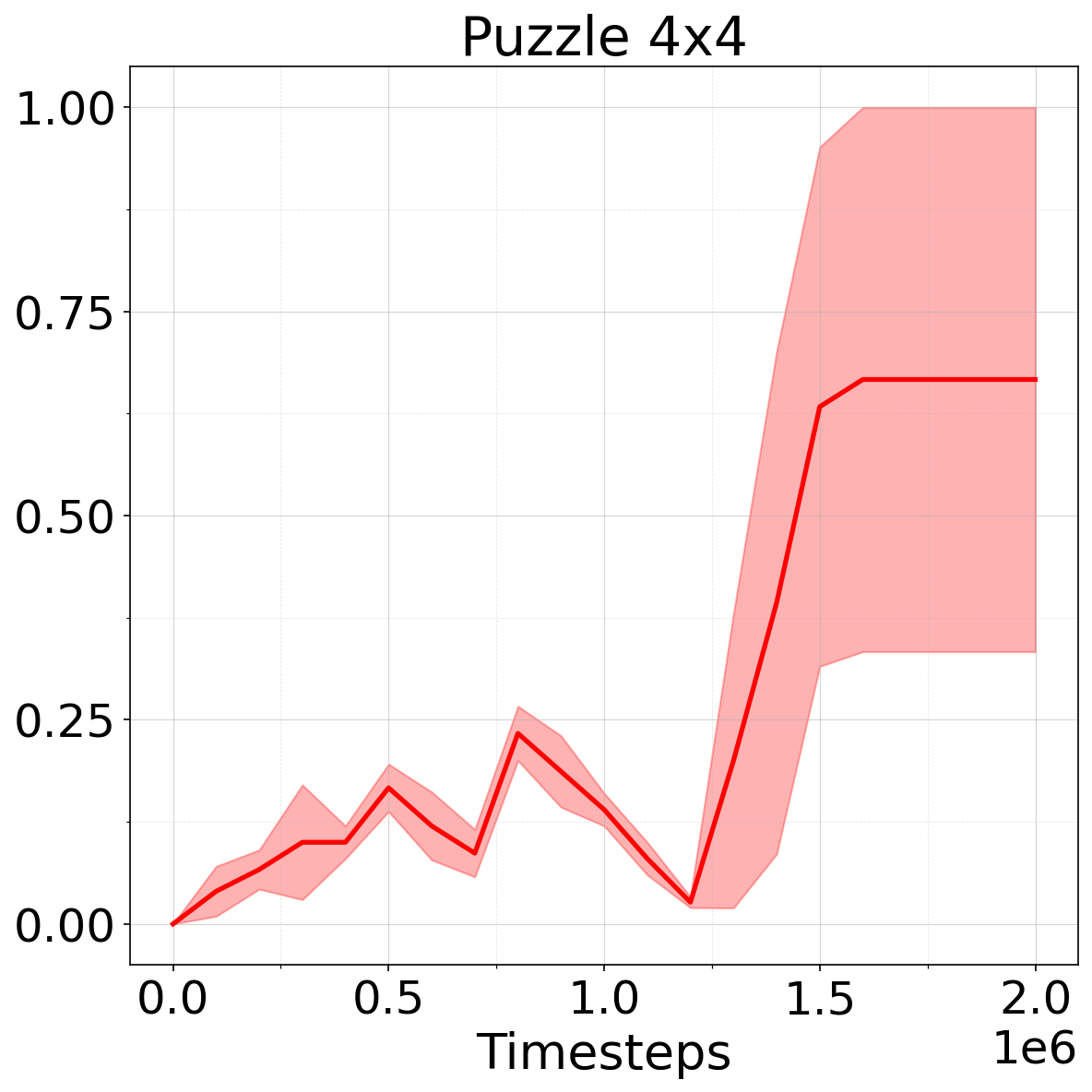}
     \end{subfigure}
        \caption{Training evaluation plots of LARBRID-I. The first 1M steps is pure offline training and 1M - 2M steps are trained online.}
        \label{fig:ogbench-larbrid-i-o2o}
\end{figure}

\begin{figure}[htbp]
     \centering
     \begin{subfigure}[b]{0.24\textwidth}
         \centering
         \includegraphics[width=\textwidth]{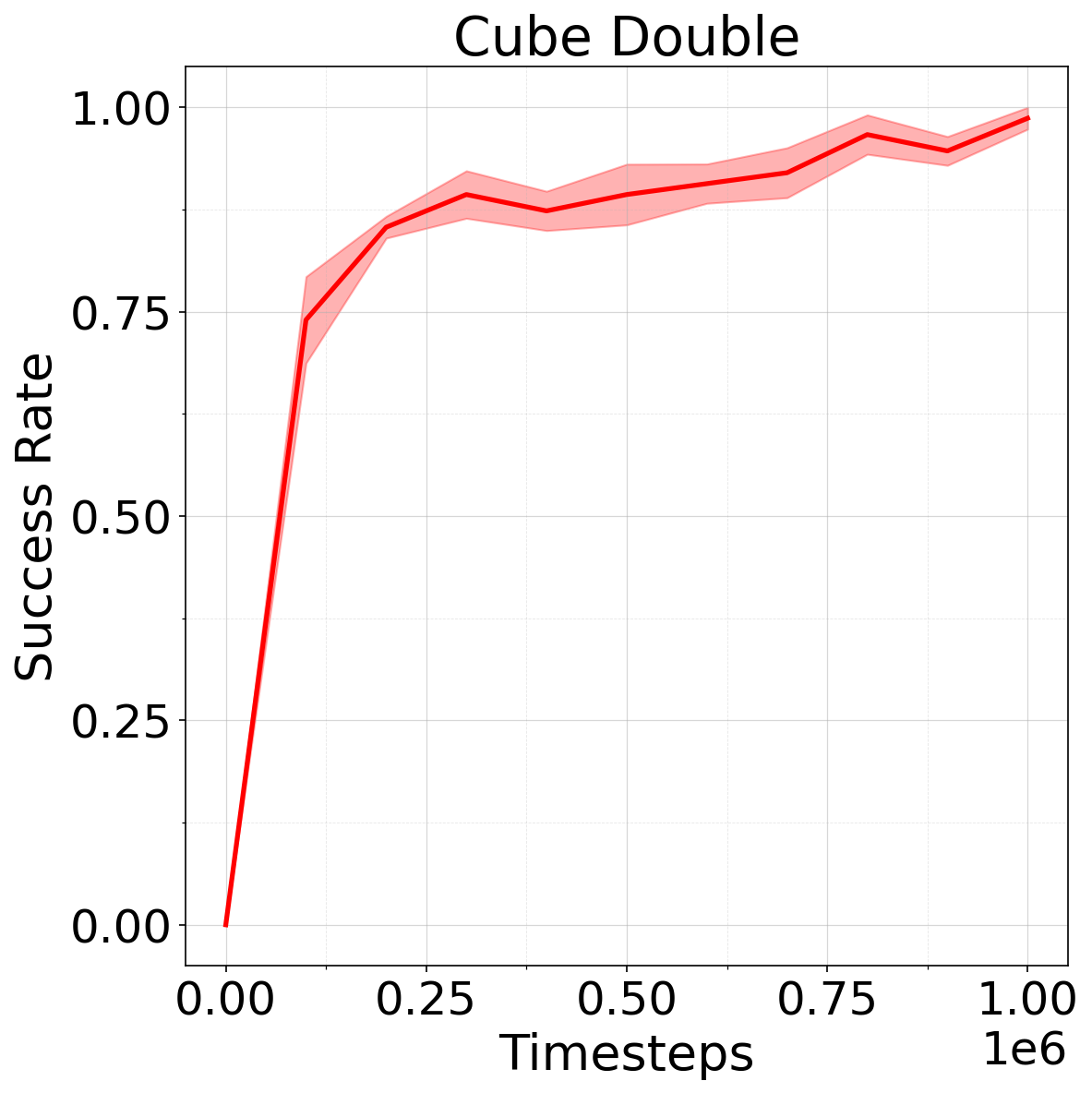}
     \end{subfigure}
     \hfill
     \begin{subfigure}[b]{0.24\textwidth}
         \centering
         \includegraphics[width=\textwidth]{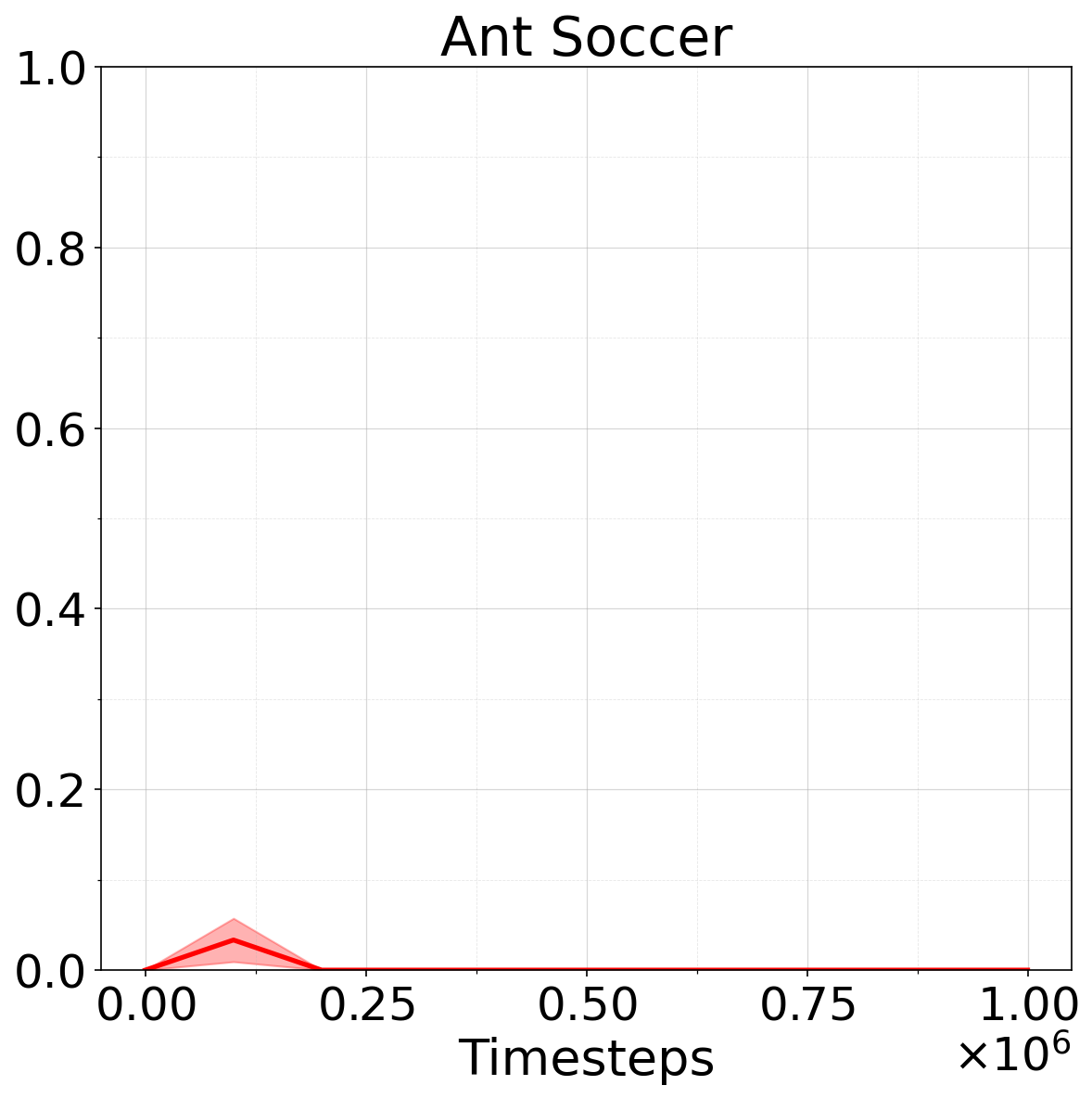}
     \end{subfigure}
     \hfill
     \begin{subfigure}[b]{0.24\textwidth}
         \centering
         \includegraphics[width=\textwidth]{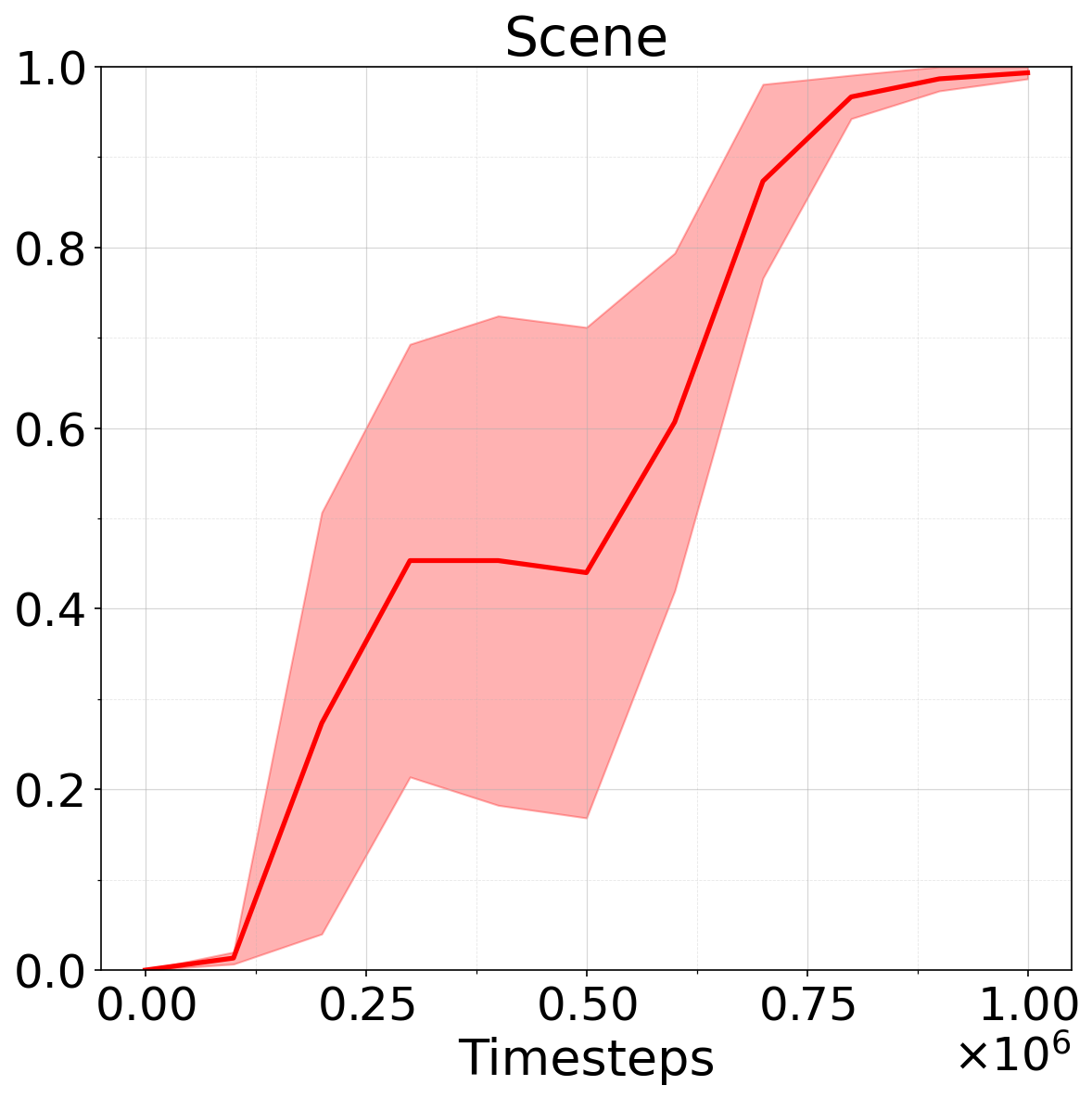}
     \end{subfigure}
     \hfill
     \begin{subfigure}[b]{0.24\textwidth}
         \centering
         \includegraphics[width=\textwidth]{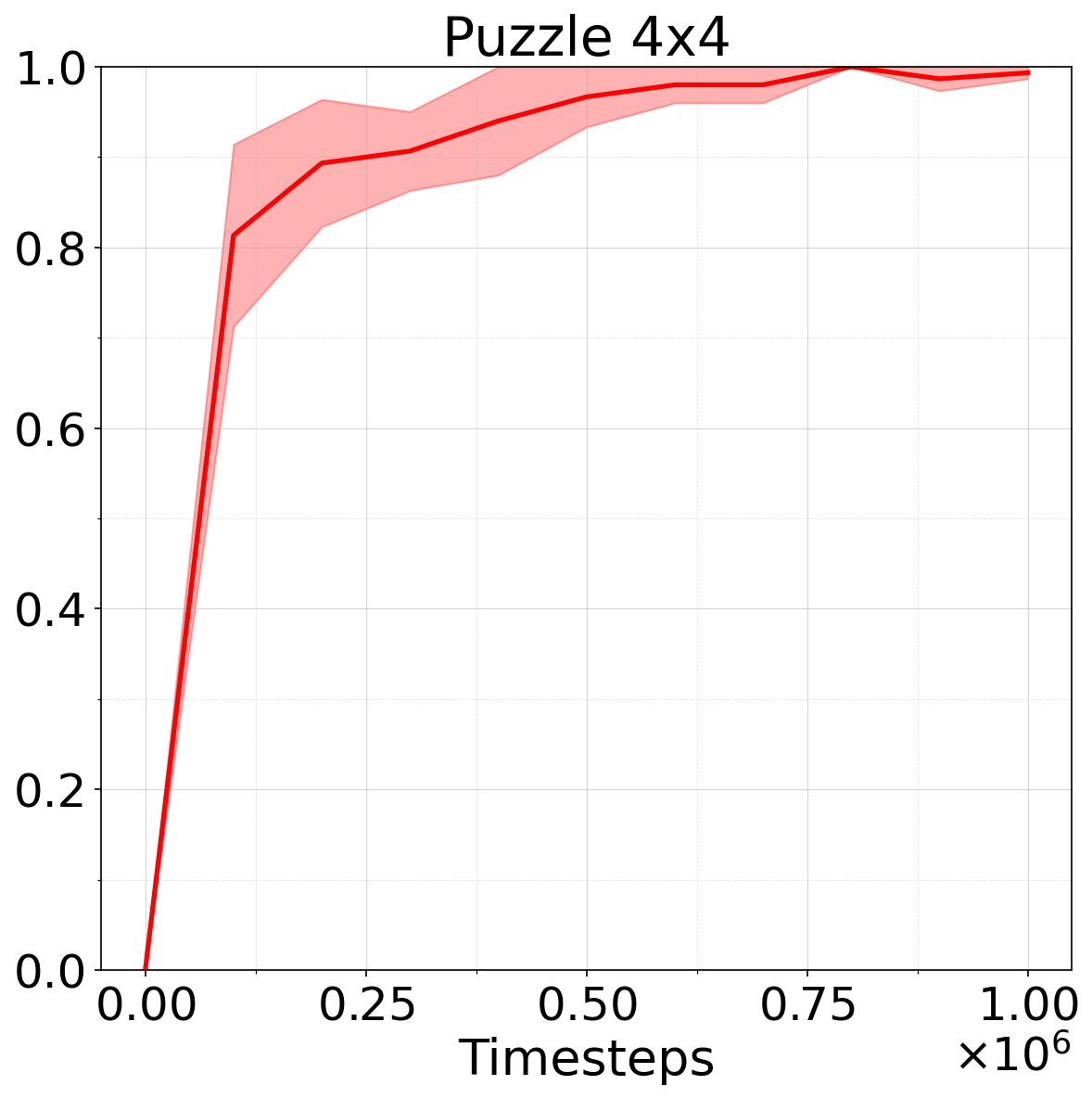}
     \end{subfigure}
        \caption{Training evaluation plots of LARBRID. It trains the BC model from FQL with 1M online steps.}
        \label{fig:ogbench-larbrid-o2o}
\end{figure}

The training evaluation plots of LARBRID and LARBRID-I are displayed in Figure~\ref{fig:ogbench-larbrid-o2o} and \ref{fig:ogbench-larbrid-i-o2o} respectively. LARBRID uses the pre-trained BC model from FQL (and offline dataset) and fine tune it online with 1M steps. LARBRID-I is trained offline first with 1M steps and then online fine tuned for another 1M steps.

\subsection{Details of Fine Tuning Foundation Models}
We use the public checkpoint at \url{s3://openpi-assets/checkpoints/pi0 libero} for the Libero tasks and the checkpoint at \url{s3://openpi-assets/checkpoints/pi0 aloha sim} for the \texttt{AlohaTransferCube} task. Both Libero tasks are pulled from the Libero-90 pool with task ID 20 for `turn on stove' and task ID 7 for `open drawer'. We use the language instruction `open the top drawer of the cabinet' for task 7 and the instruction `turn on the stove' for task 20. To ensure fairness in comparison, we follow the same design as DSRL for the shared components. We train LARBRID on a single step noise and during execution we repeat the same sample noise across all action chunks to reduce the challenge created from high dimensional action chunks. Meanwhile, we use 10 critics with mean aggregation to stabilize the TD errors, however, we only use a single DDPM to learn the representations. Unlike the visual manipulation tasks in OGBench, here we only use a CNN encoder to encode the image observation to a lower dimensional latent space. This image embedding is concatenated with the robot proprioceptive state inputs and fed to the DDPM loss. Hence, the image encoder is trained using the DDPM loss. To reduce the computation burden, we deploy LARBRID-$\psi$ instead of LARBRID to learn representations straightaway from the noise space. We found it is important to use spatial softmax and group normalisation for the image encoder output, as better task relevant information is kept.

\subsection{More Ablation Studies}\label{appendix: ablations}
\begin{figure}[htbp]
 \vspace{-2mm}
     \centering
     \begin{subfigure}[b]{0.23\textwidth}
         \centering
         \subcaption{\scriptsize SC on Halfcheetah}
         \includegraphics[width=\textwidth, height=3.3cm]{figs/halfcheetah_ablation_sc.png}
     \end{subfigure}
     \hfill
     \begin{subfigure}[b]{0.23\textwidth}
         \centering
         \subcaption{\scriptsize SC on Walker}
         \includegraphics[width=\textwidth]{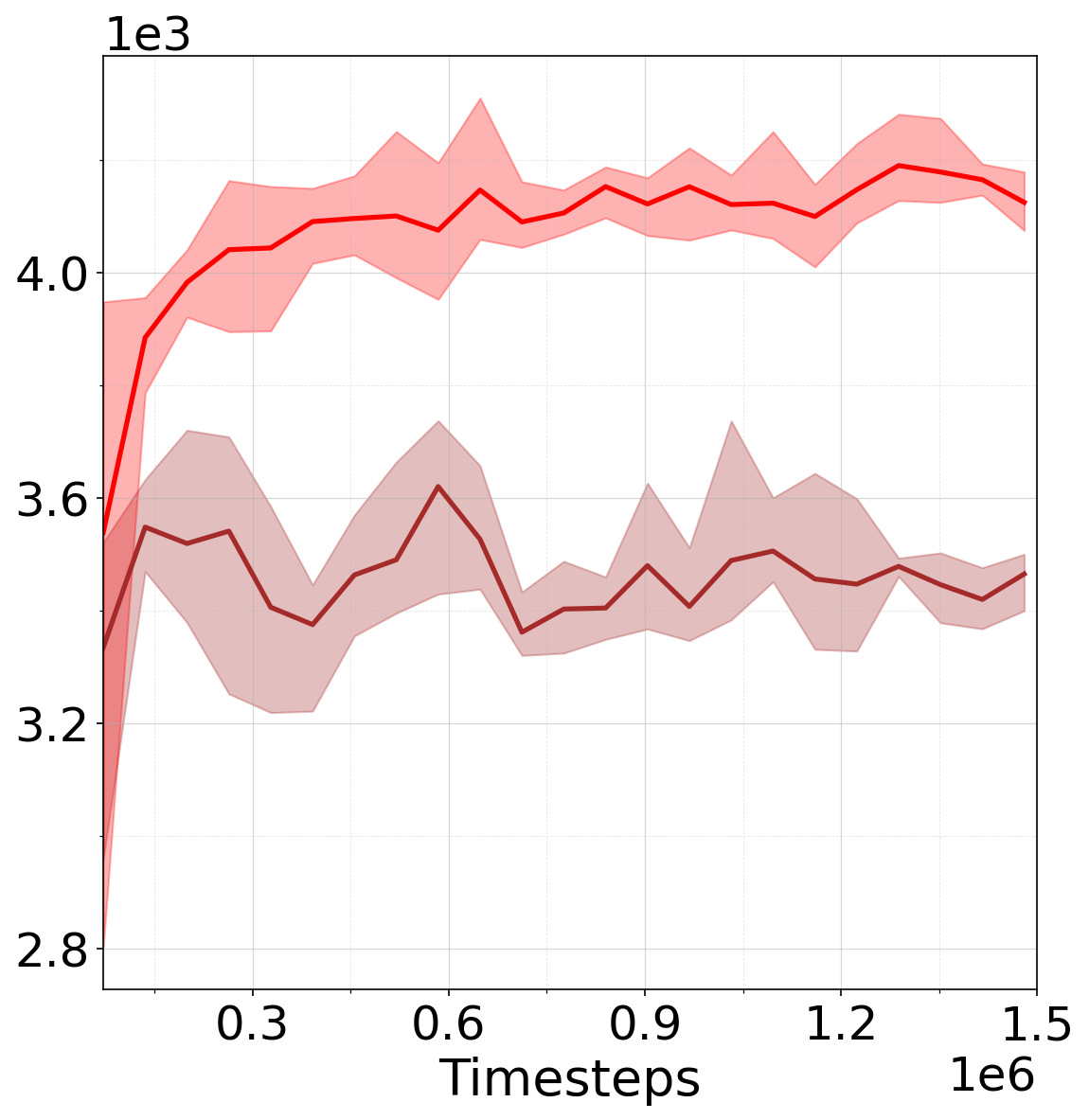}
     \end{subfigure}
     \hfill
     \begin{subfigure}[b]{0.23\textwidth}
         \centering
         \subcaption{\scriptsize Representation Sensitivity}
         \includegraphics[width=\textwidth]{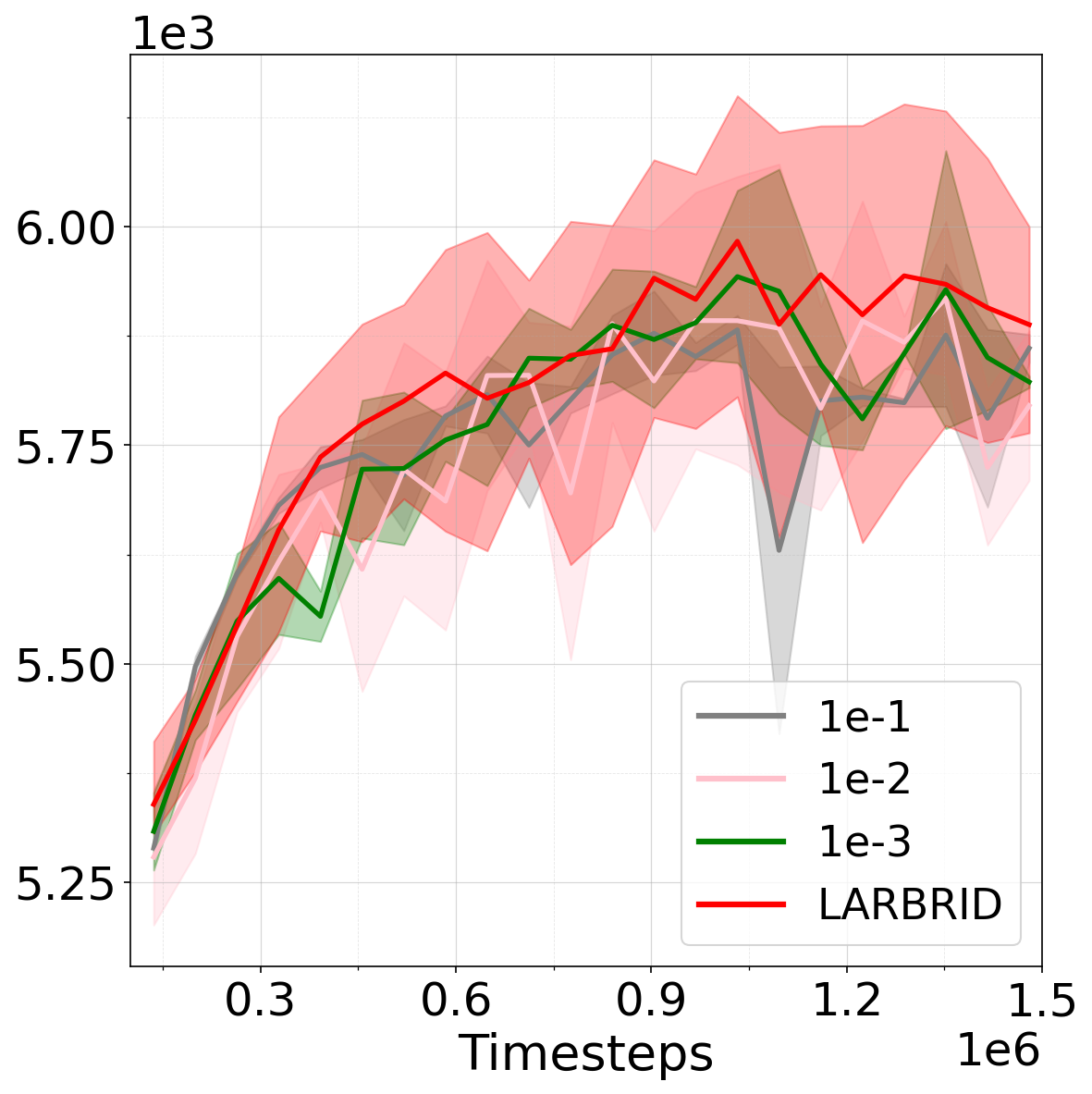}
     \end{subfigure}
     \hfill
     \begin{subfigure}[b]{0.23\textwidth}
         \centering
         \subcaption{\scriptsize Feature Matching Sensitivity}
         \includegraphics[width=\textwidth]{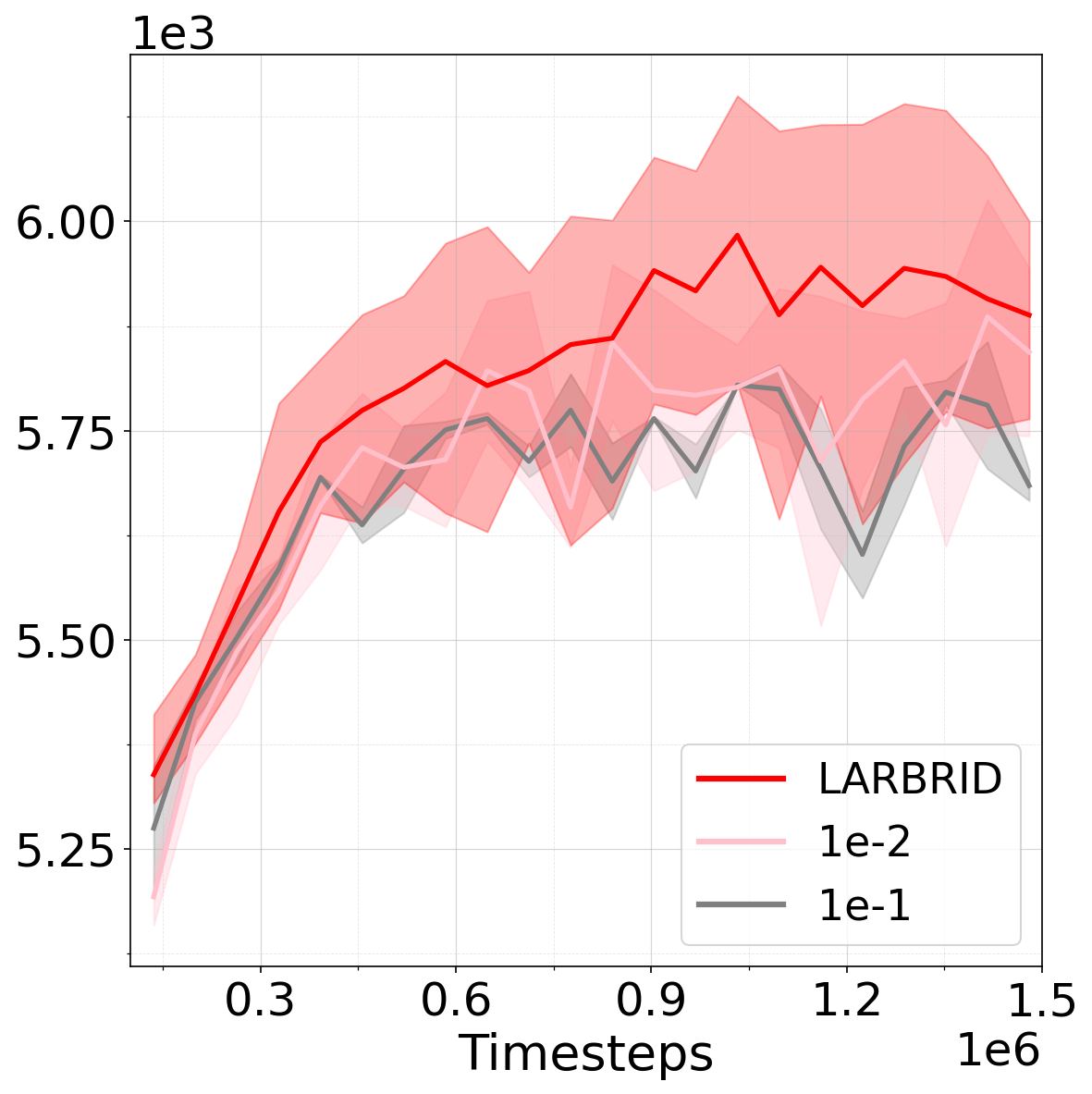}
     \end{subfigure}
\caption{Ablations on the effect of the Bridge Equation using LARBRID-SC are shown in (a) and (b). Sensitivity tests over representation loss and feature matching loss are shown in (c) and (d) respectively}\label{fig: ablation plots}
\end{figure}

The separate critic ablation studies mentioned in the main text are shown in (a) and (b) of Figure~\ref{fig: ablation plots}. Additionally, we further perform a sensitivity analysis on the learned representation $\phi$ from LARBRID. We add $\{0.1, 0.01, 0.001, 1\}$ weights to the representation loss Eq.~\ref{eq: feature score match} while keeping everything else the same as LARBRID. Note $1$ loss weight refers to the original LARBRID. All four variants produce roughly the same training curves as LARBRID as shown in Figure~\ref{fig: ablation plots}. We also perform sensitivity analysis on the MSE feature matching loss (line 5 of LARBRID) across $\{0.1, 0.01, 1\}$ loss weightings. As shown in Figure~\ref{fig: ablation plots}, there is no obvious drop in performance across all tests, hinting that our algorithm is quite robust to the fluctuations and errors from the learned representation and feature matching. LARBRID does not need to work with the restrictive settings studied in the theoretical analysis.

\section{Hyperparameters}\label{appendix:hyperparameters}
We summarize the hyperparameters used in the table below. Since our algorithm is general, we found sticking to the same parameters for the shared components with the strong baselines works well for LARBRID and LARBRID-I. We list specific hyperparameters for each experiment below. If we do not mention the specific experiment, it means the hyperparameter works across all experiments. We use $A_d$ to denote the action dimension.

\begin{table}[htpb]
\centering
\caption{Hyperparameters}
\label{tab: hyperparameters}
\begin{tabular}{ll}
\hline
\textbf{Hyperparameters}                     & \textbf{Value}          \\ \hline
representation learning rates                & 1e-4                    \\
other learning rates                         & 3e-4                    \\
Target update rate ($\tau$)                  & 5e-3                    \\
Target entropy $\bar{H}$ (online)            & 0                       \\
Target entropy $\bar{H}$ ($\pi_0$)           & $A_d / 2$              \\
Critic and actor \#. of hidden layers        & 2                       \\
Critic and actor hidden layer dimension      & 512                     \\
$\phi_{\theta_1}$ \#. of hidden layers       & 2                       \\
$\phi_{\theta_1}$ hidden dimension           & 512                     \\
$\psi_{\theta_2}$ \#. of hidden layers       & 2                       \\
$\psi_{\theta_2}$ hidden dimension           & 512                     \\
Feature dimension ($d$)                      & 512                     \\
\#. of critics (online + offline)            & 2                       \\ 
\#. of critics ($\pi_0$)                     & 10                      \\
Batch size                                   & 256                     \\
DDPM Sampling Steps                          & 50                      \\
Flow Steps (OGBench + $\pi_0$)               & 10                      \\
CNN Layers (visual tasks only)               & (32, 32, 32, 32)        \\
Image Embed Dimension                        & 128                     \\
Action Chunk Size (online)                   & 4 (8 for transport)     \\
Action Chunk Size (offline)                  & 1                       \\
Action Chunk Size ($\pi_0$)                  & 50                      \\
Gradient Steps per Update (online + $\pi_0$) & 20                      \\
Gradient Steps per Update (offline)          & 1                       \\
Image Observation Size (visual)              & $64 \times 64$          \\
Discount Factor (online)                     & 0.99                    \\
Discount Factor ($\pi_0$)                    & 0.999                   \\
Q Aggregation (online + offline)             & Min                     \\
Q Aggregation ($\pi_0$)                      & Mean                     \\
\hline
\end{tabular}
\end{table}

For LARBRID-I, the hyperparameters of the BC flow model, discount factors, and the behavior regularization coefficient $\alpha$ follow exactly the same as those recommended by \cite{fql_park2025}. Meanwhile, we found using the same dimension for the feature, critic and actors work well for fine tuning $\pi_0$, but we use the same hyperparameters as DSRL with image embed dimension 50 and layer dimension 128 across feature and hidden layers for fairness of comparisons.

\newpage
\section*{NeurIPS Paper Checklist}



\begin{enumerate}

\item {\bf Claims}
    \item[] Question: Do the main claims made in the abstract and introduction accurately reflect the paper's contributions and scope?
    \item[] Answer: \answerYes{} 
    \item[] Justification: Our theoretical results and empirical results directly reflect this.
    \item[] Guidelines:
    \begin{itemize}
        \item The answer \answerNA{} means that the abstract and introduction do not include the claims made in the paper.
        \item The abstract and/or introduction should clearly state the claims made, including the contributions made in the paper and important assumptions and limitations. A \answerNo{} or \answerNA{} answer to this question will not be perceived well by the reviewers. 
        \item The claims made should match theoretical and experimental results, and reflect how much the results can be expected to generalize to other settings. 
        \item It is fine to include aspirational goals as motivation as long as it is clear that these goals are not attained by the paper. 
    \end{itemize}

\item {\bf Limitations}
    \item[] Question: Does the paper discuss the limitations of the work performed by the authors?
    \item[] Answer: \answerYes{} 
    \item[] Justification: It is discussed in the Discussion section.
    \item[] Guidelines:
    \begin{itemize}
        \item The answer \answerNA{} means that the paper has no limitation while the answer \answerNo{} means that the paper has limitations, but those are not discussed in the paper. 
        \item The authors are encouraged to create a separate ``Limitations'' section in their paper.
        \item The paper should point out any strong assumptions and how robust the results are to violations of these assumptions (e.g., independence assumptions, noiseless settings, model well-specification, asymptotic approximations only holding locally). The authors should reflect on how these assumptions might be violated in practice and what the implications would be.
        \item The authors should reflect on the scope of the claims made, e.g., if the approach was only tested on a few datasets or with a few runs. In general, empirical results often depend on implicit assumptions, which should be articulated.
        \item The authors should reflect on the factors that influence the performance of the approach. For example, a facial recognition algorithm may perform poorly when image resolution is low or images are taken in low lighting. Or a speech-to-text system might not be used reliably to provide closed captions for online lectures because it fails to handle technical jargon.
        \item The authors should discuss the computational efficiency of the proposed algorithms and how they scale with dataset size.
        \item If applicable, the authors should discuss possible limitations of their approach to address problems of privacy and fairness.
        \item While the authors might fear that complete honesty about limitations might be used by reviewers as grounds for rejection, a worse outcome might be that reviewers discover limitations that aren't acknowledged in the paper. The authors should use their best judgment and recognize that individual actions in favor of transparency play an important role in developing norms that preserve the integrity of the community. Reviewers will be specifically instructed to not penalize honesty concerning limitations.
    \end{itemize}

\item {\bf Theory assumptions and proofs}
    \item[] Question: For each theoretical result, does the paper provide the full set of assumptions and a complete (and correct) proof?
    \item[] Answer: \answerYes{} 
    \item[] Justification: We provide full proofs and further materials in the appendix to backup everything stated in the paper.
    \item[] Guidelines:
    \begin{itemize}
        \item The answer \answerNA{} means that the paper does not include theoretical results. 
        \item All the theorems, formulas, and proofs in the paper should be numbered and cross-referenced.
        \item All assumptions should be clearly stated or referenced in the statement of any theorems.
        \item The proofs can either appear in the main paper or the supplemental material, but if they appear in the supplemental material, the authors are encouraged to provide a short proof sketch to provide intuition. 
        \item Inversely, any informal proof provided in the core of the paper should be complemented by formal proofs provided in appendix or supplemental material.
        \item Theorems and Lemmas that the proof relies upon should be properly referenced. 
    \end{itemize}

    \item {\bf Experimental result reproducibility}
    \item[] Question: Does the paper fully disclose all the information needed to reproduce the main experimental results of the paper to the extent that it affects the main claims and/or conclusions of the paper (regardless of whether the code and data are provided or not)?
    \item[] Answer: \answerYes{} 
    \item[] Justification: All algorithms are detailed in the main content or in the appendix. Best practices and hyperparameters are all given in the appendix. We will also release our code after publication. 
    \item[] Guidelines:
    \begin{itemize}
        \item The answer \answerNA{} means that the paper does not include experiments.
        \item If the paper includes experiments, a \answerNo{} answer to this question will not be perceived well by the reviewers: Making the paper reproducible is important, regardless of whether the code and data are provided or not.
        \item If the contribution is a dataset and\slash or model, the authors should describe the steps taken to make their results reproducible or verifiable. 
        \item Depending on the contribution, reproducibility can be accomplished in various ways. For example, if the contribution is a novel architecture, describing the architecture fully might suffice, or if the contribution is a specific model and empirical evaluation, it may be necessary to either make it possible for others to replicate the model with the same dataset, or provide access to the model. In general. releasing code and data is often one good way to accomplish this, but reproducibility can also be provided via detailed instructions for how to replicate the results, access to a hosted model (e.g., in the case of a large language model), releasing of a model checkpoint, or other means that are appropriate to the research performed.
        \item While NeurIPS does not require releasing code, the conference does require all submissions to provide some reasonable avenue for reproducibility, which may depend on the nature of the contribution. For example
        \begin{enumerate}
            \item If the contribution is primarily a new algorithm, the paper should make it clear how to reproduce that algorithm.
            \item If the contribution is primarily a new model architecture, the paper should describe the architecture clearly and fully.
            \item If the contribution is a new model (e.g., a large language model), then there should either be a way to access this model for reproducing the results or a way to reproduce the model (e.g., with an open-source dataset or instructions for how to construct the dataset).
            \item We recognize that reproducibility may be tricky in some cases, in which case authors are welcome to describe the particular way they provide for reproducibility. In the case of closed-source models, it may be that access to the model is limited in some way (e.g., to registered users), but it should be possible for other researchers to have some path to reproducing or verifying the results.
        \end{enumerate}
    \end{itemize}

\item {\bf Open access to data and code}
    \item[] Question: Does the paper provide open access to the data and code, with sufficient instructions to faithfully reproduce the main experimental results, as described in supplemental material?
    \item[] Answer: \answerNo{} 
    \item[] Justification: We will release our code after publication, not during the review stage.
    \item[] Guidelines:
    \begin{itemize}
        \item The answer \answerNA{} means that paper does not include experiments requiring code.
        \item Please see the NeurIPS code and data submission guidelines (\url{https://neurips.cc/public/guides/CodeSubmissionPolicy}) for more details.
        \item While we encourage the release of code and data, we understand that this might not be possible, so \answerNo{} is an acceptable answer. Papers cannot be rejected simply for not including code, unless this is central to the contribution (e.g., for a new open-source benchmark).
        \item The instructions should contain the exact command and environment needed to run to reproduce the results. See the NeurIPS code and data submission guidelines (\url{https://neurips.cc/public/guides/CodeSubmissionPolicy}) for more details.
        \item The authors should provide instructions on data access and preparation, including how to access the raw data, preprocessed data, intermediate data, and generated data, etc.
        \item The authors should provide scripts to reproduce all experimental results for the new proposed method and baselines. If only a subset of experiments are reproducible, they should state which ones are omitted from the script and why.
        \item At submission time, to preserve anonymity, the authors should release anonymized versions (if applicable).
        \item Providing as much information as possible in supplemental material (appended to the paper) is recommended, but including URLs to data and code is permitted.
    \end{itemize}

\item {\bf Experimental setting/details}
    \item[] Question: Does the paper specify all the training and test details (e.g., data splits, hyperparameters, how they were chosen, type of optimizer) necessary to understand the results?
    \item[] Answer: \answerYes{} 
    \item[] Justification: All hyperparameters, experiment settings are reported in the appendix.
    \item[] Guidelines:
    \begin{itemize}
        \item The answer \answerNA{} means that the paper does not include experiments.
        \item The experimental setting should be presented in the core of the paper to a level of detail that is necessary to appreciate the results and make sense of them.
        \item The full details can be provided either with the code, in appendix, or as supplemental material.
    \end{itemize}

\item {\bf Experiment statistical significance}
    \item[] Question: Does the paper report error bars suitably and correctly defined or other appropriate information about the statistical significance of the experiments?
    \item[] Answer: \answerYes{} 
    \item[] Justification: Each experiment is run with 4 random seeds and the variances and means are reported and compared with other baselines.
    \item[] Guidelines:
    \begin{itemize}
        \item The answer \answerNA{} means that the paper does not include experiments.
        \item The authors should answer \answerYes{} if the results are accompanied by error bars, confidence intervals, or statistical significance tests, at least for the experiments that support the main claims of the paper.
        \item The factors of variability that the error bars are capturing should be clearly stated (for example, train/test split, initialization, random drawing of some parameter, or overall run with given experimental conditions).
        \item The method for calculating the error bars should be explained (closed form formula, call to a library function, bootstrap, etc.)
        \item The assumptions made should be given (e.g., Normally distributed errors).
        \item It should be clear whether the error bar is the standard deviation or the standard error of the mean.
        \item It is OK to report 1-sigma error bars, but one should state it. The authors should preferably report a 2-sigma error bar than state that they have a 96\% CI, if the hypothesis of Normality of errors is not verified.
        \item For asymmetric distributions, the authors should be careful not to show in tables or figures symmetric error bars that would yield results that are out of range (e.g., negative error rates).
        \item If error bars are reported in tables or plots, the authors should explain in the text how they were calculated and reference the corresponding figures or tables in the text.
    \end{itemize}

\item {\bf Experiments compute resources}
    \item[] Question: For each experiment, does the paper provide sufficient information on the computer resources (type of compute workers, memory, time of execution) needed to reproduce the experiments?
    \item[] Answer: \answerYes{} 
    \item[] Justification: Yes, we state the resources and time needed in the Experiment and Appendix section.
    \item[] Guidelines:
    \begin{itemize}
        \item The answer \answerNA{} means that the paper does not include experiments.
        \item The paper should indicate the type of compute workers CPU or GPU, internal cluster, or cloud provider, including relevant memory and storage.
        \item The paper should provide the amount of compute required for each of the individual experimental runs as well as estimate the total compute. 
        \item The paper should disclose whether the full research project required more compute than the experiments reported in the paper (e.g., preliminary or failed experiments that didn't make it into the paper). 
    \end{itemize}
    
\item {\bf Code of ethics}
    \item[] Question: Does the research conducted in the paper conform, in every respect, with the NeurIPS Code of Ethics \url{https://neurips.cc/public/EthicsGuidelines}?
    \item[] Answer: \answerYes{} 
    \item[] Justification: We have followed the code of ethics.
    \item[] Guidelines:
    \begin{itemize}
        \item The answer \answerNA{} means that the authors have not reviewed the NeurIPS Code of Ethics.
        \item If the authors answer \answerNo, they should explain the special circumstances that require a deviation from the Code of Ethics.
        \item The authors should make sure to preserve anonymity (e.g., if there is a special consideration due to laws or regulations in their jurisdiction).
    \end{itemize}

\item {\bf Broader impacts}
    \item[] Question: Does the paper discuss both potential positive societal impacts and negative societal impacts of the work performed?
    \item[] Answer: \answerNA{} 
    \item[] Justification: No societal work is performed.
    \item[] Guidelines:
    \begin{itemize}
        \item The answer \answerNA{} means that there is no societal impact of the work performed.
        \item If the authors answer \answerNA{} or \answerNo, they should explain why their work has no societal impact or why the paper does not address societal impact.
        \item Examples of negative societal impacts include potential malicious or unintended uses (e.g., disinformation, generating fake profiles, surveillance), fairness considerations (e.g., deployment of technologies that could make decisions that unfairly impact specific groups), privacy considerations, and security considerations.
        \item The conference expects that many papers will be foundational research and not tied to particular applications, let alone deployments. However, if there is a direct path to any negative applications, the authors should point it out. For example, it is legitimate to point out that an improvement in the quality of generative models could be used to generate Deepfakes for disinformation. On the other hand, it is not needed to point out that a generic algorithm for optimizing neural networks could enable people to train models that generate Deepfakes faster.
        \item The authors should consider possible harms that could arise when the technology is being used as intended and functioning correctly, harms that could arise when the technology is being used as intended but gives incorrect results, and harms following from (intentional or unintentional) misuse of the technology.
        \item If there are negative societal impacts, the authors could also discuss possible mitigation strategies (e.g., gated release of models, providing defenses in addition to attacks, mechanisms for monitoring misuse, mechanisms to monitor how a system learns from feedback over time, improving the efficiency and accessibility of ML).
    \end{itemize}
    
\item {\bf Safeguards}
    \item[] Question: Does the paper describe safeguards that have been put in place for responsible release of data or models that have a high risk for misuse (e.g., pre-trained language models, image generators, or scraped datasets)?
    \item[] Answer: \answerNA{} 
    \item[] Justification: no such risk.
    \item[] Guidelines:
    \begin{itemize}
        \item The answer \answerNA{} means that the paper poses no such risks.
        \item Released models that have a high risk for misuse or dual-use should be released with necessary safeguards to allow for controlled use of the model, for example by requiring that users adhere to usage guidelines or restrictions to access the model or implementing safety filters. 
        \item Datasets that have been scraped from the Internet could pose safety risks. The authors should describe how they avoided releasing unsafe images.
        \item We recognize that providing effective safeguards is challenging, and many papers do not require this, but we encourage authors to take this into account and make a best faith effort.
    \end{itemize}

\item {\bf Licenses for existing assets}
    \item[] Question: Are the creators or original owners of assets (e.g., code, data, models), used in the paper, properly credited and are the license and terms of use explicitly mentioned and properly respected?
    \item[] Answer: \answerYes{} 
    \item[] Justification: References and citations are given.
    \item[] Guidelines:
    \begin{itemize}
        \item The answer \answerNA{} means that the paper does not use existing assets.
        \item The authors should cite the original paper that produced the code package or dataset.
        \item The authors should state which version of the asset is used and, if possible, include a URL.
        \item The name of the license (e.g., CC-BY 4.0) should be included for each asset.
        \item For scraped data from a particular source (e.g., website), the copyright and terms of service of that source should be provided.
        \item If assets are released, the license, copyright information, and terms of use in the package should be provided. For popular datasets, \url{paperswithcode.com/datasets} has curated licenses for some datasets. Their licensing guide can help determine the license of a dataset.
        \item For existing datasets that are re-packaged, both the original license and the license of the derived asset (if it has changed) should be provided.
        \item If this information is not available online, the authors are encouraged to reach out to the asset's creators.
    \end{itemize}

\item {\bf New assets}
    \item[] Question: Are new assets introduced in the paper well documented and is the documentation provided alongside the assets?
    \item[] Answer: \answerYes{} 
    \item[] Justification: Pseudoscope is given in full with detailed explanations. 
    \item[] Guidelines:
    \begin{itemize}
        \item The answer \answerNA{} means that the paper does not release new assets.
        \item Researchers should communicate the details of the dataset\slash code\slash model as part of their submissions via structured templates. This includes details about training, license, limitations, etc. 
        \item The paper should discuss whether and how consent was obtained from people whose asset is used.
        \item At submission time, remember to anonymize your assets (if applicable). You can either create an anonymized URL or include an anonymized zip file.
    \end{itemize}

\item {\bf Crowdsourcing and research with human subjects}
    \item[] Question: For crowdsourcing experiments and research with human subjects, does the paper include the full text of instructions given to participants and screenshots, if applicable, as well as details about compensation (if any)? 
    \item[] Answer: \answerNA{} 
    \item[] Justification: no such work is performed.
    \item[] Guidelines:
    \begin{itemize}
        \item The answer \answerNA{} means that the paper does not involve crowdsourcing nor research with human subjects.
        \item Including this information in the supplemental material is fine, but if the main contribution of the paper involves human subjects, then as much detail as possible should be included in the main paper. 
        \item According to the NeurIPS Code of Ethics, workers involved in data collection, curation, or other labor should be paid at least the minimum wage in the country of the data collector. 
    \end{itemize}

\item {\bf Institutional review board (IRB) approvals or equivalent for research with human subjects}
    \item[] Question: Does the paper describe potential risks incurred by study participants, whether such risks were disclosed to the subjects, and whether Institutional Review Board (IRB) approvals (or an equivalent approval/review based on the requirements of your country or institution) were obtained?
    \item[] Answer: \answerNA{} 
    \item[] Justification: We do not consider any crowdsourcing nor research with human subjects.
    \item[] Guidelines:
    \begin{itemize}
        \item The answer \answerNA{} means that the paper does not involve crowdsourcing nor research with human subjects.
        \item Depending on the country in which research is conducted, IRB approval (or equivalent) may be required for any human subjects research. If you obtained IRB approval, you should clearly state this in the paper. 
        \item We recognize that the procedures for this may vary significantly between institutions and locations, and we expect authors to adhere to the NeurIPS Code of Ethics and the guidelines for their institution. 
        \item For initial submissions, do not include any information that would break anonymity (if applicable), such as the institution conducting the review.
    \end{itemize}

\item {\bf Declaration of LLM usage}
    \item[] Question: Does the paper describe the usage of LLMs if it is an important, original, or non-standard component of the core methods in this research? Note that if the LLM is used only for writing, editing, or formatting purposes and does \emph{not} impact the core methodology, scientific rigor, or originality of the research, declaration is not required.
    \item[] Answer: \answerNA{} 
    \item[] Justification: No LLM is involved with the core method.
    \item[] Guidelines:
    \begin{itemize}
        \item The answer \answerNA{} means that the core method development in this research does not involve LLMs as any important, original, or non-standard components.
        \item Please refer to our LLM policy in the NeurIPS handbook for what should or should not be described.
    \end{itemize}

\end{enumerate}

\end{document}